\documentclass{article} 
\usepackage{iclr2027_conference,times}

\iclrfinalcopy
\newif\ifpreprint
\preprinttrue

\usepackage[utf8]{inputenc}
\usepackage[T1]{fontenc}
\usepackage{url}
\usepackage{xurl}
\usepackage{booktabs}
\usepackage{amsfonts}
\usepackage{amssymb}
\usepackage{amsmath}
\usepackage{nicefrac}
\usepackage{microtype}
\usepackage{xcolor}
\usepackage{graphicx}
\usepackage{subcaption}
\usepackage{enumitem}
\usepackage{multirow}
\usepackage{makecell}
\usepackage{tikz}
\usepackage{pgf-pie}
\usepackage{pifont}
\usepackage{wrapfig}
\usepackage{float}
\usepackage{longtable}
\usepackage{array}
\usepackage{seqsplit}
\usepackage{listings}
\usepackage{hyperref}

\ificlrfinal
  \ifpreprint
  \else
  \fi
\else
\fi

\usetikzlibrary{arrows.meta, positioning, shapes.geometric, fit, backgrounds}

\newcommand{\fastkernels}{\textsc{FastKernels}}

\title{FastKernels: Benchmarking GPU Kernel\\Generation in Production}

\author{\ignorespaces
  Gabriele Oliaro$^{1,2}$\thanks{Equal contribution.} \quad
  Jaeseong Lee$^{1*}$ \quad
  Yichao Fu$^{3*}$ \quad
  Zhaoyuan Su$^{1}$ \\
  \bf
  Owen Lu$^{2}$ \quad
  Sreeram Vennam$^{2}$ \quad
  May Jiang$^{4}$ \quad
  Junli Wang$^{3}$ \\
  \bf
  Hao Zhang$^{3}$ \quad
  Zhihao Jia$^{2}$ \quad
  Samyam Rajbhandari$^{1}$ \\
  $^{1}$Snowflake AI Research \quad
  $^{2}$CMU \quad
  $^{3}$UCSD \quad
  $^{4}$Independent Researcher
}

\begin{document}

\maketitle

\begin{abstract}
LLM-based agents for GPU kernel generation are advancing rapidly, but the benchmarks they optimize against evaluate kernels in isolation, with synthetic inputs and weak baselines, rewarding sandbox speedups that break or vanish in real inference systems.
We introduce \fastkernels{}, a benchmark of \textbf{384 tasks} drawn from 47 representative architectures across 8 categories, whose kernels suffice to reimplement \textbf{94.6\% (472/499)} of HuggingFace Transformers architectures with outputs matching the native implementations.
Each task mirrors the interface of the corresponding production module and is scored against the kernels production frameworks ship, and tasks form a compositional hierarchy, from primitives to full models, in which higher-level modules import lower-level ones.
Candidates are scored at the kernel level and end to end inside the models they come from, on the production execution path, and \textsc{MacroEval} aggregates calibrated correctness, coverage, and speedup into a leaderboard.
Seeding it with five representative agents (6{,}900 agent-hours), we find that kernel-level speedups of $1.6$--$6.6\times$ shrink to \textbf{at most $1.25\times$} end to end, \textbf{only 20\%} of winning kernel sets run correctly as-is, and kernel-level scores \textbf{mis-rank agents}: Claude Code matches or beats KDA at every level in isolation, yet KDA scores $3\times$ higher end to end.
\ificlrfinal
Code is available at \url{https://github.com/Snowflake-AI-Research/fastkernels}.
\else
Anonymized code is included in the supplementary material and will be open-sourced upon publication.
\fi
\end{abstract}

\section{Introduction}
\label{sec:introduction}

LLM-based agents for GPU kernel generation now achieve strong scores on isolated benchmarks~\citep{kernelbench, solexecbench, flashinferbench}, with state-of-the-art systems autonomously compiling, profiling, and refining CUDA or Triton code.
Yet these gains often fail to transfer into production inference frameworks such as vLLM~\citep{vllm} and SGLang~\citep{sglang}: kernels that score well in sandbox evaluation routinely regress once they encounter real serving interfaces, compilation stacks, and workloads.

The root cause is \textbf{benchmark--production misalignment}.
Current benchmarks rely heavily on synthetic inputs, single-GPU isolated kernels, simplified interfaces, and independent task levels that do not compose into full inference pipelines.
They therefore reward kernels that are fast in isolation but brittle in deployed systems, hiding interface mismatches, compilation-stack conflicts, and correctness degradation that only appears at model scale.

\subsection*{Our approach}

\begin{figure*}[t]
\centering
\includegraphics[width=\linewidth]{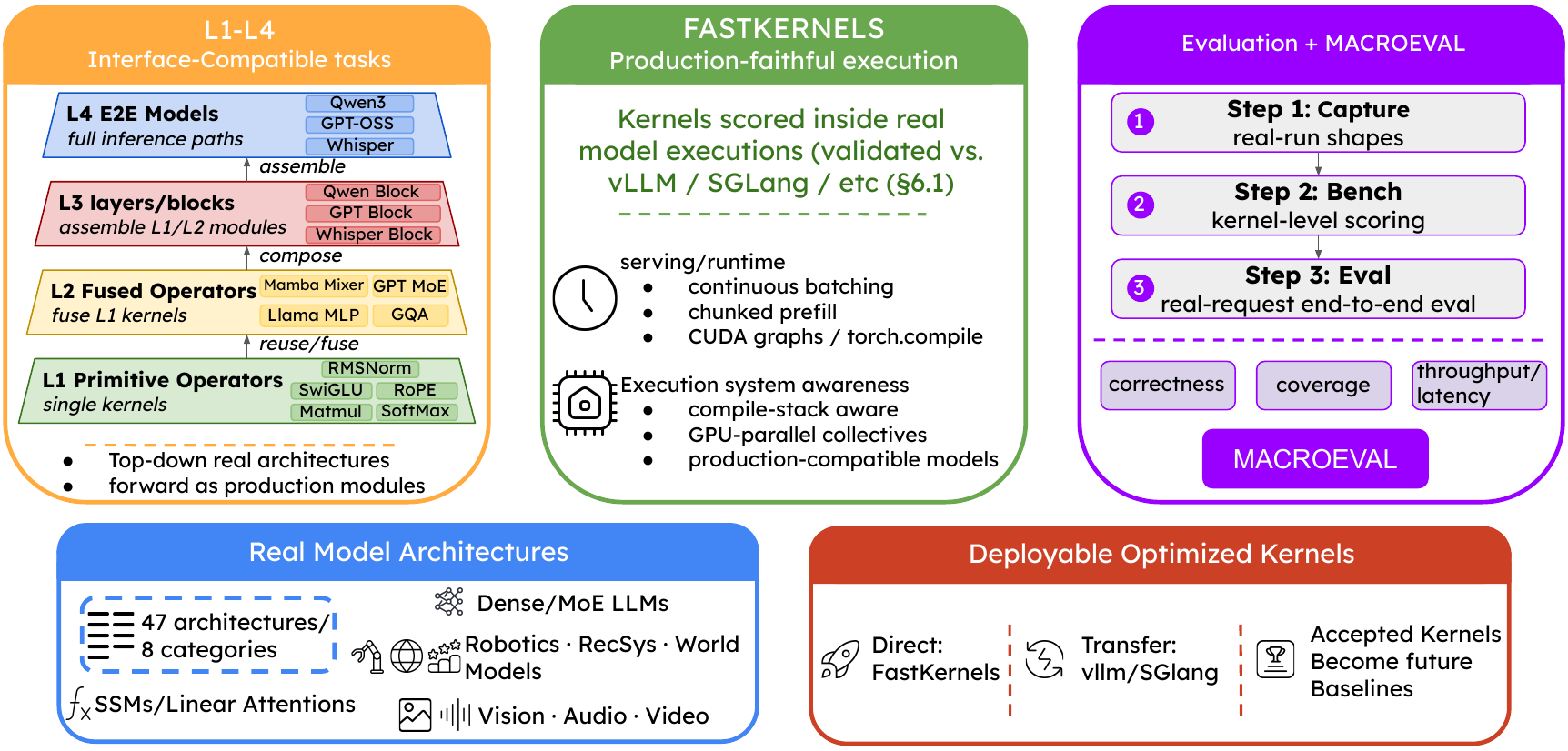}
\caption{\textbf{\fastkernels{} overview.} Tasks derived from 47 real architectures (bottom left) form a compositional L1--L4 hierarchy with production interfaces (top left) and run inside validated production model executions (center). Candidates are scored by capture, kernel-level bench, and end-to-end eval, aggregated by \textsc{MacroEval} (right); optimized kernels deploy directly or transfer into vLLM and SGLang (bottom right).}
\label{fig:overview}
\end{figure*}

We introduce \fastkernels{} (Figure~\ref{fig:overview}), a benchmark for kernel agents designed so that its scores predict deployment outcomes.
\fastkernels{} derives 384 tasks top-down from 47 real model architectures across 8 categories; each task's interface matches the corresponding module in the production library for its family, and each candidate is scored against the kernels that library ships.
Crucially, its task levels are compositional: Level~1 primitives feed Level~2 fused operators, which feed Level~3 layers and Level~4 models.
Unlike benchmarks whose levels are independent, this hierarchy enables a dynamic-programming style optimization loop: an agent can reuse an optimized lower-level kernel, such as a linear operator, when optimizing higher-level modules such as MLPs or transformer blocks, instead of rediscovering the same building block from scratch.

Because integration failures only surface at model scale, \fastkernels{} also scores every candidate end to end, inside the models its tasks come from and on their production execution path (continuous batching, chunked prefill, CUDA graphs, \texttt{torch.compile}, tensor parallelism).
This model runtime is the benchmark's measurement instrument, not a separate contribution, and we validate it against vLLM, SGLang, and the other production references before scoring any agent (Section~\ref{sec:eval-results}).
The result is an open leaderboard that ranks agents by what survives deployment, scored with \textsc{MacroEval}, and we seed it with a 6{,}900-agent-hour evaluation of five representative agents, from an RL-trained kernel model to contest-winning long-horizon harnesses.

\subsection*{Contributions}

\begin{enumerate}[leftmargin=*, itemsep=2pt, topsep=2pt]
\item \textbf{Production-aligned task suite.}
384 tasks from 47 architectures across 8 categories, from dense and MoE LLMs and SSMs to vision, audio, robotics, 3D, recommendation, and world models, with production module interfaces and speedups measured against production kernels.

\item \textbf{Compositional task hierarchy.}
Unlike flat benchmarks, the primitives-to-models hierarchy exposes \emph{composition effects}: 37\% of the agents' L3 winners break once their own lower-level kernels are active, and regenerating them on frozen lower-level winners yields 80 correct L3 winners instead of 45 (Section~\ref{sec:eval-agents}).

\item \textbf{Production-faithful evaluation.}
Kernel-level scoring on captured production shapes, with collectives on real tensor-parallel groups, and end-to-end scoring inside full models.

\item \textbf{A deployment-grounded leaderboard.}
\textsc{MacroEval} ranks agents by calibrated correctness, coverage, and end-to-end speedup across model families. Our seed evaluation of five representative agents shows kernel-level speedups of up to $6.6\times$ shrinking to at most $1.25\times$ end to end, and kernel-level scores that mis-rank agents (Section~\ref{sec:eval-e2e}; Appendix~\ref{app:leaderboard}).
\end{enumerate}

\section{Related Work}
\label{sec:related}

\paragraph{GPU kernel benchmarks.}
\textsc{KernelBench}~\citep{kernelbench} pioneered the evaluation of LLM-generated GPU kernels, drawing from an earlier generation of architectures (e.g.,~AlexNet, VGG). \textsc{robust-kbench}~\citep{robustkbench} addressed KernelBench's numerical instabilities and reward-hacking vulnerabilities. 
\textsc{SOL-ExecBench}~\citep{solexecbench} scores subgraphs against B200 speed-of-light bounds rather than software baselines.
\textsc{FlashInfer-Bench}~\citep{flashinferbench} integrates with an inference engine but is limited to FlashInfer LLM operators, while \textsc{CUDABench}~\citep{cudabench} and \textsc{TritonBench}~\citep{tritonbench} evaluate operators in isolation.
Further efforts~\citep{backendbench,multikernelbench,kernelcraft,nvidia2025computeeval,kalade2025npueval} broaden backend and robustness coverage while retaining an operator-level, single-GPU scope.
Concurrent benchmarks target multi-GPU code directly: \textsc{CommBench}~\citep{commbench} evaluates communication programs across NVLink and RDMA, and \textsc{ParallelKernelBench}~\citep{parallelkernelbench} asks agents to replace PyTorch+NCCL references with fused multi-GPU kernels; both score each program in isolation.
\fastkernels{} is complementary and fills four gaps: it is the first kernel benchmark to (i)~organize tasks into a compositional hierarchy from primitives to full models, mirroring production assembly, and to show that this hierarchy produces measurable composition effects (Section~\ref{sec:eval-agents}). It also (ii)~measures speedup against the kernels shipped in state-of-the-art inference frameworks rather than PyTorch or theoretical bounds, (iii)~matches production module interfaces, enabling copy--paste deployment, and (iv)~scores communication kernels inside the tensor-parallel models that issue them.

\paragraph{LLM-based kernel agents.}
A parallel line develops inference-time agents combining LLM reasoning with profiler feedback or evolutionary search~\citep{robustkbench,kernelagent,kernelevolve,wei2025astra,cudaforge,wang2025geak,alphaevolve,openevolve,kda,ako}, alongside specialized models trained via RL or fine-tuning~\citep{baronio2025kevin,li2025cudal1,kernelllm2025,li2025autotriton,zhu2026qimeng}. These systems are typically developed against operator-level benchmarks, which \fastkernels{} complements.
\section{Benchmark Design}
\label{sec:benchmark}

\fastkernels{} derives all tasks top-down from real model architectures, ensuring that every kernel can be traced to a specific layer in a specific model and evaluated end-to-end.

\subsection{Top-Down, Model-Driven Construction}
\label{sec:construction}

Unlike prior benchmarks that construct tasks bottom-up by combining primitive operators~\citep{kernelbench} or extracting subgraphs via automated pipelines~\citep{solexecbench}, \fastkernels{} takes a \emph{top-down} approach.
We begin with real model families---selected to represent the current and near-future frontier of AI workloads---and recursively decompose them into the kernels that constitute their inference paths.

\paragraph{Task construction.}
For each model architecture, we load the HuggingFace configuration and architecture definition, walk the forward pass, and produce standalone task implementations for each computational kernel with all configuration constants (hidden size, number of heads, data types) inlined from the model's configuration.
Every task is then audited to verify: (i)~semantic correctness of the reference implementation, (ii)~that the task captures the operator as it actually executes in the model (not a simplified proxy), (iii)~that tensor shapes and data types match the model's real configuration, and (iv)~that the task's interface matches the corresponding module in the production reference library (Section~\ref{sec:interface}).
The construction pipeline ships with \fastkernels{}, so users can add new architectures from HuggingFace with minimal effort.
No task is synthetic: unlike KernelBench's Level-2 random fusions or SOL-ExecBench's LLM-selected subgraphs, every task is an operation that runs in a real model during inference.

\subsection{Task Hierarchy}
\label{sec:hierarchy}

\fastkernels{} organizes tasks into four levels of increasing scope:

\begin{itemize}[leftmargin=*, itemsep=2pt, topsep=2pt]
\item \textbf{Level 1 --- Primitive operators.}
Individual operations: attention variants (GQA, MLA, sliding-window), normalizations (RMSNorm, LayerNorm), activations (SwiGLU, GeGLU), positional encodings (RoPE, ALiBi), and quantization/dequantization routines.

\item \textbf{Level 2 --- Fused operators.}
Multi-operation kernels representing natural fusion opportunities as they arise in real models: residual-add + RMSNorm + quantization, attention + output projection, MoE gate + dispatch + expert computation.

\item \textbf{Level 3 --- Full layers and blocks.}
Complete architectural blocks: transformer decoder layers, SSM scan blocks, MoE layers with routing, cross-attention layers, and vision encoder stages.

\item \textbf{Level 4 --- End-to-end model architectures.}
Full model inference paths evaluated as integrated systems.
The 47 representative architectures in Level~4 serve as the end-to-end evaluation set, but the underlying L1--L3 kernels cover most operators across ML models.
\end{itemize}

\paragraph{Standalone and composed evaluation.}
Crucially, the hierarchy is not just organizational---it is \emph{executable}.
L2--L4 baselines import their lower-level dependencies via relative imports (e.g., an L3 decoder layer does \texttt{from~..L2.attention~import~Attention}), mirroring how production frameworks assemble modules.
In \emph{standalone} mode (\texttt{-{}-standalone}), each candidate is benched in isolation: imported lower-level modules resolve to their production baselines regardless of what other candidate files exist.
In \emph{composed} mode, all candidate files in the candidate directory are active simultaneously, so an L3 block's lower-level imports resolve to agent-generated candidates rather than production baselines.
The standalone score measures how well the agent optimized one module; the composed score measures what a real deployment would observe when multiple agent-generated modules interact.
The gap between the two---the \emph{composition effect}---is a first-order signal that flat benchmarks are structurally unable to produce (Section~\ref{sec:eval-agents}).

\subsection{Architecture Coverage}
\label{sec:coverage}

\fastkernels{} is designed to cover as many model architectures as possible with the \emph{minimum} number of kernels, by consolidating near-identical operators (e.g., RoPE variants, normalizations, attention backends) into a single L1/L2 task and picking a small set of representative architectures whose union of operators subsumes the long tail of model families. The resulting 47 architectures span 8 categories and precision formats from 1.58-bit through FP32, and comprise 384 unique tasks: 108 L1, 155 L2, and 74 L3 kernel tasks, shared across architectures, plus the 47 end-to-end L4 models (Figure~\ref{fig:arch_pie}; counting rules and per-architecture details in Appendix~\ref{app:architectures}). 

To measure coverage, we implemented every architecture in HuggingFace Transformers (commit \texttt{89b6b17}) on \fastkernels{} without adding any operator: \textbf{94.6\% (472/499)} of them run and match native HF outputs within \texttt{dtype}-dependent tolerances, and only 5 require a genuinely new operation. The methodology and per-architecture results are in Appendix~\ref{app:hf_coverage}.

\begin{figure}[t]
\centering
\includegraphics[width=0.7\linewidth]{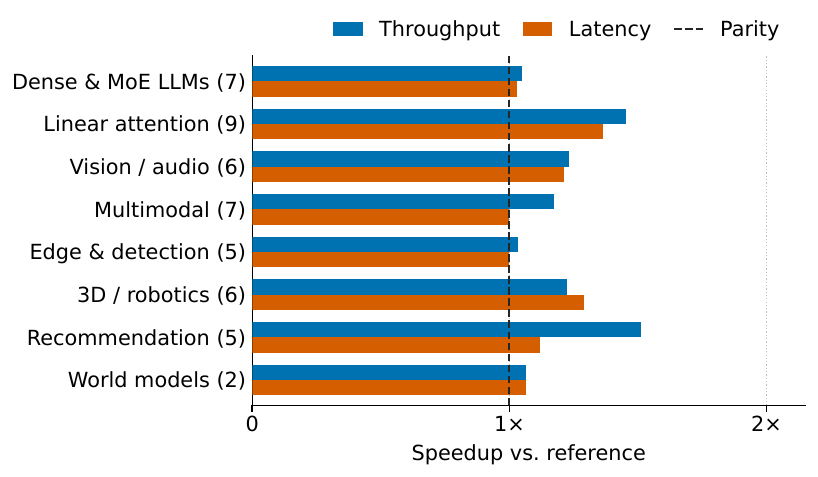}
\caption{\textbf{Composition and reference performance of \fastkernels{}}: 47 end-to-end architectures across 8 categories (count in parentheses).
Each bar is the geometric-mean end-to-end speedup of \fastkernels{} over the production reference across the category's architectures (throughput and latency).
Per-architecture details in Appendix Tables~\ref{tab:arch_detail}, \ref{tab:bench_detail}, and~\ref{tab:bench_detail_b}.}
\label{fig:arch_pie}
\label{fig:eval-e2e}
\end{figure}


\subsection{Interface-Compatible Design}
\label{sec:interface}

A central design principle of \fastkernels{} is that optimized kernels should be deployable into existing production systems with minimal effort.
For each model architecture family, we identify the corresponding state-of-the-art production library (e.g., vLLM for LLMs, SGLang for serving) and design each task's interface---its \texttt{\_\_init\_\_} constructor signature and \texttt{forward} method---to closely match the corresponding module in that reference library.

A kernel optimized within \fastkernels{} can therefore be copied into vLLM, SGLang, or another production framework without interface refactoring.
Unlike FlashInfer-Bench's FIApply, which substitutes at the kernel dispatch level (an abstraction internal to FlashInfer), \fastkernels{}'s compatibility operates at the \emph{module} level, the unit production frameworks compose.

\subsection{Multi-GPU Communication Kernels}
\label{sec:multigpu}

Production inference at scale is almost always distributed, and its collectives and communication--computation overlaps materially affect end-to-end latency.
\fastkernels{} includes the communication path of its tensor-parallel models as tasks: the custom IPC all-reduce (L1); the tensor-parallel linear, embedding, and MoE layers that issue it (L2); and a decoder layer that fuses it with residual addition and RMSNorm (L3).
The all-reduce and row-parallel linear are benched in a real NCCL group of the capture's TP degree, and all of these operators run end to end inside models served at TP 2--8.
Dedicated benchmarks~\citep{commbench,parallelkernelbench} cover more communication patterns but score each program in isolation; \fastkernels{} places every collective in the schedule it shares with the model's compute kernels, where a kernel $1.5\times$ faster on one GPU can still slow the model.

\section{Benchmarking Stack}
\label{sec:benchmarking_tooling}

\fastkernels{} exposes one evaluation stack through two tiers, \emph{kernel-level} (\texttt{fastkernels bench}) and \emph{end-to-end} (\texttt{fastkernels eval}), each usable as a diagnostic tool or in leaderboard mode.

\paragraph{Scenario sets.}
A scenario fixes a checkpoint, TP degree, dtype, and workloads from public datasets for its modality (e.g., WildChat for LLMs, Parti-prompts for diffusion, COCO for detection).
The \emph{full} set has one scenario per architecture (47 scenarios, 184 workloads, TP 1--8) and is used to validate the benchmark (Section~\ref{sec:eval-results}).
The \emph{default} set is the 11-model subset on which agents are scored, because a single long-horizon agent already spends up to 2{,}500 agent-hours on it: it spans all 8 categories, TP 1--4, and BF16/FP16/FP8/MXFP4 (48 workloads), and its models run 133 distinct L1--L3 tasks (Appendix~\ref{app:scenarios}).

\paragraph{Workload capture.}
\texttt{fastkernels capture} runs each scenario end to end and records, for every task module, a reconstructable constructor recipe and the shape, dtype, and stride of every \texttt{forward} argument.
Identical calls merge into one variant with a call count; the kernel tier then merges variants across workloads and benchmarks up to five per operator (smallest, median, and largest problem, then the most frequent): on the default B200 capture, 12.4M calls reduce to 87.7K distinct cases and 582 benchmarked ones (Appendix~\ref{app:capture}).

\paragraph{Kernel-level tier.}
The kernel tier rebuilds each candidate next to its baseline from the recipe, shares weights, and compares outputs and runtimes on the selected cases, one operator per isolated process with reward-hack checks.
Values are synthesized from the captured shapes, with offsets, indices, and cache slots kept consistent with the call; for data-dependent operators such as MoE dispatch, \fastkernels{} additionally replays golden inputs captured from real end-to-end executions, ensuring that both correctness and performance are measured on the actual tensors encountered by the model (Appendix~\ref{sec:eval-input-capture}).

\paragraph{End-to-end tier.}
The end-to-end tier swaps a candidate set into full model executions on the production path and real requests, next to a baseline run and a noise run that calibrates correctness thresholds, and produces the \textsc{MacroEval} metrics (Section~\ref{sec:metrics}).
The kernel tier gives agents fast feedback; the end-to-end tier is the ground truth that the leaderboard reports.
Both tiers integrate with profilers, NVIDIA Nsight Compute (NCU) for kernel-level analysis and Nsight Systems (NSYS) for end-to-end traces, with LLM-based extraction of the main bottlenecks, and with MLflow for tracking kernel lineage across agent runs.

\section{\textsc{MacroEval}: Cross-Architecture Metrics}
\label{sec:metrics}

\fastkernels{} evaluates candidate kernels by substituting them into end-to-end model executions and comparing each run with a reference run that uses the production baseline kernels.
The key challenge is that model families expose different correctness signals---tokens, embeddings, labels, rankings, trajectories, audio, or video---and different throughput--latency tradeoffs.
\textsc{MacroEval} addresses this by calibrating architecture-specific correctness to a common $[0,1]$ scale, measuring end-to-end speedup, and macro-averaging across families so no single architecture dominates the leaderboard.

\paragraph{Benchmark indexing.}
Let $f \in \mathcal{F}$ index architecture families, such as LLMs, diffusion models, vision models, speech and audio models, retrieval and recommender models, and robotics or world models.
For family $f$, let $i \in B_f$ index benchmark items, and let $r \in R_i$ index requests, prompts, seeds, or input instances for item $i$.
We write $y_{\mathrm{ref},i,r}$ and $y_{\mathrm{cand},i,r}$ for the reference and candidate outputs, and $T_{\mathrm{ref},i}$ and $T_{\mathrm{cand},i}$ for their measured runtimes.

\paragraph{Calibrated correctness.}
Raw output discrepancies are inherently architecture-dependent: teacher-forced top-1 disagreement with the reference tokens for autoregressive models (free-running greedy outputs diverge even between two correct runs, e.g.\ after a fresh compile), embedding or output distance for encoders and diffusion models, IoU discrepancy for detection or segmentation, WER/CER deltas for speech, ranking disagreement for retrieval, and trajectory error for robotics.
For each request, \fastkernels{} therefore computes a discrepancy $d_{i,r} = D_f(y_{\mathrm{cand},i,r}, y_{\mathrm{ref},i,r})$, with $D_f$ chosen for the output type of family $f$, and maps it to a calibrated correctness score in $[0,1]$:
\begin{equation}
C_{i,r} =
\begin{cases}
1, & d_{i,r} \le g_i, \\
\frac{f_i - d_{i,r}}{f_i - g_i}, & g_i < d_{i,r} < f_i, \\
0, & d_{i,r} \ge f_i.
\end{cases}
\end{equation}
Here $g_i$ is the threshold below which the candidate is indistinguishable from the reference, and $f_i$ is the threshold above which the output is incorrect or unusable.
Thresholds are fixed before evaluation. $g_i$ is calibrated per model from reference-vs-reference nondeterminism: a \emph{noise run} repeats the production reference in a fresh process with a cold compile cache (re-making every compilation and autotuning choice, as a candidate run does), and $g_i$ is the 95th percentile of its per-sample discrepancy against the reference (with a floor of $10^{-3}$; OpenFold3, whose noise run is bit-identical, is calibrated with a benign numerical perturbation instead, Appendix~\ref{app:e2e}).
We set $f_i = 0.5$ for every family---a candidate at least half as far from the reference as an unrelated output receives no credit---and $\tau_i = 0.9$; Appendix~\ref{app:e2e} shows that the ranking is unchanged when $g_i$ and $f_i$ are halved or doubled.

Correctness is averaged over requests, then over items within each family, then over families:
\begin{equation}
C_i = \frac{1}{|R_i|}\sum_{r \in R_i} C_{i,r},
\qquad
C_f = \frac{1}{|B_f|}\sum_{i \in B_f} C_i,
\qquad
C_{\mathrm{macro}} = \frac{1}{|\mathcal{F}|}\sum_{f \in \mathcal{F}} C_f.
\end{equation}
This macro-family weighting gives each architecture family equal influence and prevents overrepresented families, such as LLMs, from dominating the correctness axis.

\paragraph{Validity and coverage.}
Although calibrated correctness is continuous, deployment requires a hard validity decision for each benchmark item.
We define $\operatorname{valid}_i = \mathbf{1}(C_i \ge \tau_i)$ for a pre-specified item or family threshold $\tau_i$.
Items are invalid if they crash, hang, trigger shape or type errors, access memory illegally, produce NaNs, or fall below the correctness threshold.
We report both item-level coverage and macro-family coverage:
\begin{equation}
\mathrm{Coverage} = \frac{1}{N}\sum_i \operatorname{valid}_i,
\qquad
\mathrm{Coverage}_{\mathrm{macro}} =
\frac{1}{|\mathcal{F}|}\sum_{f \in \mathcal{F}}
\frac{1}{|B_f|}\sum_{i \in B_f}\operatorname{valid}_i.
\end{equation}
Macro-coverage is the preferred statistic when architecture balance matters.

\paragraph{Throughput--latency speedup.}
For valid items (invalid items receive no speedup credit), performance is measured against the production reference with a geometric blend of throughput and latency speedups:
\begin{equation}
s_i =
\left(s_i^{\mathrm{thr}}\right)^{\lambda}
\left(s_i^{\mathrm{lat}}\right)^{1-\lambda},
\qquad
s_i^{\mathrm{thr}} =
\frac{\mathrm{Throughput}_{\mathrm{cand},i}}
     {\mathrm{Throughput}_{\mathrm{ref},i}},
\qquad
s_i^{\mathrm{lat}} =
\frac{\mathrm{Latency}_{\mathrm{ref},i}}
     {\mathrm{Latency}_{\mathrm{cand},i}}.
\end{equation}
The default $\lambda = 0.5$ is neutral: an optimization that improves throughput while equivalently degrading latency receives no net credit, since such gains are workload-dependent tradeoffs rather than universal improvements; users may pre-specify a different $\lambda$.
Because speedups are multiplicative, \fastkernels{} aggregates the blended speedups with geometric means.
For each family, let $B_f^{\mathrm{valid}} = \{i \in B_f : \operatorname{valid}_i = 1\}$.
The family-level and macro-geometric speedups are
\begin{equation}
S_f =
\exp\left(
\frac{1}{|B_f^{\mathrm{valid}}|}
\sum_{i \in B_f^{\mathrm{valid}}}\log s_i
\right),
\qquad
S_{\mathrm{macro}} =
\exp\left(
\frac{1}{|\mathcal{F}_{\mathrm{valid}}|}
\sum_{f \in \mathcal{F}_{\mathrm{valid}}}\log S_f
\right),
\end{equation}
where $\mathcal{F}_{\mathrm{valid}}$ contains families with at least one valid item.
A family with no valid item is excluded from speedup aggregation, so incorrect kernels earn no performance credit, and the failure is penalized through macro-coverage instead.

\paragraph{Leaderboard.}
The leaderboard reports $S_{\mathrm{macro}}$, $C_{\mathrm{macro}}$, and $\mathrm{Coverage}_{\mathrm{macro}}$ separately, so users can rank agents by whichever their deployment prioritizes.
As the default single ranking, \fastkernels{} uses their product $\mathrm{Score}_{\mathrm{default}} = S_{\mathrm{macro}} \cdot C_{\mathrm{macro}} \cdot \mathrm{Coverage}_{\mathrm{macro}}$, which rewards agents that are simultaneously fast, correct, and broadly applicable, and it also reports throughput-only ($\lambda{=}1$) and latency-only ($\lambda{=}0$) rankings for batched and interactive deployments.
All speedups are measured empirically against the reference, without speed-of-light or roofline bounds.

\section{Evaluation}
\label{sec:evaluation}

The evaluation has two parts: Section~\ref{sec:eval-results} evaluates the \emph{benchmark itself}, checking that its baselines match each family's production framework so that speedups over them are meaningful; Sections~\ref{sec:eval-agents} and~\ref{sec:eval-e2e} evaluate \emph{kernel agents} on the default set, at the kernel level and end to end.

\subsection{Benchmark Validation: Baselines vs.\ Production Frameworks}
\label{sec:eval-results}

On the full set, we benchmark every architecture's end-to-end performance (throughput and latency on real workloads) and the correctness of each L1--L4 task against the state-of-the-art production framework for its family (e.g., vLLM for LLMs, SGLang for serving, diffusers for SDXL, FLA for linear-attention recurrences, timm for vision encoders).
Figure~\ref{fig:eval-e2e} reports the category-level geometric means: gains concentrate in linear attention ($1.45\times$ throughput) and recommendation ($1.51\times$), and the other categories sit near parity; Appendix~\ref{app:reference_perf} breaks the results down by architecture and GPU (Blackwell vs.\ Hopper).
References are pinned per release and GPU (vLLM~0.26.0 here); a daily CI job tracks their drift, 1--9\% for the median model over the four months before our pin (Appendix~\ref{app:drift}).

\subsection{Agent Evaluation: Kernel-Level Results}
\label{sec:eval-agents}

\paragraph{Agents.}
We evaluate Dr.~Kernel~\citep{drkernel}, an RL-trained model and one of the most recent KernelBench-ecosystem agents (others did not reproduce their results in our setting, e.g.\ KernelAgent~\citep{kernelagent}, or assume KernelBench's single-file tasks); KDA~\citep{kda} and AKO~\citep{ako} (adapted by the GEMM People entry), long-horizon harnesses behind winning MLSys 2026 FlashInfer contest entries~\citep{mlsys26contest}: KDA placed in all three full-agent tracks and GEMM People second on MoE; and Claude Code and Codex~\citep{codex}, the coding agents these harnesses build on.
Their campaigns total 6{,}900 agent-hours and 42B tokens, with Claude Code, KDA, and AKO spending a median of 6--12 hours per operator (Appendix~\ref{app:agents}).
We therefore evaluate fewer agents than single-turn leaderboards: the agents practitioners deploy iterate for hours per kernel (contest teams had 13 weeks for five kernels).
We omit evolutionary search, which multiplies this cost by population size and generation count and whose strongest systems~\citep{alphaevolve,kernelevolve} are not public.
Each agent optimizes every L1--L3 operator captured from the default set on B200 (47 L1, 64 L2, and 21 L3 targets) against the human-expert kernels that production frameworks ship.
L4 tasks are full models, so benchmarking them is the end-to-end evaluation of Section~\ref{sec:eval-e2e}.
We bench L1--L3 in three modes:
\emph{independent} (each operator generated and benched in isolation, over production lower-level kernels),
\emph{composed} (the independent winners---correct, geomean speedup $\geq 1\times$---activated together, so higher-level kernels run on the agent's own lower-level winners), and
\emph{sequential} (L2 and L3 regenerated on top of the frozen lower-level winners, benched composed).
Four agents reach geomean speedups of $1.6$--$6.6\times$ over the production kernels (Table~\ref{tab:kernel-levels}), but composition erodes them (Appendix Table~\ref{tab:kernel-composition}): once their children are also agent-generated, 26 of the agents' 71 L3 winners (37\%) fail correctness (13 of 18 for Codex), the surviving L3 winners of KDA and AKO lose 12--14\% of their speedup, and individual modules lose up to 76\% (Claude Code's \texttt{yolov10\_backbone}, $30.3\times \to 7.2\times$).
Sequential generation makes this an optimization target: it yields 80 correct L3 winners across agents instead of 45 composed, and the sequential L3 kernels of AKO and Codex are $1.48\times$ and $1.35\times$ faster than their composed independent ones.

\begin{table}[t]
  \centering
  \caption{\textbf{Kernel agents on \fastkernels{} (B200, default set).}
  \emph{Corr.}: operators correct on every benchmarked case; \emph{Geo.}: geomean speedup over correct operators; fast$_p$: correct and more than $p\times$ faster than production (Appendix~\ref{app:fastp}).
  Composed and sequential rows contain winners only.}
  \label{tab:kernel-levels}
  \small
  \setlength{\tabcolsep}{3pt}
  \resizebox{\linewidth}{!}{\begin{tabular}{@{}llrrrrrrrrrrrr@{}}
\toprule
 & & \multicolumn{4}{c}{\textbf{L1} (47 targets)} & \multicolumn{4}{c}{\textbf{L2} (64 targets)} & \multicolumn{4}{c}{\textbf{L3} (21 targets)} \\
\cmidrule(lr){3-6}\cmidrule(lr){7-10}\cmidrule(lr){11-14}
Agent & Mode & Corr. & Geo. & fast$_{1}$ & fast$_{2}$ & Corr. & Geo. & fast$_{1}$ & fast$_{2}$ & Corr. & Geo. & fast$_{1}$ & fast$_{2}$ \\
\midrule
\multirow{3}{*}{Dr.~Kernel} & Independent & 31 & 0.93$\times$ & 18 & 1 & 18 & 0.59$\times$ & 5 & 0 & 4 & 0.97$\times$ & 1 & 0 \\
 & Composed & 18 & 1.29$\times$ & 18 & 1 & 5 & 1.07$\times$ & 4 & 0 & 1 & 1.03$\times$ & 1 & 0 \\
 & Sequential & 18 & 1.29$\times$ & 18 & 1 & 5 & 1.40$\times$ & 5 & 0 & 2 & 1.16$\times$ & 2 & 0 \\
\midrule
\multirow{3}{*}{Claude Code} & Independent & 47 & 1.81$\times$ & 42 & 10 & 64 & 2.66$\times$ & 56 & 37 & 20 & 4.74$\times$ & 19 & 16 \\
 & Composed & 42 & 1.79$\times$ & 40 & 10 & 54 & 3.52$\times$ & 54 & 36 & 14 & 6.28$\times$ & 14 & 12 \\
 & Sequential & 42 & 1.76$\times$ & 40 & 10 & 57 & 3.21$\times$ & 57 & 37 & 20 & 4.50$\times$ & 20 & 15 \\
\midrule
\multirow{3}{*}{KDA} & Independent & 47 & 1.64$\times$ & 42 & 8 & 63 & 2.56$\times$ & 61 & 37 & 20 & 3.05$\times$ & 18 & 14 \\
 & Composed & 43 & 1.69$\times$ & 40 & 9 & 51 & 2.62$\times$ & 51 & 31 & 11 & 3.45$\times$ & 11 & 9 \\
 & Sequential & 43 & 1.68$\times$ & 40 & 8 & 57 & 2.30$\times$ & 57 & 30 & 20 & 2.86$\times$ & 20 & 14 \\
\midrule
\multirow{3}{*}{AKO} & Independent & 45 & 1.84$\times$ & 44 & 11 & 64 & 3.35$\times$ & 64 & 41 & 18 & 6.48$\times$ & 18 & 17 \\
 & Composed & 45 & 1.87$\times$ & 44 & 11 & 60 & 3.54$\times$ & 60 & 43 & 14 & 6.59$\times$ & 14 & 13 \\
 & Sequential & 45 & 1.86$\times$ & 44 & 11 & 59 & 3.51$\times$ & 59 & 40 & 20 & 6.55$\times$ & 20 & 17 \\
\midrule
\multirow{3}{*}{Codex} & Independent & 47 & 1.62$\times$ & 40 & 9 & 64 & 2.87$\times$ & 60 & 39 & 19 & 2.92$\times$ & 18 & 12 \\
 & Composed & 41 & 1.76$\times$ & 40 & 9 & 47 & 3.52$\times$ & 45 & 33 & 5 & 5.79$\times$ & 5 & 4 \\
 & Sequential & 41 & 1.75$\times$ & 38 & 9 & 56 & 2.90$\times$ & 56 & 34 & 18 & 3.95$\times$ & 18 & 11 \\
\bottomrule
\end{tabular}
}
\end{table}

\subsection{Agent Evaluation: End-to-End Leaderboard}
\label{sec:eval-e2e}

We deploy each agent's independent and sequential winner sets into the 11 default models (each receiving the kernels it instantiates) and run their real workloads on the production path, next to a baseline run and a noise run (Section~\ref{sec:metrics}).
If a set crashes the model or grossly corrupts its outputs, the offending kernel (named in the traceback, else found by bisection) is dropped and the model re-run; the first subset that runs correctly is the agent's \emph{deployable subset}, timed on the full workloads.
Table~\ref{tab:e2e-macroeval} scores both the full winner sets as-is (\emph{all winners}) and the deployable subsets (per model: Appendix~\ref{app:e2e}).

\begin{table}[t]
  \centering
  \caption{\textbf{End-to-end \textsc{MacroEval} leaderboard on the default set (B200).} \emph{Rank}: position of each agent--mode entry by Score (10 entries).
  \emph{Valid}: models with $C_i \geq \tau$; macro metrics as in Section~\ref{sec:metrics}; \emph{Thr.}/\emph{Lat.}: $\lambda{=}1$/$\lambda{=}0$.
  \emph{Deployable subset}: after dropping the kernels that crash or break the model; \emph{All winners}: every kernel-level winner as-is.}
  \label{tab:e2e-macroeval}
  \small
  \setlength{\tabcolsep}{3pt}
  \resizebox{\linewidth}{!}{
\begin{tabular}{@{}ll rrr rrr rr rr@{}}
\toprule
 & & \multicolumn{8}{c}{\textbf{Deployable subset}} & \multicolumn{2}{c}{\textbf{All winners}} \\
\cmidrule(lr){3-10}\cmidrule(lr){11-12}
 & & & & & \multicolumn{3}{c}{$S_{\mathrm{macro}}$} & & & & \\
\cmidrule(lr){6-8}
Agent & Mode & Valid & Cov$_{\mathrm{macro}}$ & $C_{\mathrm{macro}}$ & $\lambda{=}0.5$ & Thr. & Lat. & Score & Rank & Valid & Cov$_{\mathrm{macro}}$ \\
\midrule
\multirow{2}{*}{Dr.~Kernel} & Independent & 9/11 & 0.75 & 0.85 & 0.98$\times$ & 1.01$\times$ & 0.95$\times$ & 0.63 & 4 & 2/11 & 0.25 \\
 & Sequential & 9/11 & 0.75 & 0.75 & 0.99$\times$ & 1.02$\times$ & 0.96$\times$ & 0.55 & 5 & 2/11 & 0.25 \\
\midrule
\multirow{2}{*}{Claude Code} & Independent & 6/11 & 0.52 & 0.61 & 1.09$\times$ & 1.08$\times$ & 1.11$\times$ & 0.35 & 8 & 2/11 & 0.25 \\
 & Sequential & 7/11 & 0.56 & 0.55 & 1.11$\times$ & 1.09$\times$ & 1.14$\times$ & 0.34 & 9 & 1/11 & 0.12 \\
\midrule
\multirow{2}{*}{KDA} & Independent & 11/11 & 1.00 & 0.98 & 1.13$\times$ & 1.08$\times$ & 1.19$\times$ & \textbf{1.11} & \textbf{1} & 4/11 & 0.42 \\
 & Sequential & 11/11 & 1.00 & 0.98 & 1.04$\times$ & 1.03$\times$ & 1.05$\times$ & 1.03 & 2 & 6/11 & 0.58 \\
\midrule
\multirow{2}{*}{AKO} & Independent & 7/11 & 0.56 & 0.65 & 1.15$\times$ & 1.13$\times$ & 1.16$\times$ & 0.42 & 7 & 1/11 & 0.12 \\
 & Sequential & 9/11 & 0.83 & 0.82 & 1.25$\times$ & 1.08$\times$ & 1.44$\times$ & 0.85 & 3 & 2/11 & 0.25 \\
\midrule
\multirow{2}{*}{Codex} & Independent & 5/11 & 0.33 & 0.33 & 0.96$\times$ & 0.98$\times$ & 0.94$\times$ & 0.11 & 10 & 1/11 & 0.12 \\
 & Sequential & 8/11 & 0.71 & 0.71 & 1.10$\times$ & 1.11$\times$ & 1.08$\times$ & 0.55 & 6 & 1/11 & 0.12 \\
\bottomrule
\end{tabular}
}
\end{table}

\paragraph{Kernel-level winners rarely survive deployment as-is.}
Swapped in unchanged, only 22 of the 110 winner-set deployments run correctly (13--58\% of model families per agent).
Compilation-stack conflicts (e.g.\ embedding kernels calling raw C++ extensions that \texttt{torch.compile} cannot trace, rotary kernels that specialize dynamic batch shapes) and crashes on engine shapes the kernel bench never produced account for 86\% of the drops with an identified cause (Appendix Table~\ref{tab:e2e-drops}), all invisible to an isolated kernel check.
Other sets pass every kernel check yet corrupt the model (texture-only FLUX images, NaNs on Oasis), which only the end-to-end check flags.

\paragraph{Deployable end-to-end gains are small.}
Even after dropping the failing kernels, end-to-end speedups are modest ($S_{\mathrm{macro}}$ of $0.96$--$1.25\times$) despite kernel-level speedups of up to $6.6\times$: gains concentrate where the rewritten kernels dominate runtime (up to $2.2\times$ on YOLOv10, $1.7\times$ on OpenFold3, $1.5\times$ on GLA), while LLM serving stays within 6\% of the production kernels.
Sequential sets raise AKO and Codex from $1.15\times$ and $0.96\times$ to $1.25\times$ and $1.10\times$.

\paragraph{Kernel-level scores mis-rank agents.}
In isolation, Claude Code matches or beats KDA on correctness and speedup at every level (L3: 20 vs.\ 20 correct, $4.74\times$ vs.\ $3.05\times$), yet KDA's \textsc{MacroEval} score is $3\times$ higher ($1.11$ vs.\ $0.35$), the best of all agents even with halved or doubled thresholds (Appendix~\ref{app:e2e}).
The gap is deployability, not speed ($1.13\times$ vs.\ $1.09\times$): all 11 of KDA's models are valid versus 6 of Claude Code's. KDA wraps the same Claude model in a planning and review harness, which a kernel-level leaderboard cannot credit.

\section{Conclusion}
\label{sec:conclusion}

We presented \fastkernels{}, a benchmark that grounds kernel-generation evaluation in production inference.
Our seed evaluation shows that kernel-level success barely predicts deployment: speedups of up to $6.6\times$ shrink to at most $1.25\times$ end to end, only 20\% of winning kernel sets run correctly as-is, and kernel-level scores rank Claude Code above KDA, which scores $3\times$ higher.

\paragraph{Limitations.}
The $94.6\%$ HuggingFace coverage is checked on a few forward passes per architecture, some with randomly initialized weights, so it certifies functional equivalence rather than task accuracy.
Performance is measured on locked-clock B200 SXM; rankings on other GPUs may differ.
Our kernel-level results use synthesized inputs, so data-dependent kernels are scored faithfully only end to end; agents are scored on the default set only.
\textsc{MacroEval}'s family weighting, $\lambda{=}0.5$ blend, and product score encode value judgments, so we also report its components.

\subsection*{AI use statement}

LLM-based kernel-generation agents are a core experimental subject of this work: we evaluate Dr.~Kernel, Claude Code, KDA, AKO, and Codex as described in Section~\ref{sec:eval-agents} and Section~\ref{sec:eval-e2e}.
We used generative AI tools to edit software associated with the benchmark, to draft and refine portions of the manuscript, and to assist with table, figure, and reference formatting.
We have not used generative AI tools to generate synthetic datasets, formulate mathematical claims, or develop theoretical models; those tasks are not applicable to this work.
All AI-assisted text, code, figures, and experimental configurations were reviewed by the authors.
We take responsibility for the final content of this work, including text, claims, and artifacts produced with the aid of generative AI.

\ificlrfinal
  \ifpreprint
  \else
    \subsection*{Ethics statement}

This paper introduces a benchmark and evaluation methodology for GPU kernel generation.
It does not involve human subjects, crowdsourcing, or the release of a new high-risk generative model.
Tasks are derived from publicly available model architectures, libraries, and datasets.
The main dual-use consideration is that more efficient kernels can lower the cost of deploying large models; we mitigate silent failures by requiring production-faithful correctness checks before speedup credit is awarded.
The authors have read and adhere to the ICLR Code of Ethics.

  \fi
\else
  \subsection*{Ethics statement}

This paper introduces a benchmark and evaluation methodology for GPU kernel generation.
It does not involve human subjects, crowdsourcing, or the release of a new high-risk generative model.
Tasks are derived from publicly available model architectures, libraries, and datasets.
The main dual-use consideration is that more efficient kernels can lower the cost of deploying large models; we mitigate silent failures by requiring production-faithful correctness checks before speedup credit is awarded.
The authors have read and adhere to the ICLR Code of Ethics.

\fi
\subsection*{Reproducibility statement}

\ificlrfinal
The implementation of \fastkernels{} and the associated evaluation scripts are open-sourced at \url{https://github.com/Snowflake-AI-Research/fastkernels}.
\else
An anonymized implementation of \fastkernels{} and the associated evaluation scripts is included in the supplementary material and will be open-sourced upon publication.
\fi
Section~\ref{sec:benchmark} describes task construction, the compositional hierarchy, and production-matched interfaces;
Section~\ref{sec:benchmarking_tooling} describes the evaluation stack;
and Section~\ref{sec:metrics} defines \textsc{MacroEval}.
Scenario sets, the capture and case-selection pipeline, and agent configurations and costs are documented in Appendices~\ref{app:scenarios}, \ref{app:capture}, and~\ref{app:agents}; hardware, serving workloads, threshold calibration, sensitivity analyses, and per-architecture coverage in Appendix~\ref{app:experimental}, Appendix~\ref{app:e2e}, and Appendix~\ref{app:reference_perf}.

\ificlrfinal
\subsubsection*{Acknowledgments}
\label{sec:acknowledgements}

We thank Aurick Qiao for his valuable comments and suggestions.
\fi

\bibliography{references}
\bibliographystyle{iclr2027_conference}

\newpage
\appendix
\raggedbottom

\section{Benchmark Composition}
\label{app:architectures}

Table~\ref{tab:arch_detail} reports the per-architecture composition of \fastkernels{}: the reference HuggingFace checkpoint, default inference dtype, and the number of L1 (primitive), L2 (fused/composite), L3 (layer/block) and L4 (end-to-end) tasks each architecture uses.

\paragraph{Counting rule.}
A task is one module file under \texttt{tasks/baseline/L$n$/} that defines at least one benchmarkable \texttt{nn.Module} and is reachable, through the import graph, from at least one of the 47 L4 architectures; it is counted once even when it ships a CUDA sidecar (17 tasks do) or defines several module classes.
Twelve function-only helper kernels (e.g., Triton attention-merge and causal-convolution kernels) are called from inside other tasks and are optimized through them, so they are not counted separately.
This yields 108 L1, 155 L2, and 74 L3 tasks plus 47 L4 models, 384 in total.
L4 counts one task per end-to-end architecture: the encoder sub-models that some pipelines import (the CLIP and T5 text encoders, the Qwen2.5-VL vision encoder) are part of those pipelines' L4 tasks, and the L1--L3 set is identical with or without DeepSeek-V3.2, whose modules GLM-5.2 builds on.
Because tasks are shared, the per-architecture columns of Table~\ref{tab:arch_detail} sum to more than these totals.

\begin{table}[H]
  \centering
  \caption{Full architecture composition of \fastkernels{}. L1--L4 report the number of primitive, fused/composite, layer/block-level, and end-to-end tasks each architecture uses, under the counting rule above (L4 is always 1 by construction). \emph{Reference} is the HuggingFace checkpoint or upstream repository the L4 pipeline tracks; \emph{dtype} is the default inference dtype.}
  \label{tab:arch_detail}
  \scriptsize
  \setlength{\tabcolsep}{3pt}
  \begin{tabular}{@{}ll>{\raggedright\arraybackslash}p{4.5cm}lcccc@{}}
  \toprule
  \textbf{Category} & \textbf{Architecture} & \textbf{Reference} & \textbf{dtype} & \textbf{L1} & \textbf{L2} & \textbf{L3} & \textbf{L4} \\
  \midrule
  \multirow{7}{*}{Dense \& MoE LLMs}
   & Llama-3.1          & \texttt{meta-llama/\allowbreak Llama-3.1-8B-Instruct}    & BF16 & 13 & 6 & 1 & 1 \\
   & GLM-5.2            & \texttt{nvidia/\allowbreak GLM-5.2-NVFP4}                & NVFP4 & 31 & 11 & 1 & 1 \\
   & Mixtral            & \texttt{mistralai/\allowbreak Mixtral-8x7B-Instruct-v0.1} & BF16 & 19 & 8 & 1 & 1 \\
   & BitNet 1.58b       & \texttt{microsoft/\allowbreak bitnet-b1.58-2B-4T}        & W1.58/BF16 & 13 & 5 & 1 & 1 \\
   & GPT-OSS (MXFP4)    & \texttt{openai/\allowbreak gpt-oss-120b}                 & MXFP4 & 13 & 8 & 1 & 1 \\
   & EAGLE-3            & \texttt{jamesliu1/\allowbreak sglang-EAGLE3-Llama-3.1-Instruct-8B} & BF16 & 13 & 4 & 0 & 1 \\
   & Gemma-4            & \texttt{google/\allowbreak gemma-4-26B-A4B-it}           & BF16 & 20 & 8 & 1 & 1 \\
  \midrule
  \multirow{9}{*}{\makecell[l]{Linear Attention\\\& New Archs}}
   & Mamba              & \texttt{state-spaces/\allowbreak mamba-2.8b-hf}          & BF16 & 7  & 3 & 1 & 1 \\
   & Mamba2             & \texttt{mistralai/\allowbreak Mamba-Codestral-7B-v0.1}   & BF16 & 9  & 3 & 1 & 1 \\
   & RWKV-7             & \texttt{fla-hub/\allowbreak rwkv7-2.9B-g1}               & BF16 & 11 & 2 & 1 & 1 \\
   & GLA                & \texttt{fla-hub/\allowbreak gla-2.7B-100B}               & BF16 & 11 & 2 & 1 & 1 \\
   & RetNet             & \texttt{fla-hub/\allowbreak retnet-2.7B-100B}            & BF16 & 11 & 2 & 1 & 1 \\
   & Qwen-3-Next        & \texttt{Qwen/\allowbreak Qwen3-Next-80B-A3B-Instruct}    & BF16 & 22 & 7 & 1 & 1 \\
   & Kimi-Linear        & \texttt{moonshotai/\allowbreak Kimi-Linear-48B-A3B-Instruct} & BF16 & 25 & 9 & 1 & 1 \\
   & TTT-E2E            & \texttt{test-time-training/e2e}                          & BF16 & 8  & 3 & 1 & 1 \\
   & Jamba              & \texttt{ai21labs/\allowbreak AI21-Jamba-Mini-1.7}        & BF16 & 20 & 8 & 1 & 1 \\
  \midrule
  \multirow{6}{*}{\makecell[l]{Vision / Video /\\Audio}}
   & FLUX.1-Dev         & \texttt{black-forest-labs/\allowbreak FLUX.1-dev}        & BF16 & 16 & 10 & 3 & 1 \\
   & HunyuanVideo-1.5   & \texttt{hunyuanvideo-community/\allowbreak HunyuanVideo-1.5} & BF16 & 24 & 15 & 4 & 1 \\
   & SDXL               & \texttt{stabilityai/\allowbreak stable-diffusion-xl-base-1.0} & FP16 & 8  & 8  & 3 & 1 \\
   & SAM3.1             & \texttt{facebook/\allowbreak sam3.1}                     & BF16 & 8  & 6  & 8 & 1 \\
   & Whisper            & \texttt{openai/\allowbreak whisper-large-v3}             & FP16 & 13 & 6  & 2 & 1 \\
   & CosyVoice3         & \texttt{FunAudioLLM/\allowbreak Fun-CosyVoice3-0.5B-2512} & BF16 & 14 & 9  & 2 & 1 \\
  \midrule
  \multirow{7}{*}{\makecell[l]{Multimodal \&\\Encoders}}
   & Qwen2-VL           & \texttt{Qwen/\allowbreak Qwen2-VL-7B-Instruct}           & BF16 & 20 & 10 & 2 & 1 \\
   & Qwen3-VL-8B        & \texttt{Qwen/\allowbreak Qwen3-VL-8B-Instruct}           & BF16 & 27 & 13 & 3 & 1 \\
   & Qwen3-VL-235B      & \texttt{Qwen/\allowbreak Qwen3-VL-235B-A22B-Instruct-FP8} & FP8  & 27 & 13 & 3 & 1 \\
   & Qwen-2.5-Omni      & \texttt{Qwen/\allowbreak Qwen2.5-Omni-7B}                & BF16 & 20 & 10 & 2 & 1 \\
   & SigLIP-2           & \texttt{google/\allowbreak siglip2-so400m-patch16-naflex} & BF16 & 3  & 3  & 1 & 1 \\
   & DINOv3             & \texttt{facebook/\allowbreak dinov3-vit7b16-pretrain-lvd1689m} & BF16 & 4  & 2  & 1 & 1 \\
   & SwinV2             & \texttt{timm/\allowbreak swinv2\_large\_window12\_192.ms\_in22k} & FP32 & 4 & 3 & 2 & 1 \\
  \midrule
  \multirow{5}{*}{\makecell[l]{Edge \&\\Detection}}
   & MobileNetV4        & \texttt{timm/\allowbreak mobilenetv4\_conv\_medium.\allowbreak e500\_r256\_in1k} & FP32 & 5  & 2  & 1 & 1 \\
   & ConvNeXtV2         & \texttt{facebook/\allowbreak convnextv2-base-22k-384}    & FP32 & 7  & 1  & 1 & 1 \\
   & EfficientNetV2     & \texttt{timm/\allowbreak efficientnetv2\_rw\_m.agc\_in1k} & FP32 & 6  & 3  & 1 & 1 \\
   & YOLOv10            & \texttt{jameslahm/\allowbreak yolov10n}                  & FP16 & 8  & 11 & 3 & 1 \\
   & RTDetrV2           & \texttt{PekingU/\allowbreak rtdetr\_v2\_r101vd}          & FP32 & 16 & 8  & 4 & 1 \\
  \midrule
  \multirow{6}{*}{\makecell[l]{3D / Robotics /\\Science}}
   & 3DGS               & \texttt{nerfstudioteam/\allowbreak datasets} (mip-NeRF~360) & FP32 & 3  & 1  & 0 & 1 \\
   & InstantNGP         & \texttt{nerfstudioteam/\allowbreak datasets} (fox)       & FP16 & 0  & 0  & 1 & 1 \\
   & PointTransformerV3 & \texttt{Pointcept/\allowbreak PointTransformerV3}        & FP32 & 3  & 1  & 0 & 1 \\
   & OpenFold3          & \texttt{OpenFold/\allowbreak OpenFold3}                  & BF16 & 7  & 14 & 5 & 1 \\
   & Pi0                & \texttt{openpi/\allowbreak pi0\_aloha\_pen\_uncap}       & BF16 & 13 & 5  & 2 & 1 \\
   & DP3                & \texttt{3d-diffusion-policy.\allowbreak github.io}       & FP32 & 8  & 3  & 1 & 1 \\
  \midrule
  \multirow{5}{*}{\makecell[l]{Recommendation\\\& Specialized}}
   & DLRMv2             & criteo (1\,TB click logs)                                 & FP32 & 3  & 3  & 0 & 1 \\
   & LightGCN           & MovieLens / Gowalla / Amazon-Book                         & FP32 & 2  & 1  & 0 & 1 \\
   & BGE-M3             & \texttt{BAAI/\allowbreak bge-m3}                          & FP16 & 8  & 3  & 3 & 1 \\
   & ColBERTv2          & \texttt{colbert-ir/\allowbreak colbertv2.0}               & FP16 & 7  & 4  & 3 & 1 \\
   & LLaDA              & \texttt{GSAI-ML/\allowbreak LLaDA-8B-Instruct}            & BF16 & 6  & 3  & 1 & 1 \\
  \midrule
  \multirow{2}{*}{World Models}
   & Oasis              & \texttt{Etched/\allowbreak oasis-500m}                   & BF16 & 7  & 7  & 5 & 1 \\
   & V-JEPA 2           & \texttt{facebook/\allowbreak vjepa2-vitl-fpc64-256}      & BF16 & 4  & 3  & 3 & 1 \\
  \bottomrule
  \end{tabular}

  \vspace{4pt}
  {\small \textit{Note:} A value of 0 indicates that the current implementation does not expose a benchmark task at that level. Quantized configurations reuse the same high-level task graph as their underlying architecture while activating distinct quantized kernels and execution paths.}
\end{table}

\section{Interface Compatibility Examples}
\label{app:interface}

To illustrate the interface-compatible design, we show side-by-side comparisons of \fastkernels{} module interfaces with their corresponding vLLM modules.

  \medskip
  \noindent\textbf{Example: Mamba-2 mixer.}

  \noindent
  \begin{minipage}[t]{0.48\linewidth}
  \vspace{0pt}
  \textbf{\fastkernels{} interface:}
  \begin{lstlisting}
  class Mamba2Mixer(nn.Module):
    def __init__(self,
        hidden_size,
        ssm_state_size,
        conv_kernel_size,
        intermediate_size,
        use_conv_bias,
        use_bias,
        n_groups=1,
        num_heads=128,
        head_dim=64,
        rms_norm_eps=1e-5,
        activation="silu",
        use_rms_norm=True,
        ...):
      ...

    def forward(self, hidden_states):
      ...
  \end{lstlisting}
  \end{minipage}
  \hfill
  \begin{minipage}[t]{0.48\linewidth}
  \vspace{0pt}
  \textbf{vLLM interface:}
  \begin{lstlisting}
  class MambaMixer2(nn.Module):
    def __init__(self,
        hidden_size,
        ssm_state_size,
        conv_kernel_size,
        intermediate_size,
        use_conv_bias,
        use_bias,
        n_groups=1,
        num_heads=128,
        head_dim=64,
        rms_norm_eps=1e-5,
        activation="silu",
        use_rms_norm=True,
        ...):
      ...

    def forward(self, hidden_states):
      ...
  \end{lstlisting}
  \end{minipage}

  \medskip

  \noindent\textbf{Example: Mamba-2 gated RMSNorm.}

  \noindent
  \begin{minipage}[t]{0.48\linewidth}
  \vspace{0pt}
  \textbf{\fastkernels{} interface:}
  \begin{lstlisting}
  class Mixer2RMSNormGated(
      nn.Module):
    def __init__(self,
                 full_hidden_size,
                 full_n_groups,
                 use_rms_norm=True,
                 eps=1e-6):
      ...

    def forward(self, x, gate):
      ...
  \end{lstlisting}
  \end{minipage}
  \hfill
  \begin{minipage}[t]{0.48\linewidth}
  \vspace{0pt}
  \textbf{vLLM interface:}
  \begin{lstlisting}
  class Mixer2RMSNormGated(
      nn.Module):
    def __init__(self,
                 full_hidden_size,
                 full_n_groups,
                 use_rms_norm=True,
                 eps=1e-6):
      ...

    def forward(self, x, gate):
      ...
  \end{lstlisting}
  \end{minipage}

  \medskip

  The shared \texttt{\_\_init\_\_} and \texttt{forward} arguments make substitution a file swap: call sites need not change, and any residual differences are optional framework kwargs rather than a different compute interface.

\section{Scenario Sets}
\label{app:scenarios}

Scenario sets are YAML tables shipped with \fastkernels{}.
Each row fixes a checkpoint, tensor-parallel degree, dtype, and the workloads to run; per-workload datasets, request counts, and sequence lengths are defined once in a workload registry, so every tier and every run of a set sees identical inputs.
Four sets ship with the benchmark:
\begin{itemize}[leftmargin=*, itemsep=1pt, topsep=2pt]
\item \textbf{full}: one scenario per architecture (47 scenarios, 184 workloads, TP 1--8, BF16/FP16/FP32/FP8/MXFP4/NVFP4), and therefore every task. It is the set on which we validate the baselines against production references (Section~\ref{sec:eval-results}, Appendix~\ref{app:reference_perf}).
\item \textbf{default}: the 11 scenarios of Table~\ref{tab:default_scenarios} (48 workloads), on which all agent results are reported. It contains at least one model per category, which \textsc{MacroEval}'s per-family averaging requires, spans TP 1, 2, and 4 and BF16, FP16, FP8, and MXFP4 execution, and pairs throughput workloads (batched request mixes) with latency workloads (single requests or small fixed batches) for every model. Its B200 capture exercises 133 L1--L3 tasks (47 L1, 64 L2, 22 L3), which are the agents' targets; \texttt{oasis\_rollout}, a whole-model rollout driver rather than a kernel, is excluded from scoring, leaving 21 L3 targets.
\item \textbf{minimal}: Llama-3.1-8B with two workloads, a smoke test for the capture-to-bench pipeline.
\item \textbf{vllm\_only}: the 16 architectures whose reference is vLLM, used to compare vLLM releases for drift tracking (Appendix~\ref{app:drift}).
\end{itemize}
Agents are scored on the default set rather than the full set because of cost: the long-horizon agents of Section~\ref{sec:eval-agents} spend 1{,}763--2{,}512 agent-hours each on the default set's 133 operators (Appendix~\ref{app:agents}), and the full set's architectures use more than twice as many L1--L3 tasks (337 versus 159 under the counting rule of Appendix~\ref{app:architectures}, of which the default workloads invoke 133 on B200).

\begin{table}[H]
  \centering
  \caption{\textbf{The default scenario set.} Workloads are listed by name; their datasets and sizes are defined in the workload registry (Appendix~\ref{app:experimental}).}
  \label{tab:default_scenarios}
  \scriptsize
  \setlength{\tabcolsep}{3pt}
  \begin{tabular}{@{}llcl>{\raggedright\arraybackslash}p{5.6cm}@{}}
  \toprule
  \textbf{Model} & \textbf{Category} & \textbf{TP} & \textbf{dtype} & \textbf{Workloads} \\
  \midrule
  Llama-3.1-8B-Instruct & Dense \& MoE LLMs & 1 & BF16 & mixed, long-context, single request, fixed batch 32 \\
  gpt-oss-120b & Dense \& MoE LLMs & 2 & MXFP4 & mixed, long-context, single request, fixed batch 32 \\
  Qwen3-Next-80B-A3B-Instruct & Linear Attention & 2 & BF16 & mixed, long-context, single request, fixed batch 32 \\
  Kimi-Linear-48B-A3B-Instruct & Linear Attention & 2 & BF16 & mixed, long-context, single request, fixed batch 32 \\
  GLA-2.7B & Linear Attention & 1 & BF16 & mixed, long-context, single request, fixed batch 32 \\
  FLUX.1-dev & Vision / Video / Audio & 1 & BF16 & 512 and 1024 px, batched and single image \\
  Qwen3-VL-235B-A22B-Instruct-FP8 & Multimodal & 4 & FP8 & text only, image, video, single image, single video \\
  YOLOv10n & Edge \& Detection & 1 & FP16 & COCO val, single image, batch 4 \\
  OpenFold3 & 3D / Robotics / Science & 1 & BF16 & four length buckets, batched and single \\
  BGE-M3 & Recommendation & 1 & FP16 & MLDR documents, single request, fixed batch 32 \\
  Oasis-500M & World Models & 1 & FP16 & four rollout / denoising settings, batch-1 latency \\
  \bottomrule
  \end{tabular}
\end{table}

\section{Workload Capture and Case Selection}
\label{app:capture}

\paragraph{Instrumentation.}
\texttt{fastkernels capture <set>} loads each scenario into its own engine process (a TP-$N$ scenario claims $N$ GPUs; scenarios are packed onto the GPU pool and isolated so that a crash in one does not affect the others) and runs every listed workload on its real inputs.
Before the model is built, it wraps the constructor and \texttt{forward} of every \texttt{nn.Module} class under \texttt{tasks/baseline}.
A constructor call is recorded as a reconstructable \emph{recipe}: configuration dataclasses and enums with their field values, dtypes, tensor shapes, and references to other captured operators (e.g., the linear layers inside an attention block); runtime handles such as process groups are marked opaque.
A forward call is recorded as the shape, dtype, and, for non-contiguous tensors, stride of every tensor argument, together with the values of scalar arguments.
Shape capture stores no weights or activation values; golden inputs for data-dependent operators are recorded separately (below).

\paragraph{Deduplication.}
Each call is keyed by the canonical serialization of its record, and identical keys are merged into one \emph{variant} that keeps a call count.
Each forward variant also records which constructor variants built the instances it ran on, so the benchmark can rebuild exactly the module configuration that produced each shape.
Since every layer of a model and every decode step with the same batch composition produce the same key, the 12.4M L1--L3 forward calls of the default B200 capture reduce to 93.3K variants across its 48 per-workload reports, and to 87.7K distinct (constructor, forward) cases once merged across reports.
The distribution is long-tailed: half of the operators have at most seven distinct cases, while operators whose shapes follow the per-step token count (decode attention, RMSNorm, linear layers, embeddings) reach thousands.
For the few operators that read the global inference context inside \texttt{forward} (MLA attention), a coarse context summary (prefill or decode, and a power-of-two bucket of the number of prefill sequences) is folded into the key, which is enough to rebuild the code path without defeating deduplication.

\paragraph{Verification.}
Two independent checks run after every workload.
First, a forward pre-hook, registered separately from the wrappers, re-observes every operator call; its per-operator keys and counts must match the wrapper records.
Second, a mock continuous-batching and chunked-prefill scheduler, driven only by each request's real prompt and generated lengths, the engine's token and sequence budgets, and its KV-block pool, replays the serving schedule and predicts the per-step batch composition; the captured leading dimensions (token counts, sequence counts, batch sizes) must agree with it.
The replay does not model speculative decoding, so only the first check applies to EAGLE-3.

\paragraph{Case selection.}
\texttt{fastkernels bench} loads every report of a capture and deduplicates cases across reports by (constructor recipe, forward record).
When an operator has more than five distinct cases (the default budget), it keeps the smallest, median, and largest by total input elements and fills the remaining slots with the most frequently called cases, so the selection covers the size range rather than only the hot decode shape.
On the default B200 capture this yields 582 benchmarked cases over the 154 operator classes defined in the 133 target tasks.

\paragraph{Input materialization.}
Floating-point inputs are drawn from a standard normal (FP8 inputs are clamped normals cast to FP8) and masks are all ones.
Integer tensors are made valid for the call: cumulative sequence-length vectors are rebuilt as partitions consistent with the query and key token counts, token and position indices are bounded by the vocabulary size and maximum position, expert indices by the expert count, and slot mappings and block tables by the cache they index.
Captured non-contiguous layouts are reproduced with the recorded strides.
Baseline and candidate share randomly initialized weights; quantized weights (block-scaled FP8, NVFP4, MXFP4) are produced with the model's own quantization and weight-preparation routines, so kernels see the layouts they would see after checkpoint loading.
For data-dependent operators such as MoE dispatch, whose control flow depends on input values (Appendix~\ref{sec:eval-input-capture}), the kernel tier can instead replay golden inputs captured from real end-to-end executions, so that correctness and performance are measured on the tensors the model actually produces.
Cases whose inputs the baseline rejects, or whose baseline output is all zeros (so that a stub would pass), are skipped.
The all-reduce and row-parallel linear run in an NCCL group spawned with world size equal to the capture's TP degree, with every rank executing the same seeded cases.

\paragraph{Correctness and timing.}
Each case runs three rounds of fresh inputs, and an output passes when at least 99\% of its elements are within per-dtype tolerances ($\mathrm{atol},\mathrm{rtol}$ of $10^{-5},10^{-3}$ for FP32; $10^{-2},10^{-2}$ for FP16 and BF16; $0.125,0.125$ for FP8).
Timing uses CUDA events with 10 warm-up and 50 timed iterations, flushing the L2 cache and rotating input buffers between iterations, and reports the median.
Each operator runs in its own process, where the harness checks the identity of the timing primitive and of its own critical functions, background-thread counts, and output types, and reports any tampering as a reward hack.

\section{Case Study: Captured Inputs Matter for Data-Dependent Ops}
\label{sec:eval-input-capture}

\begin{figure}[t]
\centering
\includegraphics[width=0.7\textwidth]{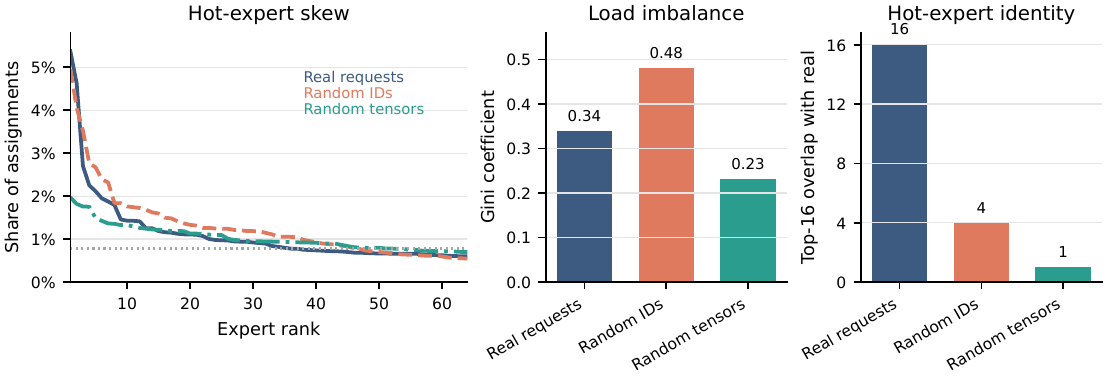}
\caption{\textbf{Input capture changes data-dependent execution.} First MoE layer of Qwen3-VL-30B-A3B-Instruct-FP8 (128 experts, top-8 routing) under three inputs: real WildChat requests, random token IDs (matched lengths), and random tensors injected at the gate. Random tensors look near-uniform; random tokens skew differently from real requests, with only 4 of the top-16 hot experts in common.}
\label{fig:moe-input-capture}
\end{figure}

Many production operators have data-dependent control flow---which experts, sparse blocks, cache slots, or communication paths execute is determined by the input.
We isolate this effect on a well-known case: the first MoE gate of Qwen3-VL-30B-A3B-Instruct-FP8 (Figure~\ref{fig:moe-input-capture}). Random tensors yield a near-uniform gate (Gini $0.232$); random token IDs over-concentrate load and identify different hot experts from real WildChat requests (Gini $0.480$ vs.\ $0.339$; only $4 / 16$ shared hot experts).
\fastkernels{}'s capture-and-replay path therefore rewards data-dependent kernels on the tensors production requests actually create, rather than on synthetic control flow: the kernel tier can replay golden inputs captured from real end-to-end executions (Appendix~\ref{app:capture}), and the end-to-end tier runs real requests.

\section{Experimental Details}
\label{app:experimental}

\paragraph{Hardware.}
End-to-end L4 validate results in Section~\ref{sec:eval-results} and Tables~\ref{tab:bench_detail}--\ref{tab:bench_detail_b} are measured on a NVIDIA DGX B200 box with 8x NVIDIA B200 SXM GPUs, 180 GB HBM3e each (1.44 TB total), interconnected via 5th-gen NVLink/NVSwitch.
Single-GPU experiments use one GPU; multi-GPU experiments use up to $8\times$B200 (tensor-parallel degree matching the production reference: e.g. TP=2 Mixtral/GPT-OSS, TP=8 GLM-5.2).
L1--L3 agent captures in Section~\ref{sec:eval-agents} use the same B200 nodes.
The Hopper results in Appendix~\ref{app:reference_perf_h200} repeat the same L4 scenario table on a node with 8x NVIDIA H200 SXM GPUs, 141 GB HBM3e each, using the same tensor-parallel degrees; GLM-5.2 is not run there because its NVFP4 kernels require Blackwell.

\paragraph{Serving benchmark.}
Autoregressive L4 throughput uses a mixed-length request mix and a long-context mix against the production server (vLLM, SGLang, or FLA); latency uses single-request (batch 1) and fixed-batch-32 decoding.
Each latency measurement is the median of 5 iterations.
Non-AR models use family-specific public workloads (Parti-prompts, SEED-TTS-Eval, LibriSpeech, COCO val, Food101, MS MARCO, ALOHA, etc.) as listed in Tables~\ref{tab:bench_detail} and \ref{tab:bench_detail_b}.

\clearpage
\section{Reference Performance and Correctness Details}
\label{app:reference_perf}

\begin{figure}[H]
  \centering
  \includegraphics[width=\linewidth]{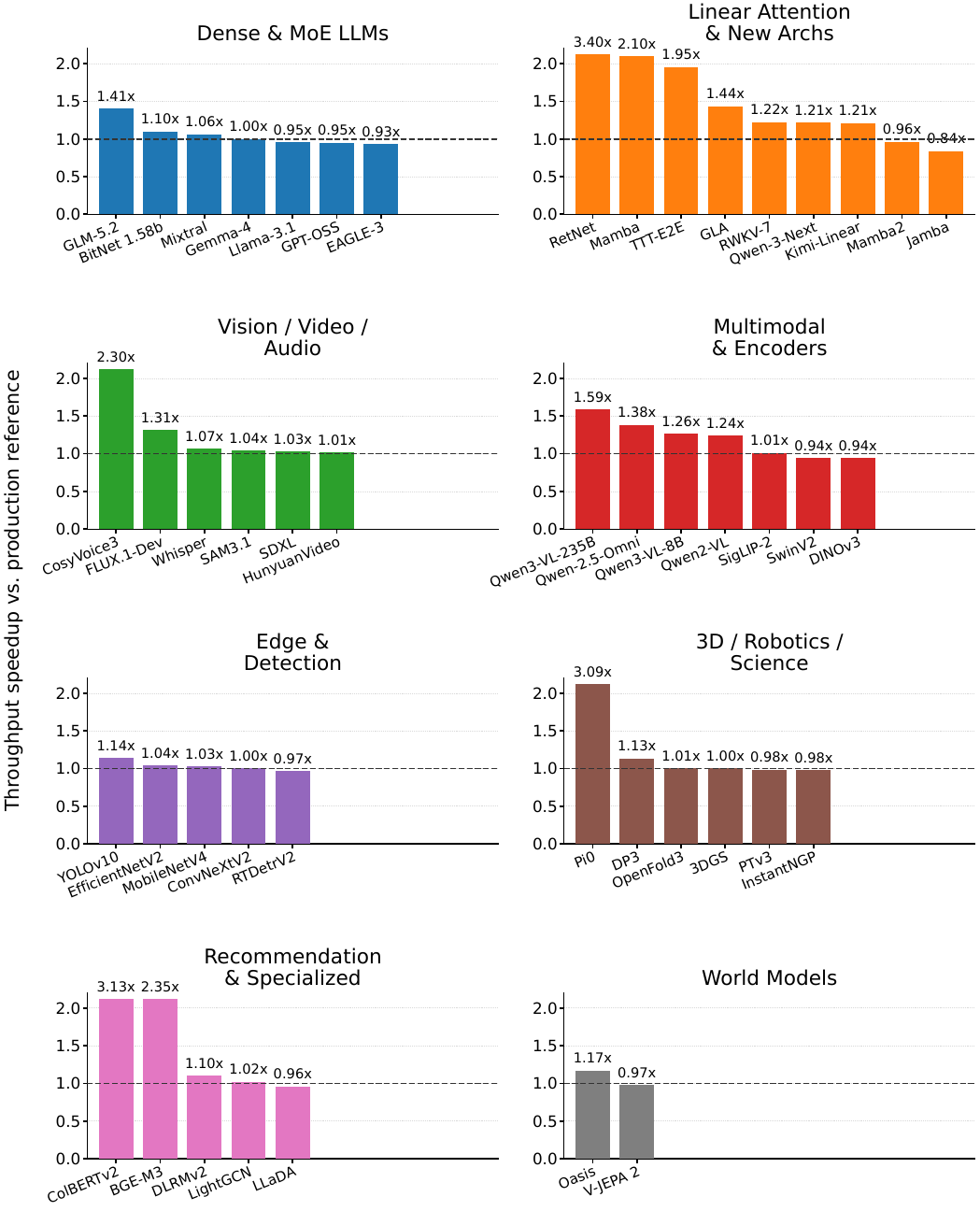}
  \caption{\textbf{Per-architecture throughput speedup on B200}, grouped by category. Bars taller than the axis are clipped and labeled with their value; the dashed line marks parity.}
  \label{fig:eval-thr-breakdown}
\end{figure}

\begin{figure}[H]
  \centering
  \includegraphics[width=\linewidth]{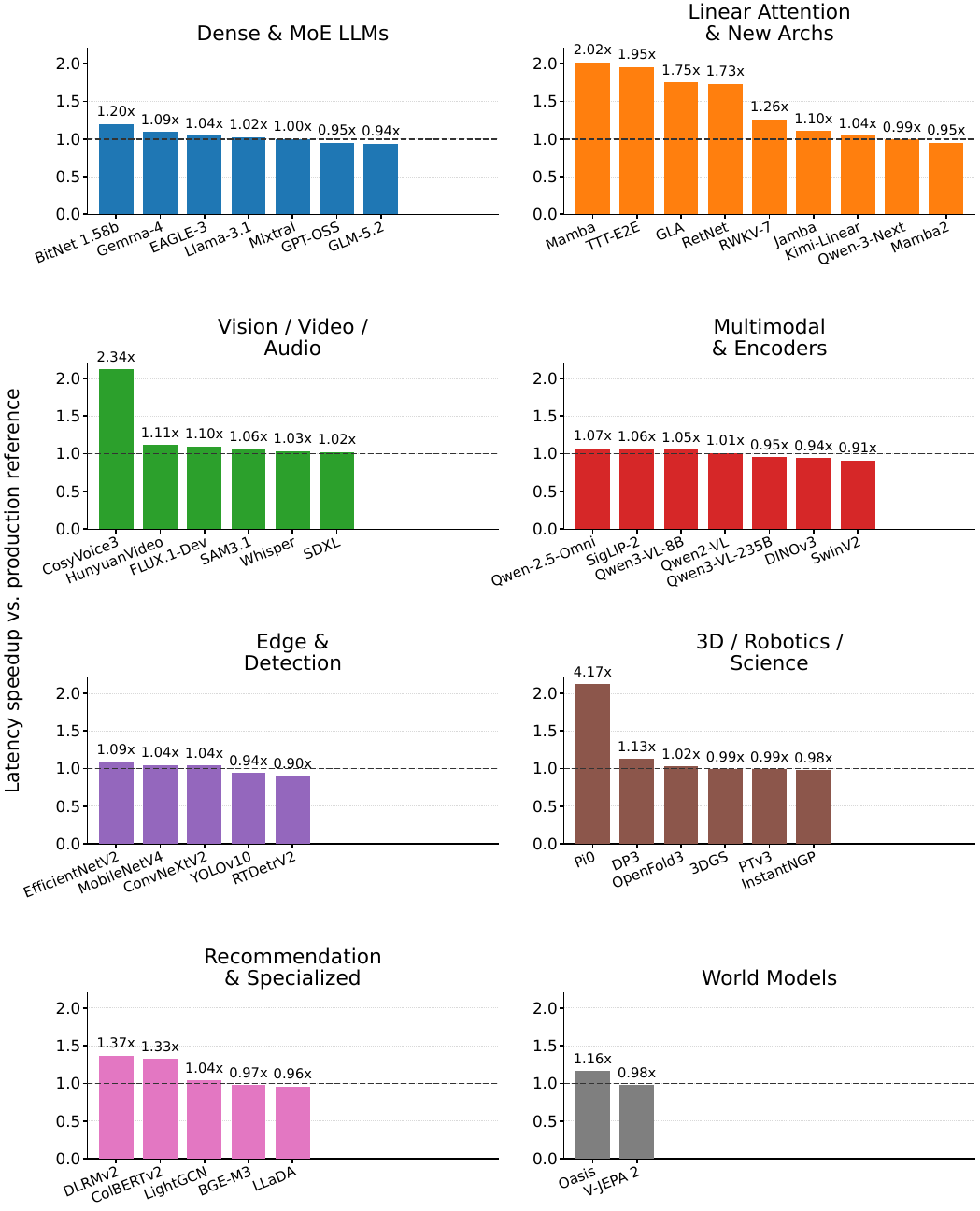}
  \caption{\textbf{Per-architecture latency speedup on B200}, drawn as in Figure~\ref{fig:eval-thr-breakdown}.}
  \label{fig:eval-lat-breakdown}
\end{figure}

Table~\ref{tab:bench_detail} and Table~\ref{tab:bench_detail_b} report the B200 L4 validate run for every architecture; Appendix~\ref{app:reference_perf_h200} repeats it on H200.
Figures~\ref{fig:eval-thr-breakdown} and~\ref{fig:eval-lat-breakdown} plot the per-architecture B200 speedups; Figures~\ref{fig:eval-e2e-h200}--\ref{fig:eval-lat-breakdown-h200} are the H200 counterparts (the category-level B200 bars are Figure~\ref{fig:eval-e2e} in the main text).
For each row we list the production reference, the public workload mix, the geometric mean of throughput and latency speedups over that mix, and the alignment metric against the reference.
Autoregressive rows summarize mean prefix-match length and exact-match rate across throughput scenarios; cosine, top-1, box/label, NLL, and pass flags are reported as published by the corresponding harness.
If a model has no latency workload, Lat.\ is ``---''.
Per-row geometric means match the aggregator used in \textsc{MacroEval} (Section~\ref{sec:metrics}) and in Section~\ref{sec:eval-results}.

\subsection{Blackwell (B200) Results}
\label{app:reference_perf_b200}

\begin{table}[H]
      \centering
      \caption{End-to-end B200 validate results for \fastkernels{} architectures (Part I).
      \emph{Thr.} and \emph{Lat.} are geometric-mean speedups versus the named production reference.
      \emph{Correctness} is the harness alignment summary (prefix-match length and exact-match rate for autoregressive models; cosine, top-1, or pass flags otherwise).}
      \label{tab:bench_detail}
      \scriptsize
      \setlength{\tabcolsep}{2.5pt}
      \resizebox{\linewidth}{!}{%
      \begin{tabular}{@{}ll>{\raggedright\arraybackslash}p{1.55cm}>{\raggedright\arraybackslash}p{2.6cm}cc>{\raggedright\arraybackslash}p{2.35cm}@{}}
      \toprule
      \textbf{Category} & \textbf{Architecture} & \textbf{Reference} & \textbf{Workload} & \textbf{Thr.} & \textbf{Lat.} & \textbf{Correctness} \\
      \midrule
      \multirow{7}{*}{Dense \& MoE LLMs}
       & Llama-3.1 & \texttt{vLLM} & WildChat mixed / long-context & $0.95\times$ & $1.02\times$ & prefix 136.3 / EM 28.6\% \\
       & Mixtral & \texttt{vLLM} & WildChat mixed / long-context & $1.06\times$ & $1.00\times$ & prefix 86.3 / EM 12.2\% \\
       & GLM-5.2 (NVFP4) & \texttt{vLLM} & WildChat mixed / long-context & $1.41\times$ & $0.94\times$ & prefix 33.5 / EM 8.9\% \\
       & BitNet 1.58b & \makecell[l]{Microsoft \\ BitNet GPU} & WildChat mixed / long-context & $1.10\times$ & $1.20\times$ & prefix 122.1 / EM 11.0\% \\
       & GPT-OSS (MXFP4) & \texttt{vLLM} & WildChat mixed / long-context & $0.95\times$ & $0.95\times$ & prefix 53.8 / EM 8.2\% \\
       & EAGLE-3 & \texttt{SGLang} & WildChat EAGLE-3 speculative decoding & $0.93\times$ & $1.04\times$ & prefix 160.6 / EM 37.5\% \\
       & Gemma-4 & \texttt{vLLM} & WildChat mixed / long-context & $1.00\times$ & $1.09\times$ & prefix 70.0 / EM 9.6\% \\
      \midrule
      \multirow{9}{*}{\makecell[l]{Linear Attention\\\& New Archs}}
       & Mamba & \texttt{vLLM} & WildChat mixed / long-context & $2.10\times$ & $2.02\times$ & prefix 129.5 / EM 31.9\% \\
       & Mamba2 & \texttt{vLLM} & WildChat mixed / long-context & $0.96\times$ & $0.95\times$ & prefix 203.1 / EM 54.6\% \\
       & RWKV-7 & \texttt{FLA} & mixed / long-context serving & $1.22\times$ & $1.26\times$ & prefix 100.7 / EM 14.8\% \\
       & GLA & \texttt{FLA} & mixed / long-context serving & $1.44\times$ & $1.75\times$ & prefix 262.1 / EM 70.9\% \\
       & RetNet & \texttt{FLA} & mixed / long-context serving & $3.40\times$ & $1.73\times$ & prefix 94.4 / EM 29.9\% \\
       & Qwen-3-Next & \texttt{vLLM} & mixed / long-context serving & $1.21\times$ & $0.99\times$ & prefix 11.7 / EM 1.4\% \\
       & Kimi-Linear & \texttt{vLLM} & mixed / long-context serving & $1.21\times$ & $1.04\times$ & prefix 63.8 / EM 9.2\% \\
       & TTT-E2E & JAX ref. & pretrain + meta forward & $1.95\times$ & $1.95\times$ & token NLL $\Delta$ 0.022 \\
       & Jamba & \texttt{vLLM} & mixed / long-context serving & $0.84\times$ & $1.10\times$ & prefix 91.4 / EM 15.6\% \\
      \midrule
      \multirow{6}{*}{\makecell[l]{Vision / Video /\\Audio}}
       & FLUX.1-Dev & \texttt{vllm-omni} & Parti-prompts; 512 / 1024 & $1.31\times$ & $1.10\times$ & min cos 0.949 \\
       & HunyuanVideo-1.5 & \texttt{vllm-omni} & 480p short / medium T2V & $1.01\times$ & $1.11\times$ & min cos 0.959 \\
       & SDXL & \texttt{diffusers} & Parti-prompts; 512 / 1024 & $1.03\times$ & $1.02\times$ & min cos 0.919 \\
       & SAM3.1 & \texttt{sam3} & image pipeline + video frames & $1.04\times$ & $1.06\times$ & pass / min cos 0.871 \\
       & Whisper & \texttt{vLLM} & LibriSpeech test.clean & $1.07\times$ & $1.03\times$ & prefix 385.9 / EM 63.5\% \\
       & CosyVoice3 & \texttt{vllm-omni} & SEED-TTS-Eval; short / medium / long & $2.30\times$ & $2.34\times$ & mel cos 0.991 \\
      \midrule
      \multirow{7}{*}{\makecell[l]{Multimodal \&\\Encoders}}
       & Qwen2-VL & \texttt{vLLM} & text / image / video & $1.24\times$ & $1.01\times$ & prefix 246.0 / EM 43.9\% \\
       & Qwen3-VL-8B & \texttt{vLLM} & text / image / video & $1.26\times$ & $1.05\times$ & prefix 141.1 / EM 20.2\% \\
       & Qwen3-VL-235B (FP8) & \texttt{vLLM} & text / image / video & $1.59\times$ & $0.95\times$ & prefix 40.1 / EM 2.9\% \\
       & Qwen-2.5-Omni & \texttt{vLLM} & text / image / video / audio & $1.38\times$ & $1.07\times$ & prefix 161.4 / EM 26.9\% \\
       & SigLIP-2 & \texttt{timm} & default-res + high-res & $1.01\times$ & $1.06\times$ & min cos 1.000 \\
       & DINOv3 & \texttt{timm} & default-res + high-res & $0.94\times$ & $0.94\times$ & min cos 1.000 \\
       & SwinV2 & \texttt{timm} & default-res + high-res & $0.94\times$ & $0.91\times$ & min cos 1.000 \\
      \bottomrule
      \end{tabular}%
      }
  \end{table}

  \begin{table}[H]
      \centering
      \caption{End-to-end B200 validate results for \fastkernels{} architectures (Part II).}
      \label{tab:bench_detail_b}
      \scriptsize
      \setlength{\tabcolsep}{2.5pt}
      \resizebox{\linewidth}{!}{%
      \begin{tabular}{@{}ll>{\raggedright\arraybackslash}p{1.55cm}>{\raggedright\arraybackslash}p{2.6cm}cc>{\raggedright\arraybackslash}p{2.35cm}@{}}
      \toprule
      \textbf{Category} & \textbf{Architecture} & \textbf{Reference} & \textbf{Workload} & \textbf{Thr.} & \textbf{Lat.} & \textbf{Correctness} \\
      \midrule
      \multirow{5}{*}{\makecell[l]{Edge \&\\Detection}}
       & MobileNetV4 & \texttt{timm} & default-res + high-res & $1.03\times$ & $1.04\times$ & min cos 1.000 \\
       & ConvNeXtV2 & \texttt{transformers} & Food101 validation & $1.00\times$ & $1.04\times$ & top-1 1.000 \\
       & EfficientNetV2 & \texttt{timm} & Food101 validation & $1.04\times$ & $1.09\times$ & top-1 1.000 \\
       & YOLOv10 & \makecell[l]{THU-MIG/ \\ yolov10} & COCO val detection & $1.14\times$ & $0.94\times$ & box 0.997 / score 1.000 / lbl 100\% \\
       & RTDetrV2 & \texttt{transformers} & COCO val detection & $0.97\times$ & $0.90\times$ & box 0.928 / score 1.000 / lbl 73.4\% \\
      \midrule
      \multirow{6}{*}{\makecell[l]{3D / Robotics /\\Science}}
       & 3DGS & \texttt{gsplat} & deterministic orbit render & $1.00\times$ & $0.99\times$ & pass / RGB cos 1.000 \\
       & InstantNGP & \texttt{pyngp} & fox render & $0.98\times$ & $0.98\times$ & pass / RGBA cos 1.000 \\
       & PointTransformerV3 & PTv3 ref. & ScanObjectNN + batch-8 & $0.98\times$ & $0.99\times$ & feat.\ cos 0.986 (fail) \\
       & OpenFold3 & OpenFold3 ref. & 4 length buckets & $1.01\times$ & $1.02\times$ & align 100\% \\
       & Pi0 & \texttt{OpenPI} & ALOHA 3-cam / 1-cam & $3.09\times$ & $4.17\times$ & pass / action cos 1.000 \\
       & DP3 & \makecell[l]{3D-Diffusion- \\ Policy} & 1-env + batch-8 & $1.13\times$ & $1.13\times$ & min cos 1.000 \\
      \midrule
      \multirow{5}{*}{\makecell[l]{Recommendation\\\& Specialized}}
       & DLRMv2 & \texttt{TorchRec} & Adult Census CTR batch & $1.10\times$ & $1.37\times$ & cosine undefined \\
       & LightGCN & \texttt{PyG} & recommend-batch & $1.02\times$ & $1.04\times$ & pass / cos 1.000 \\
       & BGE-M3 & \texttt{vLLM} & MLDR documents & $2.35\times$ & $0.97\times$ & pass / cos 0.999 \\
       & ColBERTv2 & \texttt{vLLM} & MS MARCO passages & $3.13\times$ & $1.33\times$ & pass / cos 1.000 \\
       & LLaDA & \texttt{Fast-dLLM} & HumanEval Fast-dLLM dual & $0.96\times$ & $0.96\times$ & prefix 168.6 / EM 56.1\% \\
      \midrule
      \multirow{2}{*}{World Models}
       & Oasis & \texttt{open-oasis} & Minecraft clips; 4 DDIM settings & $1.17\times$ & $1.16\times$ & pass / min cos 0.997 \\
       & V-JEPA 2 & \texttt{transformers} & predictor / encoder & $0.97\times$ & $0.98\times$ & pass / cos 1.000 \\
      \bottomrule
      \end{tabular}%
      }

      \vspace{4pt}
      {\small \textit{Note:} Speedup values are geometric means over all validate scenarios for that row; values below $1$ indicate slower execution than the reference.
      CosyVoice3 e2e correctness is the overall median mel-spectrogram cosine from the run log (the compact markdown table left this field blank).
      DLRMv2 reports \texttt{pass=False} with NaN cosine against TorchRec on this checkpoint; PointTransformerV3 reports \texttt{pass=False} at feature cosine $0.986$.
      RetNet's $3.40\times$ throughput geomean still mixes a $6.83\times$ mixed scenario (prefix 21.2) with $1.69\times$ long-context (prefix 167.6).}
  \end{table}

\subsection{Hopper (H200) Results}
\label{app:reference_perf_h200}

\begin{figure}[H]
  \centering
  \includegraphics[width=0.7\linewidth]{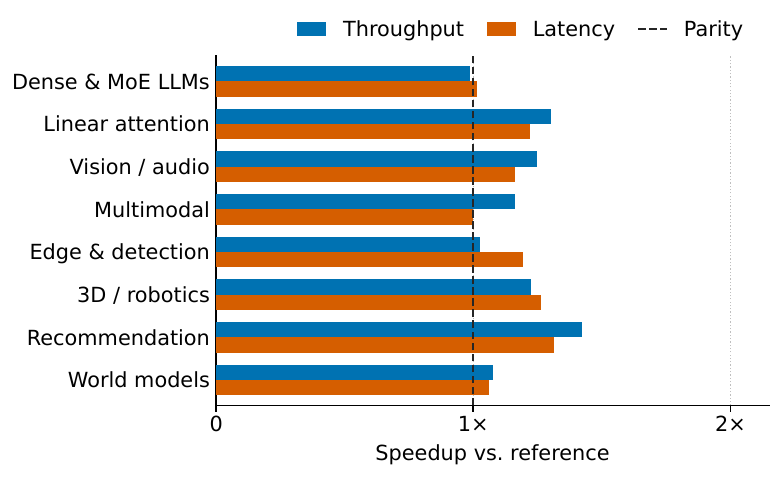}
  \caption{\textbf{End-to-end speedup by category on H200.}
  Counterpart of Figure~\ref{fig:eval-e2e}: each bar is the geometric mean over the architectures of that category in Tables~\ref{tab:bench_detail_h200} and~\ref{tab:bench_detail_h200_b} (GLM-5.2 is not run).
  The edge \& detection latency bar includes YOLOv10, which fails alignment on H200; without it, that bar is $0.98\times$.}
  \label{fig:eval-e2e-h200}
\end{figure}

\begin{figure}[H]
  \centering
  \includegraphics[width=\linewidth]{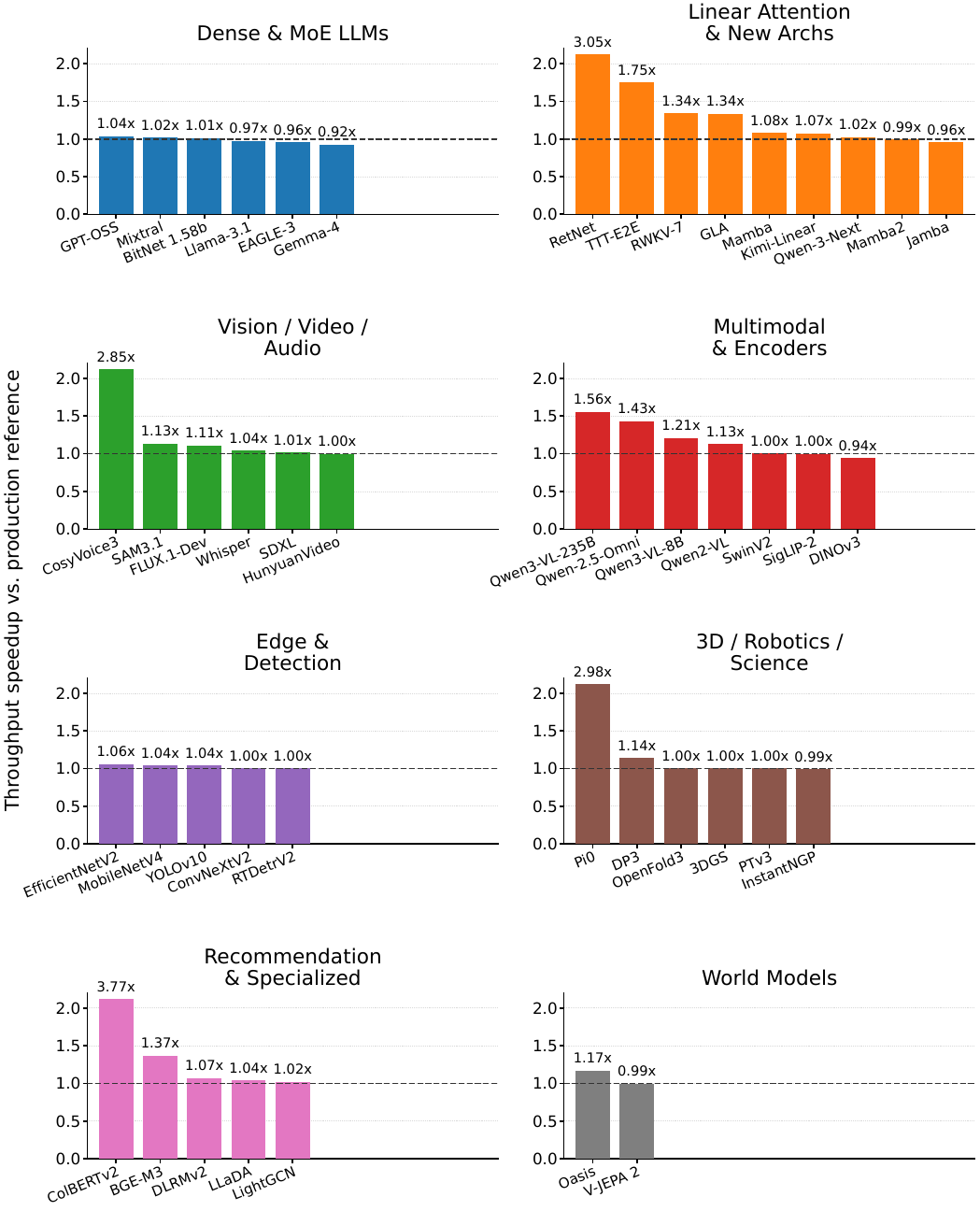}
  \caption{\textbf{Per-architecture throughput speedup on H200}, grouped by category. Bars taller than the axis are clipped and labeled with their value; the dashed line marks parity.}
  \label{fig:eval-thr-breakdown-h200}
\end{figure}

\begin{figure}[H]
  \centering
  \includegraphics[width=\linewidth]{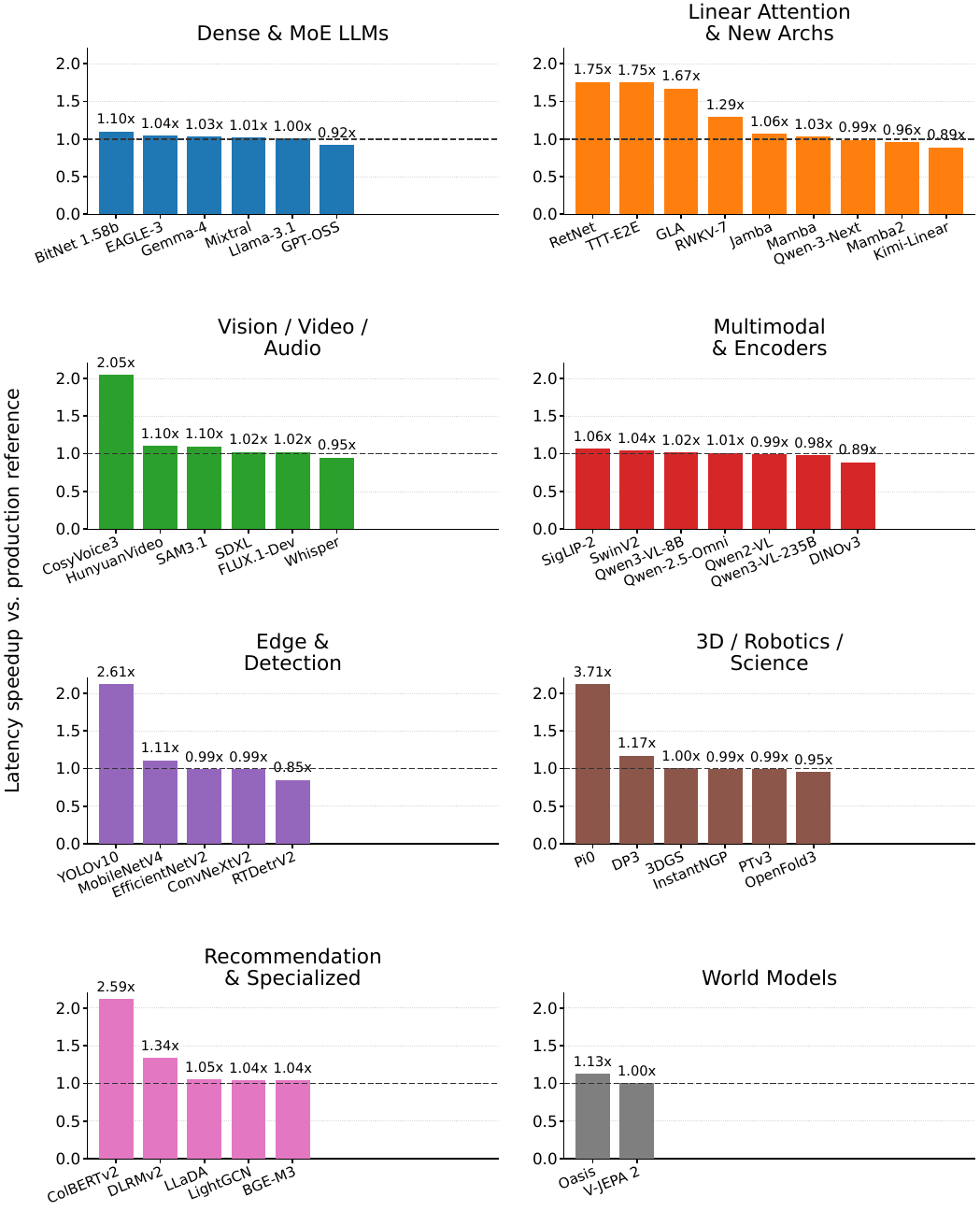}
  \caption{\textbf{Per-architecture latency speedup on H200}, drawn as in Figure~\ref{fig:eval-thr-breakdown-h200}. YOLOv10's $2.61\times$ comes from a run that fails alignment (Table~\ref{tab:bench_detail_h200_b}).}
  \label{fig:eval-lat-breakdown-h200}
\end{figure}

Tables~\ref{tab:bench_detail_h200} and~\ref{tab:bench_detail_h200_b} repeat the L4 validate run on $8\times$H200 (Appendix~\ref{app:experimental}) with the same scenario table, workloads, references, and aggregation as the B200 tables above.
Figure~\ref{fig:eval-e2e-h200} is the H200 counterpart of Figure~\ref{fig:eval-e2e}, and Figures~\ref{fig:eval-thr-breakdown-h200} and~\ref{fig:eval-lat-breakdown-h200} are the H200 counterparts of Figures~\ref{fig:eval-thr-breakdown} and~\ref{fig:eval-lat-breakdown}.
Over the 46 architectures that run on both GPUs, the throughput geomean is $1.19\times$ on H200 versus $1.22\times$ on B200, and the latency geomean is $1.15\times$ on both.
The four largest H200 throughput gains (ColBERTv2, RetNet, Pi0, CosyVoice3) are also among the five largest on B200, and dense and MoE LLM serving stays at parity ($0.99\times$); the largest drops relative to B200 are Mamba ($2.10\times \to 1.08\times$) and BGE-M3 ($2.35\times \to 1.37\times$).

\begin{table}[H]
      \centering
      \caption{End-to-end H200 validate results for \fastkernels{} architectures (Part I). Columns as in Table~\ref{tab:bench_detail}.
      $^\dagger$Re-run serially on a second $8\times$H200 node (see the note under Table~\ref{tab:bench_detail_h200_b}).}
      \label{tab:bench_detail_h200}
      \scriptsize
      \setlength{\tabcolsep}{2.5pt}
      \resizebox{\linewidth}{!}{%
      \begin{tabular}{@{}ll>{\raggedright\arraybackslash}p{1.55cm}>{\raggedright\arraybackslash}p{2.6cm}cc>{\raggedright\arraybackslash}p{2.35cm}@{}}
      \toprule
      \textbf{Category} & \textbf{Architecture} & \textbf{Reference} & \textbf{Workload} & \textbf{Thr.} & \textbf{Lat.} & \textbf{Correctness} \\
      \midrule
      \multirow{7}{*}{Dense \& MoE LLMs}
       & Llama-3.1 & \texttt{vLLM} & WildChat mixed / long-context & $0.97\times$ & $1.00\times$ & prefix 140.3 / EM 28.5\% \\
       & Mixtral & \texttt{vLLM} & WildChat mixed / long-context & $1.02\times$ & $1.01\times$ & prefix 72.9 / EM 9.7\% \\
       & GLM-5.2 (NVFP4) & \texttt{vLLM} & WildChat mixed / long-context & --- & --- & not run (NVFP4 is Blackwell-only) \\
       & BitNet 1.58b & \makecell[l]{Microsoft \\ BitNet GPU} & WildChat mixed / long-context & $1.01\times$ & $1.10\times$ & prefix 108.2 / EM 7.8\% \\
       & GPT-OSS (MXFP4) & \texttt{vLLM} & WildChat mixed / long-context & $1.04\times$ & $0.92\times$ & prefix 63.0 / EM 8.8\% \\
       & EAGLE-3$^\dagger$ & \texttt{SGLang} & WildChat EAGLE-3 speculative decoding & $0.96\times$ & $1.04\times$ & prefix 179.2 / EM 50.0\% \\
       & Gemma-4 & \texttt{vLLM} & WildChat mixed / long-context & $0.92\times$ & $1.03\times$ & prefix 72.3 / EM 7.3\% \\
      \midrule
      \multirow{9}{*}{\makecell[l]{Linear Attention\\\& New Archs}}
       & Mamba & \texttt{vLLM} & WildChat mixed / long-context & $1.08\times$ & $1.03\times$ & prefix 147.7 / EM 36.9\% \\
       & Mamba2 & \texttt{vLLM} & WildChat mixed / long-context & $0.99\times$ & $0.96\times$ & prefix 207.1 / EM 53.5\% \\
       & RWKV-7 & \texttt{FLA} & mixed / long-context serving & $1.34\times$ & $1.29\times$ & prefix 99.8 / EM 12.7\% \\
       & GLA & \texttt{FLA} & mixed / long-context serving & $1.34\times$ & $1.67\times$ & prefix 264.1 / EM 71.1\% \\
       & RetNet & \texttt{FLA} & mixed / long-context serving & $3.05\times$ & $1.75\times$ & prefix 98.8 / EM 32.2\% \\
       & Qwen-3-Next & \texttt{vLLM} & mixed / long-context serving & $1.02\times$ & $0.99\times$ & prefix 11.4 / EM 1.3\% \\
       & Kimi-Linear & \texttt{vLLM} & mixed / long-context serving & $1.07\times$ & $0.89\times$ & prefix 18.3 / EM 4.0\% \\
       & TTT-E2E & JAX ref. & pretrain + meta forward & $1.75\times$ & $1.75\times$ & token NLL $\Delta$ 0.022 \\
       & Jamba$^\dagger$ & \texttt{vLLM} & mixed / long-context serving & $0.96\times$ & $1.06\times$ & prefix 98.3 / EM 16.4\% \\
      \midrule
      \multirow{6}{*}{\makecell[l]{Vision / Video /\\Audio}}
       & FLUX.1-Dev & \texttt{vllm-omni} & Parti-prompts; 512 / 1024 & $1.11\times$ & $1.02\times$ & min cos 0.981 \\
       & HunyuanVideo-1.5 & \texttt{vllm-omni} & 480p short / medium T2V & $1.00\times$ & $1.10\times$ & min cos 0.943 \\
       & SDXL & \texttt{diffusers} & Parti-prompts; 512 / 1024 & $1.01\times$ & $1.02\times$ & min cos 0.923 \\
       & SAM3.1 & \texttt{sam3} & image pipeline + video frames & $1.13\times$ & $1.10\times$ & pass / min cos 0.871 \\
       & Whisper & \texttt{vLLM} & LibriSpeech test.clean & $1.04\times$ & $0.95\times$ & prefix 386.1 / EM 63.6\% \\
       & CosyVoice3 & \texttt{vllm-omni} & SEED-TTS-Eval; short / medium / long & $2.85\times$ & $2.05\times$ & mel cos 0.991 \\
      \midrule
      \multirow{7}{*}{\makecell[l]{Multimodal \&\\Encoders}}
       & Qwen2-VL & \texttt{vLLM} & text / image / video & $1.13\times$ & $0.99\times$ & prefix 248.2 / EM 44.8\% \\
       & Qwen3-VL-8B & \texttt{vLLM} & text / image / video & $1.21\times$ & $1.02\times$ & prefix 139.5 / EM 19.9\% \\
       & Qwen3-VL-235B (FP8) & \texttt{vLLM} & text / image / video & $1.56\times$ & $0.98\times$ & prefix 41.9 / EM 2.8\% \\
       & Qwen-2.5-Omni & \texttt{vLLM} & text / image / video / audio & $1.43\times$ & $1.01\times$ & prefix 158.9 / EM 25.0\% \\
       & SigLIP-2$^\dagger$ & \texttt{timm} & default-res + high-res & $1.00\times$ & $1.06\times$ & min cos 0.986 \\
       & DINOv3$^\dagger$ & \texttt{timm} & default-res + high-res & $0.94\times$ & $0.89\times$ & min cos 1.000 \\
       & SwinV2$^\dagger$ & \texttt{timm} & default-res + high-res & $1.00\times$ & $1.04\times$ & min cos 1.000 \\
      \bottomrule
      \end{tabular}%
      }
  \end{table}

  \begin{table}[H]
      \centering
      \caption{End-to-end H200 validate results for \fastkernels{} architectures (Part II).}
      \label{tab:bench_detail_h200_b}
      \scriptsize
      \setlength{\tabcolsep}{2.5pt}
      \resizebox{\linewidth}{!}{%
      \begin{tabular}{@{}ll>{\raggedright\arraybackslash}p{1.55cm}>{\raggedright\arraybackslash}p{2.6cm}cc>{\raggedright\arraybackslash}p{2.35cm}@{}}
      \toprule
      \textbf{Category} & \textbf{Architecture} & \textbf{Reference} & \textbf{Workload} & \textbf{Thr.} & \textbf{Lat.} & \textbf{Correctness} \\
      \midrule
      \multirow{5}{*}{\makecell[l]{Edge \&\\Detection}}
       & MobileNetV4$^\dagger$ & \texttt{timm} & default-res + high-res & $1.04\times$ & $1.11\times$ & min cos 1.000 \\
       & ConvNeXtV2 & \texttt{transformers} & Food101 validation & $1.00\times$ & $0.99\times$ & top-1 1.000 \\
       & EfficientNetV2 & \texttt{timm} & Food101 validation & $1.06\times$ & $0.99\times$ & top-1 1.000 \\
       & YOLOv10 & \makecell[l]{THU-MIG/ \\ yolov10} & COCO val detection & $1.04\times$ & $2.61\times$ & box 0.461 / score 0.549 / lbl 94.6\% (fail) \\
       & RTDetrV2 & \texttt{transformers} & COCO val detection & $1.00\times$ & $0.85\times$ & box 0.928 / score 1.000 / lbl 73.4\% \\
      \midrule
      \multirow{6}{*}{\makecell[l]{3D / Robotics /\\Science}}
       & 3DGS & \texttt{gsplat} & deterministic orbit render & $1.00\times$ & $1.00\times$ & pass / RGB cos 1.000 \\
       & InstantNGP & \texttt{pyngp} & fox render & $0.99\times$ & $0.99\times$ & pass / RGBA cos 1.000 \\
       & PointTransformerV3 & PTv3 ref. & ScanObjectNN + batch-8 & $1.00\times$ & $0.99\times$ & feat.\ cos 0.986 (fail) \\
       & OpenFold3 & OpenFold3 ref. & 4 length buckets & $1.00\times$ & $0.95\times$ & align 100\% \\
       & Pi0 & \texttt{OpenPI} & ALOHA 3-cam / 1-cam & $2.98\times$ & $3.71\times$ & pass / action cos 1.000 \\
       & DP3 & \makecell[l]{3D-Diffusion- \\ Policy} & 1-env + batch-8 & $1.14\times$ & $1.17\times$ & min cos 1.000 \\
      \midrule
      \multirow{5}{*}{\makecell[l]{Recommendation\\\& Specialized}}
       & DLRMv2 & \texttt{TorchRec} & Adult Census CTR batch & $1.07\times$ & $1.34\times$ & cosine undefined \\
       & LightGCN & \texttt{PyG} & recommend-batch & $1.02\times$ & $1.04\times$ & pass / cos 1.000 \\
       & BGE-M3 & \texttt{vLLM} & MLDR documents & $1.37\times$ & $1.04\times$ & pass / cos 0.997 \\
       & ColBERTv2 & \texttt{vLLM} & MS MARCO passages & $3.77\times$ & $2.59\times$ & pass / cos 1.000 \\
       & LLaDA & \texttt{Fast-dLLM} & HumanEval Fast-dLLM dual & $1.04\times$ & $1.05\times$ & prefix 166.2 / EM 53.7\% \\
      \midrule
      \multirow{2}{*}{World Models}
       & Oasis & \texttt{open-oasis} & Minecraft clips; 4 DDIM settings & $1.17\times$ & $1.13\times$ & pass / min cos 0.996 \\
       & V-JEPA 2 & \texttt{transformers} & predictor / encoder & $0.99\times$ & $1.00\times$ & pass / cos 1.000 \\
      \bottomrule
      \end{tabular}%
      }

      \vspace{4pt}
      {\small \textit{Note:} Aggregation as in Tables~\ref{tab:bench_detail}--\ref{tab:bench_detail_b}.
      $^\dagger$EAGLE-3, Jamba, and the four \texttt{timm} encoders failed in the parallel run for environment reasons (an SGLang~0.5.20 \texttt{ServerArgs} API change, a CUDA illegal-memory-access crash that did not reproduce, and gated ImageNet access, respectively) and were re-run serially on a second $8\times$H200 node with the same software stack; the EAGLE-3 re-run also needed a two-line compatibility fix in the SGLang harness.
      YOLOv10 fails alignment on H200 (box cosine $0.461$ vs.\ $0.997$ on B200), so its H200 speedups, notably the $2.61\times$ latency, are not like-for-like and should not be read as gains.
      DLRMv2 again reports a degenerate cosine ($0$ here, NaN on B200) with a maximum absolute difference of $0$; PointTransformerV3 fails at feature cosine $0.986$ as on B200.
      RetNet's $3.05\times$ throughput geomean mixes a $5.61\times$ mixed scenario (prefix 19.1) with $1.66\times$ long-context (prefix 178.4).}
  \end{table}

\section{Baseline Versioning and Drift}
\label{app:drift}

Production kernels are a moving target, so \fastkernels{} is versioned: each release pins the exact revision of every reference library (vLLM, FlashInfer, FLA, diffusers, \dots) per target GPU, and the results in this paper use vLLM~0.26.0 (released 2026-07-25).
A speedup over a \fastkernels{} baseline is therefore a speedup over production at that release.
To track drift, \fastkernels{} ships a daily CI job (\texttt{scripts/baseline\_drift.sh}): whenever a new vLLM release appears, it re-benchmarks only the reference models under that release on each pinned GPU type, in isolated environments with one job per node, and computes per-family drift coefficients (per-model geometric mean over workloads, models weighted equally within a family).
Dividing a speedup measured against the pinned baseline by the family coefficient estimates the speedup against the newer release; when any family moves by more than 10\%, a re-release with re-pinned references is prepared.

To estimate the cadence this implies, we ran the same measurement over the 127 days between vLLM~0.18.0 (2026-03-20) and 0.26.0 on 8$\times$H200 and 8$\times$B200, on the 13 reference models both releases serve, with identical workloads (Table~\ref{tab:drift}).
The median model moved by only $1.03\times$ (throughput) and $1.01\times$ (latency) on H200, and by $1.07\times$ and $1.09\times$ on B200.
Drift is concentrated rather than uniform: one family crossed the 10\% threshold on H200 (linear attention and SSMs, $1.15\times$ throughput) and two on B200, where the LLM family is dominated by gpt-oss-120b's MXFP4 path maturing on Blackwell ($2.8\times$, against $1.04\times$ on H200).
Newer releases are not uniformly faster: Qwen3-VL-235B-FP8 lost throughput on both GPUs ($0.83\times$ and $0.87\times$).
This implies a re-release at least every four months, sooner on newer hardware whose support for new number formats is still maturing---comparable to the cadence at which any benchmark that stays current must add newly released architectures.

\begin{table}[H]
  \centering
  \caption{\textbf{Reference drift from vLLM~0.18.0 to 0.26.0} (127 days), as computed by the baseline-drift CI job: speedup of the newer release (${>}1$: faster), per-model geometric mean over workloads, models weighted equally within a family.
  Latencies are 5-iteration medians (${<}1\%$ spread); throughput is one timed pass per workload, which three repeated runs per release of Qwen3-VL-235B-FP8 place within 1.5\%.
  On B200, six models needed \texttt{nvidia-cutlass-dsl==4.5.3} to import under vLLM~0.18.0.}
  \label{tab:drift}
  \small
\begin{tabular}{@{}lr rr rr@{}}
\toprule
 & & \multicolumn{2}{c}{\textbf{H200}} & \multicolumn{2}{c}{\textbf{B200}} \\
\cmidrule(lr){3-4}\cmidrule(lr){5-6}
Model family & $n$ & Thr. & Lat. & Thr. & Lat. \\
\midrule
LLM (dense + MoE) & 3 & 1.09$\times$ & 1.08$\times$ & 1.44$\times$ & 1.37$\times$ \\
Linear attention / SSM / hybrid & 5 & 1.15$\times$ & 1.03$\times$ & 1.14$\times$ & 1.18$\times$ \\
Multimodal / VLM & 4 & 0.98$\times$ & 1.05$\times$ & 1.01$\times$ & 1.07$\times$ \\
Audio & 1 & 1.02$\times$ & 1.02$\times$ & 1.02$\times$ & 1.07$\times$ \\
\midrule
Median model & 13 & 1.03$\times$ & 1.01$\times$ & 1.07$\times$ & 1.09$\times$ \\
\bottomrule
\end{tabular}

\end{table}

\section{Agent Selection, Configuration, and Cost}
\label{app:agents}

\paragraph{Selection.}
We chose agents to cover the three lines of work that currently produce the strongest kernels.
From the KernelBench ecosystem we run Dr.~Kernel~\citep{drkernel}, one of its most recent agents; of the others we tried, KernelAgent~\citep{kernelagent} did not reproduce its reported results in our setting, and the rest assume KernelBench's single-file tasks with a self-contained reference and cannot load production-shaped modules that require configuration objects, KV-cache wrappers, packed quantized weights, or a distributed context.
From the MLSys 2026 FlashInfer AI Kernel Generation Contest~\citep{mlsys26contest}, whose tasks were drawn from production LLM workloads and scored against FlashInfer's expert kernels, we run KDA~\citep{kda}, whose full-agent entry placed first on MoE, second on DSA, and third on GDN, and AKO~\citep{ako} (its AKO4X harness was adapted by the GEMM People entry, which placed second on MoE), whose release reports beating the FlashInfer expert kernels on 9 of 10 operator families on B200.
Finally, we run Claude Code and Codex~\citep{codex} directly, with no harness.
Because KDA and AKO drive the same Claude model as our Claude Code runs (KDA adds a GPT reviewer), comparing them with Claude Code isolates the contribution of the harness.

\paragraph{Configurations.}
Table~\ref{tab:agent-config} summarizes how each agent was run.
All agents receive the same task: the baseline module, its captured cases, and \texttt{fastkernels bench} on a GPU that is leased only while they benchmark or profile.
A candidate becomes a \emph{winner} only if every non-skipped case passes and its geometric-mean speedup is at least $1\times$; an unchanged baseline that passes counts as correct at $1\times$ but not as a fast$_1$ win.
The independent campaign optimizes all 133 operators in standalone mode; the sequential campaign regenerates the 86 L2 and L3 operators on top of the frozen lower-level winners.

\begin{table}[H]
  \centering
  \caption{\textbf{Agent configurations.} \emph{Cap}: hang cap on one agent session (the candidate is kept on timeout), not a target budget.}
  \label{tab:agent-config}
  \scriptsize
  \setlength{\tabcolsep}{3pt}
  \begin{tabular}{@{}l>{\raggedright\arraybackslash}p{3.4cm}>{\raggedright\arraybackslash}p{6.2cm}l@{}}
  \toprule
  \textbf{Agent} & \textbf{Model(s)} & \textbf{Loop per operator} & \textbf{Cap} \\
  \midrule
  Dr.~Kernel & released \texttt{drkernel-14b} & 8 samples $\times$ 3 generate--execute--feedback turns (temperature 1.0, top-$p$ 0.95) & --- \\
  Claude Code & Claude Opus 5 (1M context, high effort) & one session with the default tools & 5\,h \\
  Codex & GPT-5.6 sol (high) & one session, same prompt as Claude Code & 5\,h \\
  KDA & Claude Opus 5 (draft, plan, implement); GPT-5.6 sol (review) & one planning cycle, then two implement--review rounds, with KernelWiki and Nsight Compute skills & 6\,h \\
  AKO & Claude Opus 5 (master and sub-agents) & two master-directed rounds, each a fresh sub-agent rewriting the kernel along one direction; improving kernels are archived & 5\,h/round \\
  \bottomrule
  \end{tabular}
\end{table}

\paragraph{Budget and the number of agents.}
Table~\ref{tab:agent-cost} reports what each agent spent: 6{,}914 agent-hours and 41.9B tokens (266M output tokens) in total, of which Claude Code, KDA, and AKO account for 1{,}763, 2{,}512, and 2{,}234 agent-hours.
These are the budgets at which such agents are used in practice: the MLSys 2026 contest gave teams 13 weeks, from the release of the starter kit (January 22) to the kernel deadline (April 24), to optimize five kernels~\citep{mlsys26contest}.
A leaderboard that samples many models for a few turns per task is far cheaper per entry, but it measures a different system than the long-horizon agents practitioners deploy, and the ranking of those agents is what \fastkernels{} is meant to report.
Each additional long-horizon agent costs about 2{,}000 agent-hours and 8--22B tokens on the default set.
Codex is the exception worth noting: its kernel-level results are comparable to the Claude-based agents' at 321 agent-hours.
\fastkernels{} imposes no structure on the agent, so any agent that writes candidate files can be scored with the same harness.

\paragraph{Evolutionary search.}
AlphaEvolve~\citep{alphaevolve} and KernelEvolve~\citep{kernelevolve} are not publicly available.
OpenEvolve~\citep{openevolve} is, and was the organizers' agent baseline in the MLSys 2026 contest, but evolutionary frameworks evaluate a population of candidates in every generation, and on \fastkernels{} each evaluation compiles a kernel and benchmarks it on its production-shaped cases (for collectives, on a full tensor-parallel group); their cost per operator therefore scales with population size times the number of generations, on top of the LLM calls that produce each candidate, which puts a 133-operator campaign beyond our budget.
The AKO authors also report that, during the contest, search-based methods performed far worse than letting a coding agent optimize kernels directly~\citep{ako}.

\begin{table}[H]
  \centering
  \caption{\textbf{Agent cost per campaign} (133 operators independently, 86 regenerated sequentially).
  \emph{Agent-hours}: sum over operators of each operator's wall-clock generation time, including relaunches; operators run in parallel, so this is not campaign wall-clock time.
  \emph{Tokens} as reported by each agent's CLI or API: input, output, and cached (re-read) context; output includes reasoning tokens.
  Dr.~Kernel runs a local model without prompt caching; KDA counts both its Claude and Codex sessions.}
  \label{tab:agent-cost}
  \small
  \setlength{\tabcolsep}{4pt}
\begin{tabular}{@{}ll r rr rrr@{}}
\toprule
 & & & \multicolumn{2}{c}{\textbf{Agent-hours}} & \multicolumn{3}{c}{\textbf{Tokens}} \\
\cmidrule(lr){4-5}\cmidrule(lr){6-8}
Agent & Campaign & Ops & Total & Median/op & Total (B) & Output (M) & Cached (\%) \\
\midrule
\multirow{2}{*}{Dr.~Kernel} & Independent & 133 & 53 & 0.26 & 0.03 & 10.4 & --- \\
 & Sequential & 86 & 32 & 0.34 & 0.02 & 8.0 & --- \\
\midrule
\multirow{2}{*}{Claude Code} & Independent & 133 & 1,171 & 7.27 & 4.35 & 31.4 & 97 \\
 & Sequential & 86 & 592 & 6.06 & 3.58 & 22.5 & 97 \\
\midrule
\multirow{2}{*}{KDA} & Independent & 133 & 1,751 & 12.33 & 12.31 & 62.8 & 96 \\
 & Sequential & 86 & 762 & 8.22 & 9.48 & 42.2 & 97 \\
\midrule
\multirow{2}{*}{AKO} & Independent & 133 & 1,421 & 9.43 & 5.47 & 47.4 & 97 \\
 & Sequential & 86 & 813 & 8.70 & 4.86 & 35.3 & 97 \\
\midrule
\multirow{2}{*}{Codex} & Independent & 133 & 253 & 1.86 & 1.02 & 4.0 & 98 \\
 & Sequential & 86 & 67 & 0.71 & 0.74 & 2.4 & 98 \\
\bottomrule
\end{tabular}

\end{table}

\section{Additional Kernel-Level Results}
\label{app:fastp}

\begin{figure}[H]
  \centering
  \includegraphics[width=\linewidth]{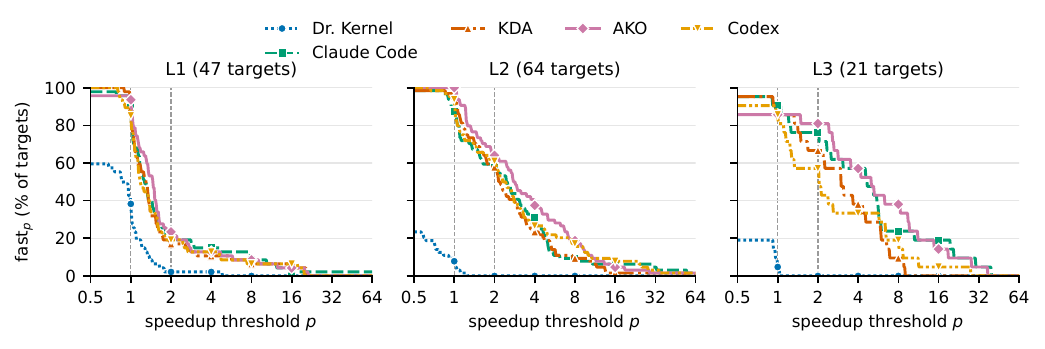}
  \caption{\textbf{fast$_p$ by level (independent mode).} Share of targets whose kernel is correct and more than $p\times$ faster than the production baseline; the value at the left edge approaches the share of correct kernels. Dashed lines mark $p=1$ and $p=2$.}
  \label{fig:kernel-fastp}
\end{figure}

Figure~\ref{fig:kernel-fastp} extends the fast$_1$ and fast$_2$ columns of Table~\ref{tab:kernel-levels} to every speedup threshold $p$, for the independent mode.

Table~\ref{tab:kernel-composition} compares the same operators across the three kernel-bench modes (Section~\ref{sec:eval-agents}).

\begin{table}[H]
  \centering
  \caption{\textbf{Composition effect and sequential generation}, over operators correct in both modes ($n$).
  \emph{Ratio}: geometric mean of the per-operator speedup ratio (composed / independent, or sequential / composed).
  \emph{Broke}: independent winners that fail correctness once composed.}
  \label{tab:kernel-composition}
  \small
  \setlength{\tabcolsep}{3pt}
  \begin{tabular}{@{}l rrr rrr rr rr@{}}
\toprule
 & \multicolumn{6}{c}{\makecell{\textbf{Composition effect}\\(Independent $\to$ Composed)}} & \multicolumn{4}{c}{\makecell{\textbf{Sequential}\\(Composed $\to$ Seq.)}} \\
\cmidrule(lr){2-7}\cmidrule(lr){8-11}
 & \multicolumn{3}{c}{L2} & \multicolumn{3}{c}{L3} & \multicolumn{2}{c}{L2} & \multicolumn{2}{c}{L3} \\
\cmidrule(lr){2-4}\cmidrule(lr){5-7}\cmidrule(lr){8-9}\cmidrule(lr){10-11}
Agent & $n$ & Ratio & Broke & $n$ & Ratio & Broke & $n$ & Ratio & $n$ & Ratio \\
\midrule
Dr.~Kernel & 5 & 0.99$\times$ & 0 & 1 & 1.00$\times$ & 0 & 1 & 1.78$\times$ & 0 & --- \\
Claude Code & 54 & 1.09$\times$ & 2 & 14 & 0.99$\times$ & 4 & 50 & 0.99$\times$ & 14 & 0.97$\times$ \\
KDA & 51 & 0.98$\times$ & 10 & 11 & 0.88$\times$ & 6 & 47 & 0.90$\times$ & 11 & 1.07$\times$ \\
AKO & 60 & 1.00$\times$ & 4 & 14 & 0.86$\times$ & 3 & 55 & 1.05$\times$ & 14 & 1.48$\times$ \\
Codex & 47 & 0.97$\times$ & 13 & 5 & 1.04$\times$ & 13 & 43 & 0.94$\times$ & 5 & 1.35$\times$ \\
\bottomrule
\end{tabular}

\end{table}

\section{End-to-End Results per Model}
\label{app:e2e}

Qwen3-VL-235B uses 200 requests per throughput workload and 32 correctness samples; GLA and OpenFold3 were re-run on cloud B200 nodes after a harness fix, and Codex's sets on a separate B200 node (its Llama baseline throughput is within 0.02\% of the other node's). Every set's speedups and discrepancies are relative to the baseline measured in the same run.
OpenFold3 runs eagerly and deterministically, so its noise run reproduces the reference bit for bit while its diffusion sampling amplifies any numerical change; its $g_i$ therefore comes from a calibration run that instead disables cuBLAS reduced-precision bf16 reductions, a benign numerical change.
Table~\ref{tab:e2e-drops} counts the kernels each set requests and keeps across models, and why kernels were dropped; \emph{Ind.\,+\,Seq.} pools an agent's two sets.
Table~\ref{tab:e2e-per-model} lists, for every model and agent set, the outcome of the all-winners run and of the deployable subset, the number of dropped kernels, the throughput and latency speedups and the mean correctness discrepancy $\bar d$, next to the model's noise floor ($\bar d$ of the noise run and the calibrated $g_i$).
Table~\ref{tab:e2e-sensitivity} reports the agent ranking under $0.5\times$--$2\times$ scaling of the correctness thresholds $g_i$ and $f_i$.

\begin{table}[H]
  \centering
  \caption{\textbf{Why kernel-level winners are dropped end to end.} Kernels are counted per (model, kernel) pair.
  Causes: \emph{Comp.}: breaks the compilation stack (not traceable by \texttt{torch.compile}, or breaks dynamic shapes); \emph{Run.}: crashes at run time on real engine shapes; \emph{T/O}: hangs; \emph{Imp.}: fails to import; \emph{Inc.}: runs but produces wrong outputs.
  \emph{AW crash / broken}: models where the full winner set crashes / runs with wrong outputs; \emph{Not conv.}: no deployable subset found within the attempt budget.}
  \label{tab:e2e-drops}
  \small
  \setlength{\tabcolsep}{3pt}
  \resizebox{\linewidth}{!}{
\begin{tabular}{@{}ll rr rrrrr rrrrr@{}}
\toprule
 & & \multicolumn{2}{c}{\textbf{Kernels}} & \multicolumn{5}{c}{\textbf{Dropped, by cause}} & \multicolumn{5}{c}{\textbf{Models}} \\
\cmidrule(lr){3-4}\cmidrule(lr){5-9}\cmidrule(lr){10-14}
Agent & Mode & Req. & Kept & Comp. & Run. & T/O & Imp. & Inc. & \makecell{AW\\crash} & \makecell{AW\\broken} & Empty & \makecell{Not\\conv.} & N/A \\
\midrule
\multirow{3}{*}{Dr.~Kernel} & Independent & 53 & 31 & 6 & 7 & 0 & 0 & 4 & 6 & 3 & 0 & 1 & 0 \\
 & Sequential & 54 & 25 & 6 & 7 & 0 & 0 & 5 & 6 & 3 & 0 & 2 & 0 \\
 & Ind.\,+\,Seq. & 107 & 56 & 12 & 14 & 0 & 0 & 9 & 12 & 6 & 0 & 3 & 0 \\
\midrule
\multirow{3}{*}{Claude Code} & Independent & 175 & 85 & 11 & 11 & 1 & 0 & 2 & 6 & 3 & 0 & 4 & 0 \\
 & Sequential & 190 & 93 & 12 & 8 & 1 & 0 & 3 & 8 & 2 & 0 & 4 & 0 \\
 & Ind.\,+\,Seq. & 365 & 178 & 23 & 19 & 2 & 0 & 5 & 14 & 5 & 0 & 8 & 0 \\
\midrule
\multirow{3}{*}{KDA} & Independent & 171 & 160 & 2 & 9 & 0 & 0 & 0 & 7 & 0 & 0 & 0 & 0 \\
 & Sequential & 193 & 178 & 9 & 5 & 0 & 1 & 0 & 5 & 0 & 0 & 0 & 0 \\
 & Ind.\,+\,Seq. & 364 & 338 & 11 & 14 & 0 & 1 & 0 & 12 & 0 & 0 & 0 & 0 \\
\midrule
\multirow{3}{*}{AKO} & Independent & 190 & 117 & 1 & 14 & 1 & 0 & 3 & 7 & 2 & 0 & 3 & 0 \\
 & Sequential & 201 & 122 & 14 & 19 & 0 & 1 & 2 & 8 & 1 & 0 & 2 & 0 \\
 & Ind.\,+\,Seq. & 391 & 239 & 15 & 33 & 1 & 1 & 5 & 15 & 3 & 0 & 5 & 0 \\
\midrule
\multirow{3}{*}{Codex} & Independent & 153 & 33 & 9 & 23 & 0 & 0 & 4 & 10 & 0 & 0 & 6 & 0 \\
 & Sequential & 180 & 96 & 14 & 24 & 0 & 2 & 4 & 9 & 1 & 0 & 2 & 0 \\
 & Ind.\,+\,Seq. & 333 & 129 & 23 & 47 & 0 & 2 & 8 & 19 & 1 & 0 & 8 & 0 \\
\bottomrule
\end{tabular}
}
\end{table}

\begin{table}[H]
  \centering
  \caption{\textbf{End-to-end results per model.} AW\,/\,Dep.: status of the all-winners run and of the deployable subset (n.c.: no deployable subset within the attempt budget); Drop: dropped kernels; $s_{\mathrm{thr}}$, $s_{\mathrm{lat}}$: throughput and latency speedup of the deployable subset over the production baseline; $\bar d$: mean per-sample discrepancy; $^\ddagger$: $g_i$ from the benign-numerics calibration run instead of the noise run.}
  \label{tab:e2e-per-model}
  \scriptsize
  \setlength{\tabcolsep}{2pt}
  \resizebox{\linewidth}{!}{
\begin{tabular}{@{}lrrl@{\hspace{12pt}}lrrrr@{\hspace{12pt}}lrrrr@{}}
\toprule
 & \multicolumn{2}{c}{\textbf{Noise}} & & \multicolumn{5}{c}{\textbf{Independent}} & \multicolumn{5}{c}{\textbf{Sequential}} \\
\cmidrule(lr){2-3}\cmidrule(lr){5-9}\cmidrule(lr){10-14}
Model & $\bar d$ & $g_i$ & Agent & AW\,/\,Dep. & Drop & $s_{\mathrm{thr}}$ & $s_{\mathrm{lat}}$ & $\bar d$ & AW\,/\,Dep. & Drop & $s_{\mathrm{thr}}$ & $s_{\mathrm{lat}}$ & $\bar d$ \\
\midrule
\multirow{5}{*}{Llama-3.1-8B} & \multirow{5}{*}{0.005} & \multirow{5}{*}{0.018} & Dr.~Kernel & crash\,/\,ok & 3 & 0.96 & 0.99 & 0.000 & crash\,/\,ok & 3 & 1.01 & 1.00 & 0.000 \\
 &  &  & Claude Code & crash\,/\,n.c. & 3 & --- & --- & --- & crash\,/\,n.c. & 4 & --- & --- & --- \\
 &  &  & KDA & crash\,/\,ok & 2 & 1.02 & 1.02 & 0.002 & crash\,/\,ok & 6 & 1.01 & 1.01 & 0.000 \\
 &  &  & AKO & crash\,/\,ok & 2 & 1.02 & 1.02 & 0.006 & crash\,/\,ok & 6 & 1.02 & 1.02 & 0.000 \\
 &  &  & Codex & crash\,/\,ok & 2 & 0.98 & 1.00 & 0.004 & crash\,/\,ok & 3 & 1.01 & 1.02 & 0.004 \\
\midrule
\multirow{5}{*}{GPT-OSS-120B} & \multirow{5}{*}{0.000} & \multirow{5}{*}{0.001} & Dr.~Kernel & crash\,/\,ok & 2 & 1.01 & 0.98 & 0.000 & crash\,/\,ok & 2 & 1.01 & 0.98 & 0.000 \\
 &  &  & Claude Code & crash\,/\,ok & 3 & 1.03 & 0.99 & 0.014 & crash\,/\,ok & 3 & 1.02 & 1.01 & 0.021 \\
 &  &  & KDA & crash\,/\,ok & 3 & 1.02 & 0.99 & 0.014 & crash\,/\,ok & 3 & 1.01 & 1.03 & 0.014 \\
 &  &  & AKO & crash\,/\,n.c. & 3 & --- & --- & --- & crash\,/\,ok & 7 & 1.00 & 0.93 & 0.012 \\
 &  &  & Codex & crash\,/\,ok & 4 & 0.98 & 1.02 & 0.000 & crash\,/\,ok & 5 & 0.99 & 1.02 & 0.000 \\
\midrule
\multirow{5}{*}{Qwen3-Next-80B} & \multirow{5}{*}{0.000} & \multirow{5}{*}{0.001} & Dr.~Kernel & broken\,/\,ok & 1 & 1.02 & 0.99 & 0.021 & broken\,/\,ok & 1 & 1.02 & 0.99 & 0.021 \\
 &  &  & Claude Code & crash\,/\,n.c. & 3 & --- & --- & --- & crash\,/\,ok & 1 & 1.02 & 1.03 & 0.019 \\
 &  &  & KDA & crash\,/\,ok & 1 & 1.02 & 1.00 & 0.019 & ok\,/\,ok & 0 & 1.03 & 1.00 & 0.017 \\
 &  &  & AKO & crash\,/\,ok & 3 & 1.03 & 1.01 & 0.019 & crash\,/\,ok & 4 & 1.05 & 1.08 & 0.019 \\
 &  &  & Codex & crash\,/\,ok & 3 & 1.01 & 1.04 & 0.019 & crash\,/\,ok & 2 & 1.15 & 1.06 & 0.356$^\dagger$ \\
\midrule
\multirow{5}{*}{Kimi-Linear-48B} & \multirow{5}{*}{0.013} & \multirow{5}{*}{0.041} & Dr.~Kernel & crash\,/\,ok & 2 & 1.01 & 0.97 & 0.008 & crash\,/\,ok & 2 & 0.99 & 0.96 & 0.000 \\
 &  &  & Claude Code & crash\,/\,ok & 5 & 1.01 & 0.97 & 0.024 & crash\,/\,ok & 5 & 1.01 & 1.02 & 0.024 \\
 &  &  & KDA & crash\,/\,ok & 1 & 1.01 & 0.97 & 0.018 & crash\,/\,ok & 3 & 1.03 & 1.00 & 0.018 \\
 &  &  & AKO & crash\,/\,ok & 3 & 0.99 & 1.08 & 0.023 & crash\,/\,n.c. & 5 & --- & --- & --- \\
 &  &  & Codex & crash\,/\,n.c. & 5 & --- & --- & --- & crash\,/\,ok & 9 & 0.99 & 0.97 & 0.013 \\
\midrule
\multirow{5}{*}{GLA-2.7B} & \multirow{5}{*}{0.000} & \multirow{5}{*}{0.001} & Dr.~Kernel & broken\,/\,ok & 1 & 0.90 & 0.84 & 0.001 & broken\,/\,ok & 2 & 0.91 & 0.82 & 0.001 \\
 &  &  & Claude Code & broken\,/\,ok & 1 & 1.27 & 1.70 & 0.001 & broken\,/\,ok & 2 & 1.27 & 1.75 & 0.001 \\
 &  &  & KDA & ok\,/\,ok & 0 & 1.11 & 1.23 & 0.001 & ok\,/\,ok & 0 & 1.11 & 1.15 & 0.001 \\
 &  &  & AKO & broken\,/\,ok & 2 & 1.14 & 1.18 & 0.001 & broken\,/\,ok & 3 & 1.17 & 1.31 & 0.004 \\
 &  &  & Codex & crash\,/\,ok & 2 & 0.74 & 0.58 & 0.009 & crash\,/\,ok & 2 & 0.76 & 0.60 & 0.005 \\
\midrule
\multirow{5}{*}{FLUX.1-dev} & \multirow{5}{*}{0.000} & \multirow{5}{*}{0.001} & Dr.~Kernel & broken\,/\,ok & 1 & 1.01 & 0.92 & 0.013 & broken\,/\,n.c. & 1 & --- & --- & --- \\
 &  &  & Claude Code & broken\,/\,ok & 1 & 1.16 & 1.31 & 0.010 & broken\,/\,ok & 1 & 1.16 & 1.47 & 0.012 \\
 &  &  & KDA & ok\,/\,ok & 0 & 1.22 & 1.25 & 0.014 & ok\,/\,ok & 0 & 1.20 & 1.31 & 0.011 \\
 &  &  & AKO & ok\,/\,ok & 0 & 1.24 & 1.34 & 0.132$^\dagger$ & crash\,/\,ok & 2 & 1.18 & 1.49 & 0.012 \\
 &  &  & Codex & crash\,/\,n.c. & 4 & --- & --- & --- & broken\,/\,n.c. & 2 & --- & --- & --- \\
\midrule
\multirow{5}{*}{Qwen3-VL-235B} & \multirow{5}{*}{0.025} & \multirow{5}{*}{0.045} & Dr.~Kernel & crash\,/\,ok & 3 & 0.80 & 1.00 & 0.090$^\dagger$ & crash\,/\,n.c. & 3 & --- & --- & --- \\
 &  &  & Claude Code & crash\,/\,n.c. & 3 & --- & --- & --- & crash\,/\,n.c. & 4 & --- & --- & --- \\
 &  &  & KDA & crash\,/\,ok & 1 & 1.13 & 1.08 & 0.024 & crash\,/\,ok & 1 & 1.14 & 1.14 & 0.026 \\
 &  &  & AKO & crash\,/\,n.c. & 1 & --- & --- & --- & crash\,/\,n.c. & 3 & --- & --- & --- \\
 &  &  & Codex & crash\,/\,n.c. & 5 & --- & --- & --- & crash\,/\,n.c. & 5 & --- & --- & --- \\
\midrule
\multirow{5}{*}{YOLOv10n} & \multirow{5}{*}{0.000} & \multirow{5}{*}{0.001} & Dr.~Kernel & crash\,/\,n.c. & 2 & --- & --- & --- & crash\,/\,ok & 1 & 1.08 & 0.97 & 0.000 \\
 &  &  & Claude Code & broken\,/\,n.c. & 0 & --- & --- & --- & crash\,/\,n.c. & 2 & --- & --- & --- \\
 &  &  & KDA & crash\,/\,ok & 2 & 1.02 & 1.26 & 0.003 & crash\,/\,ok & 2 & 1.19 & 0.86 & 0.002 \\
 &  &  & AKO & crash\,/\,ok & 4 & 1.32 & 1.37 & 0.003 & ok\,/\,ok & 0 & 1.32 & 3.56 & 0.003 \\
 &  &  & Codex & crash\,/\,n.c. & 5 & --- & --- & --- & crash\,/\,ok & 6 & 1.15 & 1.25 & 0.040 \\
\midrule
\multirow{5}{*}{OpenFold3} & \multirow{5}{*}{0.000} & \multirow{5}{*}{0.157$^\ddagger$} & Dr.~Kernel & ok\,/\,ok & 0 & 1.06 & 1.03 & 0.121 & ok\,/\,ok & 0 & 1.04 & 1.00 & 0.121 \\
 &  &  & Claude Code & crash\,/\,ok & 6 & 1.40 & 1.45 & 0.207$^\dagger$ & crash\,/\,n.c. & 1 & --- & --- & --- \\
 &  &  & KDA & ok\,/\,ok & 0 & 1.07 & 1.66 & 0.112 & ok\,/\,ok & 0 & 0.62 & 0.84 & 0.129 \\
 &  &  & AKO & crash\,/\,ok & 1 & 1.27 & 1.72 & 0.113 & crash\,/\,ok & 5 & 1.25 & 1.48 & 0.130 \\
 &  &  & Codex & crash\,/\,n.c. & 3 & --- & --- & --- & crash\,/\,ok & 6 & 1.65 & 1.81 & 0.157 \\
\midrule
\multirow{5}{*}{BGE-M3} & \multirow{5}{*}{0.000} & \multirow{5}{*}{0.001} & Dr.~Kernel & ok\,/\,ok & 0 & 1.03 & 1.03 & 0.000 & ok\,/\,ok & 0 & 1.03 & 1.04 & 0.000 \\
 &  &  & Claude Code & ok\,/\,ok & 0 & 1.07 & 1.07 & 0.000 & ok\,/\,ok & 0 & 1.05 & 1.09 & 0.000 \\
 &  &  & KDA & ok\,/\,ok & 0 & 1.11 & 1.10 & 0.000 & ok\,/\,ok & 0 & 1.07 & 1.07 & 0.000 \\
 &  &  & AKO & ok\,/\,ok & 0 & 1.03 & 0.81 & 0.000 & ok\,/\,ok & 0 & 0.96 & 1.09 & 0.000 \\
 &  &  & Codex & ok\,/\,ok & 0 & 1.09 & 1.05 & 0.005 & ok\,/\,ok & 0 & 1.07 & 1.08 & 0.005 \\
\midrule
\multirow{5}{*}{Oasis-500M} & \multirow{5}{*}{0.000} & \multirow{5}{*}{0.001} & Dr.~Kernel & crash\,/\,ok & 2 & 1.03 & 0.84 & 0.003 & crash\,/\,ok & 3 & 1.01 & 0.85 & 0.003 \\
 &  &  & Claude Code & ok\,/\,ok & 0 & 1.02 & 0.93 & 0.035 & crash\,/\,ok & 1 & 1.13 & 0.96 & 0.034 \\
 &  &  & KDA & crash\,/\,ok & 1 & 1.01 & 1.26 & 0.040 & ok\,/\,ok & 0 & 1.14 & 1.22 & 0.039 \\
 &  &  & AKO & broken\,/\,n.c. & 0 & --- & --- & --- & crash\,/\,ok & 1 & 0.81 & 1.28 & 0.036 \\
 &  &  & Codex & crash\,/\,n.c. & 3 & --- & --- & --- & crash\,/\,ok & 4 & 1.07 & 0.84 & 0.026 \\
\bottomrule
\end{tabular}
}
\end{table}

\begin{table}[H]
  \centering
  \caption{\textbf{Sensitivity of the \textsc{MacroEval} ranking} to scaling the correctness thresholds $g_i$ and $f_i$.}
  \label{tab:e2e-sensitivity}
  \small
\begin{tabular}{@{}ll rr rr rr@{}}
\toprule
 & & \multicolumn{2}{c}{$(0.5{\times})\,g_i, f_i$} & \multicolumn{2}{c}{$(1{\times})\,g_i, f_i$} & \multicolumn{2}{c}{$(2{\times})\,g_i, f_i$} \\
\cmidrule(lr){3-4}\cmidrule(lr){5-6}\cmidrule(lr){7-8}
Agent & Mode & Score & Rank & Score & Rank & Score & Rank \\
\midrule
\multirow{3}{*}{Dr.~Kernel} & Independent & 0.48 & 4/10 & 0.63 & 4/10 & 0.74 & 4/10 \\
 & Sequential & 0.44 & 5/10 & 0.55 & 5/10 & 0.56 & 6/10 \\
 & Ind.\,+\,Seq. & 0.46 & 2/5 & 0.59 & 3/5 & 0.64 & 2/5 \\
\midrule
\multirow{3}{*}{Claude Code} & Independent & 0.24 & 8/10 & 0.35 & 8/10 & 0.47 & 7/10 \\
 & Sequential & 0.26 & 7/10 & 0.34 & 9/10 & 0.35 & 9/10 \\
 & Ind.\,+\,Seq. & 0.25 & 4/5 & 0.35 & 4/5 & 0.41 & 4/5 \\
\midrule
\multirow{3}{*}{KDA} & Independent & 0.78 & 1/10 & 1.11 & 1/10 & 1.12 & 1/10 \\
 & Sequential & 0.76 & 2/10 & 1.03 & 2/10 & 1.04 & 2/10 \\
 & Ind.\,+\,Seq. & 0.77 & 1/5 & 1.07 & 1/5 & 1.08 & 1/5 \\
\midrule
\multirow{3}{*}{AKO} & Independent & 0.28 & 6/10 & 0.42 & 7/10 & 0.43 & 8/10 \\
 & Sequential & 0.57 & 3/10 & 0.85 & 3/10 & 0.86 & 3/10 \\
 & Ind.\,+\,Seq. & 0.41 & 3/5 & 0.61 & 2/5 & 0.62 & 3/5 \\
\midrule
\multirow{3}{*}{Codex} & Independent & 0.10 & 10/10 & 0.11 & 10/10 & 0.11 & 10/10 \\
 & Sequential & 0.20 & 9/10 & 0.55 & 6/10 & 0.56 & 5/10 \\
 & Ind.\,+\,Seq. & 0.15 & 5/5 & 0.30 & 5/5 & 0.30 & 5/5 \\
\bottomrule
\end{tabular}

\end{table}

\section{Leaderboard and Submission Portal}
\label{app:leaderboard}

\begin{figure}[tbp]
  \centering
  \includegraphics[width=\linewidth]{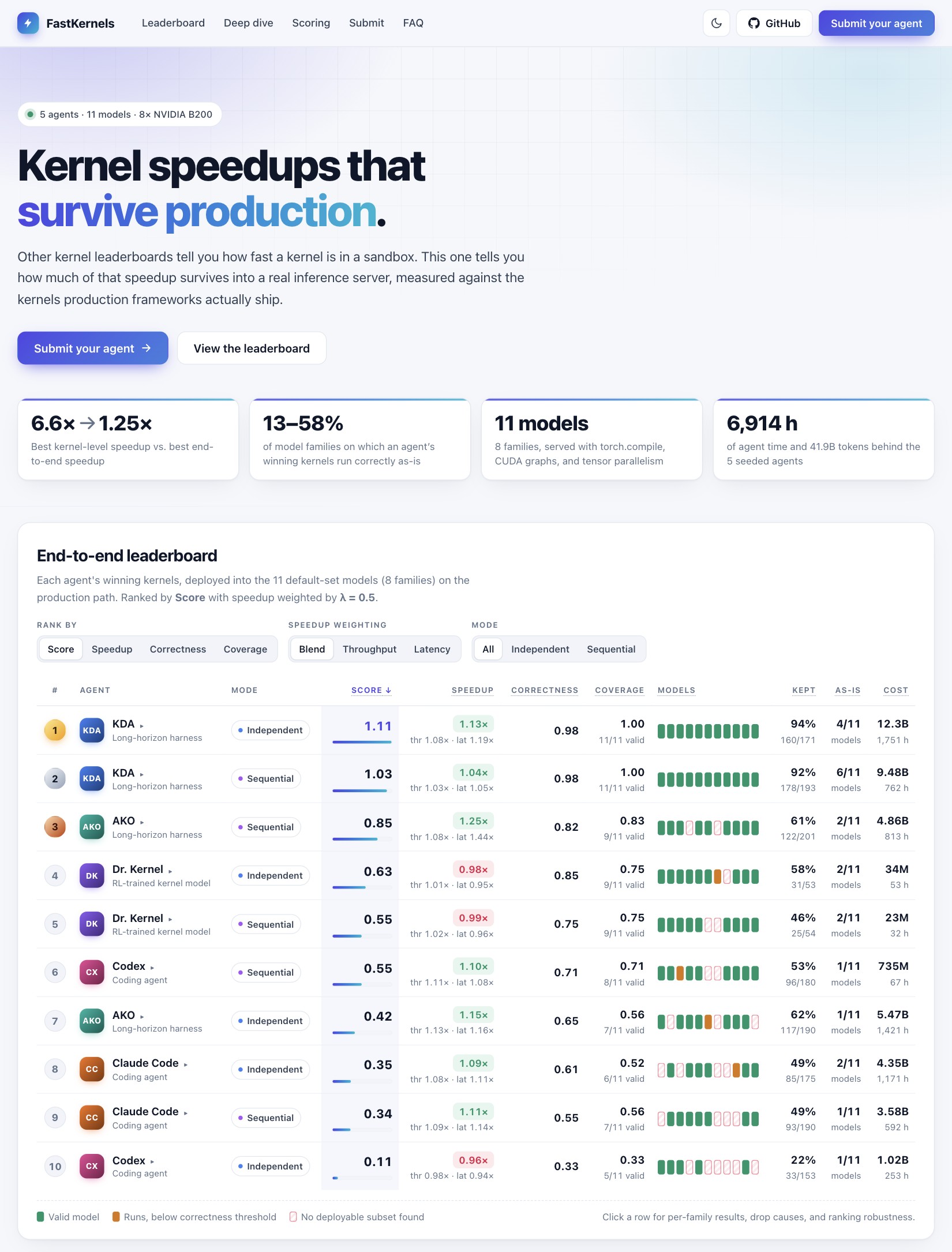}
  \caption{\textbf{Leaderboard.}}
  \label{fig:lb-overview}
\end{figure}

\begin{figure}[tbp]
  \centering
  \includegraphics[width=\linewidth]{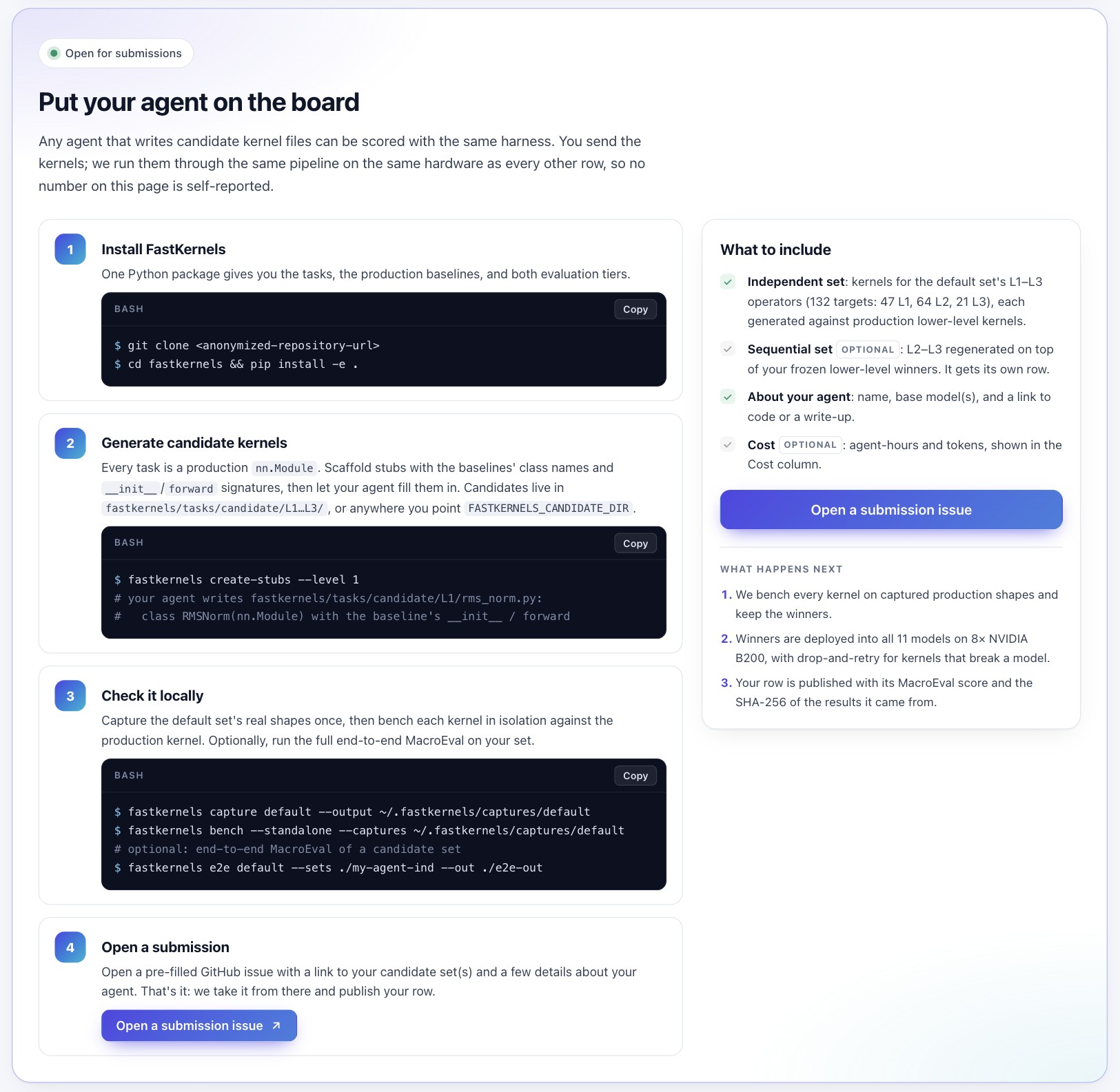}
  \ificlrfinal
  \caption{\textbf{Submission portal.}}
  \else
  \caption{\textbf{Submission portal.} Repository URL anonymized for review.}
  \fi
  \label{fig:lb-submit}
\end{figure}

Figure~\ref{fig:lb-overview} shows the leaderboard (the ranking of Table~\ref{tab:e2e-macroeval}) and Figure~\ref{fig:lb-submit} the submission portal.

\clearpage
\section{HuggingFace Transformers Coverage}
\label{app:hf_coverage}

We measure coverage empirically, by implementing every architecture in HuggingFace Transformers (commit \texttt{89b6b17}) on \fastkernels{}.
For each architecture, an agent implemented the model on top of the \fastkernels{} repository using only its existing operators: minimal patches were allowed, but no new operators or primitives.
To check correctness, we ran a few forward passes on a fixed batch whose input sizes exercise the model's relevant computation (e.g., for models with windowed attention, a sequence longer than the window): one step for non-autoregressive, embedding, and prefill-only models, and three steps (one prefill, two decodes) for autoregressive models.
We then compared the output of the \fastkernels{} implementation with that of native HF and consider the architecture supported if the implementation runs without errors and the relative L2 difference is within a \texttt{dtype}-dependent threshold: 0.01\% (FP32), 0.5\% (FP16), 2\% (BF16), 5\% (FP8), and 10--20\% (FP4); mixed-precision models use the threshold of their lowest-precision format.
Architectures without a public checkpoint use the constructor defaults of their HF configuration with random weights.

Table~\ref{tab:hf-coverage} lists 488 architectures: 468 are within their threshold, 17 run but exceed it (marked $^\dagger$), and 4 (marked $^{*}$) are flagged by the HF audit only because the HF implementation differs from the version in the reference library \fastkernels{} pins (vLLM for \nolinkurl{kimi_linear}; FLA's RWKV-7 for \nolinkurl{rwkv}; the SAM~3 release for \nolinkurl{sam3_video}; timm for \nolinkurl{timm_wrapper}); we count these as supported because \fastkernels{} supports a valid version of each.
The remaining architectures fall in three groups.
\emph{Requiring new operations} (5): \nolinkurl{autoformer}, \nolinkurl{informer}, \nolinkurl{rag}, \nolinkurl{time_series_transformer}, \nolinkurl{xlstm}.
\emph{HF reference issues} (6), where the native HF implementation itself fails, so no reference output exists: \nolinkurl{deepseek_v4}, \nolinkurl{gemma4_assistant}, \nolinkurl{granite4_vision}, \nolinkurl{minicpmv4_7}, \nolinkurl{nllb_moe}, \nolinkurl{sam_hq}.
\emph{Excluded} (5): generic wrappers that compose other architectures and have no modeling code of their own: \nolinkurl{encoder_decoder}, \nolinkurl{speech_encoder_decoder}, \nolinkurl{timm_backbone}, \nolinkurl{vision_encoder_decoder}, \nolinkurl{vision_text_dual_encoder}.
Coverage is therefore \textbf{94.6\% (472/499)}: excluding the wrappers, the architectures above their threshold (unless marked $^{*}$), those that require new operations, and those without a working HF reference count as unsupported.

\begingroup\scriptsize\setlength{\tabcolsep}{2pt}\setlength{\LTleft}{0pt}\setlength{\LTright}{0pt}
\begin{longtable}{@{}>{\raggedright\arraybackslash}p{.28\linewidth}>{\raggedright\arraybackslash}p{.45\linewidth}>{\raggedright\arraybackslash}p{.1\linewidth}>{\raggedleft\arraybackslash}p{.13\linewidth}@{}}
\caption{\textbf{HuggingFace Transformers coverage audit} (commit \texttt{89b6b17}): every architecture implemented on \fastkernels{} and the relative L2 difference (\%) of its output from native HF. $^\dagger$: above the threshold for its \texttt{dtype}. $^{*}$: flagged only because HF differs from the version \fastkernels{} pins; counted as supported.}\label{tab:hf-coverage}\\
\toprule
Architecture & HF model & \texttt{dtype} & Rel.\ L2 (\%) \\
\midrule
\endfirsthead
\multicolumn{4}{@{}l}{\tablename~\thetable\ (continued)}\\[2pt]
\toprule
Architecture & HF model & \texttt{dtype} & Rel.\ L2 (\%) \\
\midrule
\endhead
\bottomrule
\endfoot
\nolinkurl{afmoe} & \nolinkurl{arcee-ai/Trinity-Mini} & bf16 & 1.36 \\
\nolinkurl{aimv2} & \nolinkurl{apple/aimv2-large-patch14-224-lit} & bf16 & 0 \\
\nolinkurl{albert} & \nolinkurl{albert/albert-base-v2} & bf16 & 0.871 \\
\nolinkurl{align} & \nolinkurl{kakaobrain/align-base} & bf16 & 0 \\
\nolinkurl{altclip} & \nolinkurl{BAAI/AltCLIP} & bf16 & 0 \\
\nolinkurl{apertus} & \nolinkurl{swiss-ai/Apertus-8B-Instruct-2509} & bf16 & 0 \\
\nolinkurl{arcee} & \nolinkurl{arcee-ai/AFM-4.5B} & bf16 & 0.832 \\
\nolinkurl{aria} & \nolinkurl{Rhymes-AI/Aria} & bf16 & 0.518 \\
\nolinkurl{audio_spectrogram_transformer} & Constructor defaults & bf16 & 0 \\
\nolinkurl{audioflamingo3} & \nolinkurl{nvidia/audio-flamingo-3-hf} & bf16 & 0.639 \\
\nolinkurl{axk1} & \nolinkurl{skt/A.X-K1} & bf16 & 0.395 \\
\nolinkurl{axk2} & Constructor defaults & bf16 & 0.171 \\
\nolinkurl{aya_vision} & \nolinkurl{unsloth/aya-vision-8b} & bf16 & 0 \\
\nolinkurl{bamba} & \nolinkurl{ibm-ai-platform/Bamba-9B-v2} & bf16 & 0 \\
\nolinkurl{bark} & \nolinkurl{suno/bark} & bf16 & 0.913 \\
\nolinkurl{bart} & \nolinkurl{facebook/bart-large-cnn} & bf16 & 0 \\
\nolinkurl{beit} & Constructor defaults & bf16 & 0 \\
\nolinkurl{bert} & \nolinkurl{google-bert/bert-base-uncased} & bf16 & 0 \\
\nolinkurl{bert_generation} & \nolinkurl{google/bert_for_seq_generation_L-24_bbc_encoder} & bf16 & 0 \\
\nolinkurl{big_bird} & \nolinkurl{google/bigbird-roberta-base} & bf16 & 0 \\
\nolinkurl{bigbird_pegasus} & \nolinkurl{google/bigbird-pegasus-large-arxiv} & bf16 & 0 \\
\nolinkurl{biogpt} & \nolinkurl{microsoft/biogpt} & bf16 & 0 \\
\nolinkurl{bit} & Constructor defaults & bf16 & 0 \\
\nolinkurl{bitnet} & \nolinkurl{microsoft/bitnet-b1.58-2B-4T} & bf16 & 0 \\
\nolinkurl{blenderbot} & \nolinkurl{facebook/blenderbot-400M-distill} & bf16 & 0 \\
\nolinkurl{blenderbot_small} & \nolinkurl{facebook/blenderbot_small-90M} & bf16 & 0 \\
\nolinkurl{blip} & \nolinkurl{Salesforce/blip-image-captioning-base} & bf16 & 0 \\
\nolinkurl{blip_2} & \nolinkurl{Salesforce/blip2-opt-2.7b} & bf16 & 0 \\
\nolinkurl{blip_text} & Constructor defaults & bf16 & 0 \\
\nolinkurl{bloom} & \nolinkurl{bigscience/bloom} & bf16 & 0 \\
\nolinkurl{blt} & \nolinkurl{itazap/blt-1b-hf} & bf16 & 0 \\
\nolinkurl{bridgetower} & \nolinkurl{BridgeTower/bridgetower-base} & bf16 & 0 \\
\nolinkurl{bros} & \nolinkurl{jinho8345/bros-base-uncased} & bf16 & 0 \\
\nolinkurl{camembert} & \nolinkurl{almanach/camembert-base} & bf16 & 0 \\
\nolinkurl{canary} & \nolinkurl{nvidia/canary-1b-v2} & bf16 & 0 \\
\nolinkurl{canine} & Constructor defaults & fp32 & $1.95\!\times\!10^{-4}$ \\
\nolinkurl{chameleon} & \nolinkurl{nopperl/chameleon-7b-hf} & bf16 & 0.465 \\
\nolinkurl{chinese_clip} & \nolinkurl{OFA-Sys/chinese-clip-vit-base-patch16} & bf16 & 0 \\
\nolinkurl{chmv2} & Constructor defaults & bf16 & $6.32\!\times\!10^{-6}$ \\
\nolinkurl{clap} & \nolinkurl{laion/clap-htsat-unfused} & bf16 & 0 \\
\nolinkurl{clip} & \nolinkurl{openai/clip-vit-base-patch32} & bf16 & 0 \\
\nolinkurl{clipseg} & \nolinkurl{CIDAS/clipseg-rd64-refined} & bf16 & 1.35 \\
\nolinkurl{clvp} & \nolinkurl{susnato/clvp_dev} & fp32 & 0 \\
\nolinkurl{codegen} & \nolinkurl{Salesforce/codegen-350M-mono} & fp16 & 0 \\
\nolinkurl{cohere} & Constructor defaults & bf16 & 0 \\
\nolinkurl{cohere2} & Constructor defaults & bf16 & 0 \\
\nolinkurl{cohere2_moe} & \nolinkurl{CohereLabs/command-a-plus-05-2026} & bf16 & 0.528 \\
\nolinkurl{cohere2_vision} & \nolinkurl{mlx-community/command-a-vision-07-2025-4bit} & bf16 & 0.638 \\
\nolinkurl{cohere_asr} & Constructor defaults & bf16 & 0 \\
\nolinkurl{cohere_compass} & \nolinkurl{CohereLabs/North-Micro-Vision-Instruct} & bf16 & 0.789 \\
\nolinkurl{colmodernvbert} & ModernVBERT/colmodernvbert-merged + ColModernVBertConfig wrapper & bf16 & 0 \\
\nolinkurl{colpali} & vidore/colpaligemma-3b-pt-448-base + ColPaliConfig wrapper & bf16 & 1.01 \\
\nolinkurl{colqwen2} & \nolinkurl{vidore/colqwen2-v1.0-hf} & bf16 & 0.829 \\
\nolinkurl{conditional_detr} & \nolinkurl{microsoft/conditional-detr-resnet-50} & bf16 & 0 \\
\nolinkurl{convbert} & \nolinkurl{YituTech/conv-bert-base} & bf16 & 0 \\
\nolinkurl{convnext} & Constructor defaults & bf16 & 0 \\
\nolinkurl{convnextv2} & Constructor defaults & bf16 & 0 \\
\nolinkurl{cosmos3_edge} & \nolinkurl{nvidia/Cosmos3-Edge} & bf16 & 0 \\
\nolinkurl{cosmos3_omni} & \nolinkurl{nvidia/Cosmos3-Nano} & bf16 & 1.2 \\
\nolinkurl{cpmant} & \nolinkurl{openbmb/cpm-ant-10b} & bf16 & 0 \\
\nolinkurl{csm} & Constructor defaults & bf16 & 0.00206 \\
\nolinkurl{ctrl} & \nolinkurl{Salesforce/ctrl} & bf16 & 0.366 \\
\nolinkurl{cvt} & Constructor defaults & bf16 & 0 \\
\nolinkurl{cwm} & Constructor defaults & bf16 & 0.884 \\
\nolinkurl{d_fine} & \nolinkurl{ustc-community/dfine-xlarge-coco} & bf16 & 0 \\
\nolinkurl{dab_detr} & \nolinkurl{IDEA-Research/dab-detr-resnet-50} & fp32 & 0 \\
\nolinkurl{dac} & \nolinkurl{descript/dac_16khz} & bf16 & 0 \\
\nolinkurl{data2vec_audio} & Constructor defaults & bf16 & 0 \\
\nolinkurl{data2vec_text} & \nolinkurl{facebook/data2vec-text-base} & bf16 & 0 \\
\nolinkurl{data2vec_vision} & Constructor defaults & bf16 & 0 \\
\nolinkurl{dbrx} & alpindale/dbrx-instruct (mirror) & bf16 & 0 \\
\nolinkurl{deberta} & \nolinkurl{microsoft/deberta-base} & fp32 & $1.64\!\times\!10^{-4}$ \\
\nolinkurl{deberta_v2} & \nolinkurl{microsoft/deberta-v2-xlarge} & bf16 & 0 \\
\nolinkurl{decision_transformer} & \nolinkurl{edbeeching/decision-transformer-gym-hopper-medium} & bf16 & 0 \\
\nolinkurl{deepseek_ocr2} & \nolinkurl{deepseek-community/DeepSeek-OCR-2} & bf16 & 1.77 \\
\nolinkurl{deepseek_v2} & \nolinkurl{deepseek-ai/DeepSeek-V2-Lite} & bf16 & 1 \\
\nolinkurl{deepseek_v3} & \nolinkurl{deepseek-ai/DeepSeek-V3} & fp8/bf16 & 0 \\
\nolinkurl{deepseek_v32} & Constructor defaults & bf16 & 1.64 \\
\nolinkurl{deepseek_vl} & \nolinkurl{deepseek-community/deepseek-vl-1.3b-chat} & fp16 & 0.0769 \\
\nolinkurl{deepseek_vl_hybrid} & \nolinkurl{deepseek-community/deepseek-vl-7b-chat} & fp16 & 0.0693 \\
\nolinkurl{deformable_detr} & \nolinkurl{SenseTime/deformable-detr} & bf16 & 0 \\
\nolinkurl{deimv2} & \nolinkurl{harshaljanjani/DEIMv2_HGNetv2_N_COCO_Transformers} & bf16 & 0 \\
\nolinkurl{deit} & Constructor defaults & bf16 & 0 \\
\nolinkurl{depth_anything} & \nolinkurl{LiheYoung/depth-anything-small-hf} & bf16 & 0.141 \\
\nolinkurl{depth_pro} & \nolinkurl{apple/DepthPro-hf} & fp16 & 0.121 \\
\nolinkurl{detr} & \nolinkurl{facebook/detr-resnet-50-panoptic} & bf16 & 0 \\
\nolinkurl{dia} & Constructor defaults & bf16 & 0.537 \\
\nolinkurl{diffllama} & \nolinkurl{kajuma/DiffLlama-0.3B-handcut} & bf16 & 0.746 \\
\nolinkurl{diffusion_gemma} & \nolinkurl{google/diffusiongemma-26B-A4B-it} & bf16 & 81$^\dagger$ \\
\nolinkurl{dinat} & Constructor defaults & bf16 & 0 \\
\nolinkurl{dinov2} & Constructor defaults & bf16 & 0 \\
\nolinkurl{dinov2_with_registers} & Constructor defaults & bf16 & 0 \\
\nolinkurl{dinov3_convnext} & Constructor defaults & bf16 & 0 \\
\nolinkurl{dinov3_vit} & Constructor defaults & bf16 & 0 \\
\nolinkurl{distilbert} & \nolinkurl{distilbert/distilbert-base-uncased} & bf16 & 0.855 \\
\nolinkurl{doge} & Constructor defaults & bf16 & 0 \\
\nolinkurl{donut_swin} & Constructor defaults & bf16 & 0 \\
\nolinkurl{dots1} & \nolinkurl{rednote-hilab/dots.llm1.base} & bf16 & 0.239 \\
\nolinkurl{dpr} & \nolinkurl{facebook/dpr-question_encoder-single-nq-base} & bf16 & 0.972 \\
\nolinkurl{dpt} & Constructor defaults & bf16 & 0 \\
\nolinkurl{edgetam} & \nolinkurl{yonigozlan/EdgeTAM-hf} & bf16 & 0 \\
\nolinkurl{edgetam_video} & \nolinkurl{yonigozlan/EdgeTAM-hf} & bf16 & 0 \\
\nolinkurl{efficientloftr} & \nolinkurl{zju-community/efficientloftr} & bf16 & 0.0292 \\
\nolinkurl{efficientnet} & Constructor defaults & bf16 & 0 \\
\nolinkurl{electra} & \nolinkurl{google/electra-small-generator} & bf16 & 0.885 \\
\nolinkurl{emu3} & \nolinkurl{BAAI/Emu3-Chat-hf} & bf16 & 0 \\
\nolinkurl{encodec} & \nolinkurl{facebook/encodec_24khz} & bf16 & 0.0579 \\
\nolinkurl{eomt} & \nolinkurl{tue-mps/coco_panoptic_eomt_large_640} & bf16 & 0 \\
\nolinkurl{eomt_dinov3} & Constructor defaults & bf16 & 0 \\
\nolinkurl{ernie} & \nolinkurl{nghuyong/ernie-1.0-base-zh} & bf16 & 0 \\
\nolinkurl{ernie4_5} & \nolinkurl{baidu/ERNIE-4.5-0.3B-PT} & bf16 & 0 \\
\nolinkurl{ernie4_5_moe} & \nolinkurl{baidu/ERNIE-4.5-21B-A3B-PT} & bf16 & 0.542 \\
\nolinkurl{ernie4_5_vl_moe} & \nolinkurl{baidu/ERNIE-4.5-VL-28B-A3B-PT} & bf16 & 0.42 \\
\nolinkurl{esm} & \nolinkurl{facebook/esm-1b} & bf16 & 1.73 \\
\nolinkurl{esmc} & \nolinkurl{biohub/ESMC-6B-hf} & bf16 & 0 \\
\nolinkurl{esmfold} & \nolinkurl{facebook/esmfold_v1} & bf16 & $4.80\!\times\!10^{-6}$ \\
\nolinkurl{esmfold2} & \nolinkurl{biohub/ESMFold2-hf} & fp32 & $2.37\!\times\!10^{-5}$ \\
\nolinkurl{eurobert} & \nolinkurl{EuroBERT/EuroBERT-210m} & bf16 & 0 \\
\nolinkurl{evolla} & \nolinkurl{westlake-repl/Evolla-10B-DPO-hf} & bf16 & 0 \\
\nolinkurl{exaone4} & \nolinkurl{LGAI-EXAONE/EXAONE-4.0-32B} & bf16 & 0 \\
\nolinkurl{exaone4_5} & \nolinkurl{LGAI-EXAONE/EXAONE-4.5-33B} & bf16 & 1.27 \\
\nolinkurl{exaone_moe} & \nolinkurl{LGAI-EXAONE/K-EXAONE-236B-A23B} & bf16 & 1.65 \\
\nolinkurl{falcon} & \nolinkurl{tiiuae/falcon-7b-instruct} & bf16 & 1.45 \\
\nolinkurl{falcon_h1} & \nolinkurl{tiiuae/Falcon-H1-7B-Instruct} & bf16 & 1.25 \\
\nolinkurl{falcon_mamba} & \nolinkurl{tiiuae/falcon-mamba-7b-instruct} & bf16 & 0 \\
\nolinkurl{fast_vlm} & \nolinkurl{KamilaMila/FastVLM-0.5B} & bf16 & 0.69 \\
\nolinkurl{fastspeech2_conformer} & \nolinkurl{espnet/fastspeech2_conformer_with_hifigan} & fp32 & $1.91\!\times\!10^{-4}$ \\
\nolinkurl{flaubert} & \nolinkurl{flaubert/flaubert_base_uncased} & bf16 & 1.09 \\
\nolinkurl{flava} & \nolinkurl{facebook/flava-full} & bf16 & 0 \\
\nolinkurl{flex_olmo} & \nolinkurl{allenai/FlexOlmo-7x7B-1T} & bf16 & 0.86 \\
\nolinkurl{florence2} & \nolinkurl{florence-community/Florence-2-large} & fp16 & 0 \\
\nolinkurl{fnet} & \nolinkurl{google/fnet-base} & fp32 & $1.70\!\times\!10^{-4}$ \\
\nolinkurl{focalnet} & Constructor defaults & bf16 & 0 \\
\nolinkurl{fsmt} & \nolinkurl{facebook/wmt19-ru-en} & bf16 & 0 \\
\nolinkurl{fun_asr_nano} & \nolinkurl{FunAudioLLM/Fun-ASR-Nano-2512-hf} & bf16 & 0 \\
\nolinkurl{funnel} & \nolinkurl{funnel-transformer/small} & bf16 & 0 \\
\nolinkurl{fuyu} & \nolinkurl{adept/fuyu-8b} & bf16 & 0 \\
\nolinkurl{gemma} & Constructor defaults & bf16 & 0.451 \\
\nolinkurl{gemma2} & Constructor defaults & bf16 & 0.389 \\
\nolinkurl{gemma3} & Pinned Gemma3 4B recipe & bf16 & 0 \\
\nolinkurl{gemma3n} & Constructor defaults & bf16 & 0 \\
\nolinkurl{gemma4} & \nolinkurl{google/gemma-4-e2b-it} & bf16 & 0 \\
\nolinkurl{gemma4_unified} & \nolinkurl{google/gemma-4-12B-it} & bf16 & 5.88$^\dagger$ \\
\nolinkurl{gemma4_unified_assistant} & \nolinkurl{google/gemma-4-e2b-it-assistant} & bf16 & 0 \\
\nolinkurl{git} & \nolinkurl{microsoft/git-base-coco} & bf16 & 0 \\
\nolinkurl{glm} & \nolinkurl{THUDM/glm-4-9b-chat} & bf16 & 0.905 \\
\nolinkurl{glm4} & Constructor defaults & bf16 & 0.761 \\
\nolinkurl{glm46v} & \nolinkurl{zai-org/GLM-4.1V-9B-Thinking} & bf16 & 0 \\
\nolinkurl{glm4_moe} & \nolinkurl{zai-org/GLM-4.5} & bf16 & 0.212 \\
\nolinkurl{glm4_moe_lite} & \nolinkurl{zai-org/GLM-4.7-Flash} & bf16 & 0.99 \\
\nolinkurl{glm4v} & \nolinkurl{zai-org/GLM-4.1V-9B-Thinking} & bf16 & 0 \\
\nolinkurl{glm4v_moe} & \nolinkurl{zai-org/GLM-4.5V} & bf16 & 0.143 \\
\nolinkurl{glm5_next} & \nolinkurl{zai-org/GLM-5.3-Flash} & fp8/bf16 & 19.5$^\dagger$ \\
\nolinkurl{glm_image} & \nolinkurl{zai-org/GLM-Image} & bf16 & 0 \\
\nolinkurl{glm_moe_dsa} & \nolinkurl{zai-org/GLM-5} & bf16 & 1.92 \\
\nolinkurl{glm_ocr} & \nolinkurl{zai-org/GLM-OCR} & bf16 & 0 \\
\nolinkurl{glmasr} & \nolinkurl{zai-org/GLM-ASR-Nano-2512} & bf16 & 0 \\
\nolinkurl{glpn} & \nolinkurl{vinvino02/glpn-kitti} & bf16 & 0.062 \\
\nolinkurl{got_ocr2} & \nolinkurl{stepfun-ai/GOT-OCR-2.0-hf} & bf16 & 0 \\
\nolinkurl{gpt2} & \nolinkurl{openai-community/gpt2} & bf16 & 0 \\
\nolinkurl{gpt_bigcode} & \nolinkurl{bigcode/gpt_bigcode-santacoder} & bf16 & 0 \\
\nolinkurl{gpt_neo} & \nolinkurl{EleutherAI/gpt-neo-1.3B} & bf16 & 0 \\
\nolinkurl{gpt_neox} & \nolinkurl{EleutherAI/gpt-neox-20b} & fp16 & 0.0522 \\
\nolinkurl{gpt_neox_japanese} & \nolinkurl{abeja/gpt-neox-japanese-2.7b} & bf16 & 0.687 \\
\nolinkurl{gpt_oss} & \nolinkurl{openai/gpt-oss-20b} & mxfp4/bf16 & 0 \\
\nolinkurl{gptj} & \nolinkurl{EleutherAI/gpt-j-6B} & bf16 & 0 \\
\nolinkurl{granite} & \nolinkurl{ibm-granite/granite-3.0-8b-base} & bf16 & 0.275 \\
\nolinkurl{granite_speech} & \nolinkurl{ibm-granite/granite-speech-3.3-2b} & bf16 & 0 \\
\nolinkurl{granite_speech5} & \nolinkurl{ibm-granite/granite-speech-5.0-470m-turboctc} & bf16 & 0 \\
\nolinkurl{granite_speech_plus} & \nolinkurl{ibm-granite/granite-speech-4.1-2b-plus} & bf16 & 0 \\
\nolinkurl{granite_swa} & \nolinkurl{ibm-granite/granite-swash-2b} & bf16 & 0 \\
\nolinkurl{granitemoe} & \nolinkurl{ibm/PowerMoE-3b} & bf16 & 1.69 \\
\nolinkurl{granitemoe_swa} & \nolinkurl{ibm-granite/granite-swash-3b-a600m} & bf16 & 9.34$^\dagger$ \\
\nolinkurl{granitemoehybrid} & \nolinkurl{ibm-granite/granite-4.0-h-tiny} & bf16 & 0 \\
\nolinkurl{granitemoeshared} & \nolinkurl{ibm-research/moe-7b-1b-active-shared-experts} & bf16 & 0 \\
\nolinkurl{grounding_dino} & \nolinkurl{IDEA-Research/grounding-dino-tiny} & fp32 & $6.4\!\times\!10^{-5}$ \\
\nolinkurl{groupvit} & \nolinkurl{nvidia/groupvit-gcc-yfcc} & bf16 & 0 \\
\nolinkurl{helium} & \nolinkurl{kyutai/helium-1-preview} & bf16 & 0 \\
\nolinkurl{hgnet_v2} & Constructor defaults & bf16 & 0 \\
\nolinkurl{hiera} & Constructor defaults & bf16 & 0 \\
\nolinkurl{higgs_audio_v2} & \nolinkurl{eustlb/higgs-audio-v2-generation-3B-base} & bf16 & 0 \\
\nolinkurl{higgs_audio_v2_tokenizer} & \nolinkurl{eustlb/higgs-audio-v2-tokenizer} & bf16 & 0 \\
\nolinkurl{hrm_text} & \nolinkurl{sapientinc/HRM-Text-1B} & bf16 & 0 \\
\nolinkurl{hubert} & \nolinkurl{facebook/hubert-large-ls960-ft} & bf16 & 0 \\
\nolinkurl{hunyuan_v1_dense} & \nolinkurl{tencent/Hunyuan-0.5B-Pretrain} & bf16 & 0 \\
\nolinkurl{hunyuan_v1_moe} & \nolinkurl{tencent/Hunyuan-A13B-Instruct} & bf16 & 0 \\
\nolinkurl{hunyuan_vl} & \nolinkurl{tencent/HunyuanOCR} & bf16 & 0.724 \\
\nolinkurl{hy_v3} & \nolinkurl{tencent/Hy3-preview} & bf16 & 0.712 \\
\nolinkurl{hy_v4} & \nolinkurl{tencent/Hy4-preview} & bf16 & 2.55$^\dagger$ \\
\nolinkurl{hyperclovax} & \nolinkurl{naver-hyperclovax/HyperCLOVAX-SEED-Think-14B} & bf16 & 0 \\
\nolinkurl{hyperclovax_vision_v2} & \nolinkurl{naver-hyperclovax/HyperCLOVAX-SEED-Think-32B} & bf16 & 0.78 \\
\nolinkurl{ibert} & \nolinkurl{kssteven/ibert-roberta-base} & bf16 & 1.33 \\
\nolinkurl{idefics} & \nolinkurl{HuggingFaceM4/idefics-9b} & bf16 & 0.651 \\
\nolinkurl{idefics2} & \nolinkurl{HuggingFaceM4/idefics2-8b-base} & bf16 & 0.749 \\
\nolinkurl{idefics3} & \nolinkurl{HuggingFaceM4/Idefics3-8B-Llama3} & bf16 & 1.12 \\
\nolinkurl{ijepa} & Constructor defaults & bf16 & 0 \\
\nolinkurl{imagegpt} & \nolinkurl{openai/imagegpt-small} & bf16 & 0 \\
\nolinkurl{inkling} & \nolinkurl{thinkingmachines/Inkling} & bf16 & 4.15$^\dagger$ \\
\nolinkurl{instructblip} & \nolinkurl{Salesforce/instructblip-vicuna-7b} & bf16 & 0.539 \\
\nolinkurl{instructblipvideo} & \nolinkurl{Salesforce/instructblip-vicuna-7b} & bf16 & 0.483 \\
\nolinkurl{internvl} & \nolinkurl{OpenGVLab/InternVL3-1B-hf} & bf16 & 0 \\
\nolinkurl{jais2} & Constructor defaults & bf16 & 0 \\
\nolinkurl{jamba} & \nolinkurl{ai21labs/Jamba-v0.1} & bf16 & 1.02 \\
\nolinkurl{janus} & \nolinkurl{deepseek-community/Janus-Pro-1B} & bf16 & 0 \\
\nolinkurl{jetmoe} & \nolinkurl{jetmoe/jetmoe-8b} & bf16 & 0.341 \\
\nolinkurl{jina_embeddings_v3} & \nolinkurl{jinaai/jina-embeddings-v3-hf} & bf16 & 1.5 \\
\nolinkurl{kimi_k25} & \nolinkurl{moonshotai/Kimi-K2.6} & int4/bf16 & 7.61 \\
\nolinkurl{kimi_linear}$^{*}$ & \nolinkurl{moonshotai/Kimi-Linear-48B-A3B-Instruct} & bf16 & 16.9$^\dagger$ \\
\nolinkurl{kosmos2} & \nolinkurl{microsoft/kosmos-2-patch14-224} & bf16 & 0 \\
\nolinkurl{kosmos2_5} & \nolinkurl{microsoft/kosmos-2.5} & bf16 & 0 \\
\nolinkurl{kyutai_speech_to_text} & \nolinkurl{kyutai/stt-2.6b-en-trfs} & bf16 & 0 \\
\nolinkurl{laguna} & \nolinkurl{poolside/laguna-XS.2} & bf16 & 0 \\
\nolinkurl{lasr} & Constructor defaults & bf16 & 0.727 \\
\nolinkurl{layoutlm} & \nolinkurl{microsoft/layoutlm-base-uncased} & bf16 & 0 \\
\nolinkurl{layoutlmv2} & \nolinkurl{microsoft/layoutlmv2-base-uncased} & bf16 & 0 \\
\nolinkurl{layoutlmv3} & \nolinkurl{microsoft/layoutlmv3-base} & bf16 & 0 \\
\nolinkurl{led} & \nolinkurl{allenai/led-large-16384-arxiv} & bf16 & 0 \\
\nolinkurl{levit} & Constructor defaults & bf16 & 0 \\
\nolinkurl{lfm2} & \nolinkurl{LiquidAI/LFM2-1.2B} & bf16 & 1.83 \\
\nolinkurl{lfm2_moe} & \nolinkurl{LiquidAI/LFM2-8B-A1B} & bf16 & 1.45 \\
\nolinkurl{lfm2_vl} & \nolinkurl{LiquidAI/LFM2-VL-1.6B} & bf16 & 0.787 \\
\nolinkurl{lightglue} & \nolinkurl{ETH-CVG/lightglue_superpoint} & bf16 & 0 \\
\nolinkurl{lighton_ocr} & \nolinkurl{lightonai/LightOnOCR-1B-1025} & bf16 & 0 \\
\nolinkurl{lilt} & \nolinkurl{SCUT-DLVCLab/lilt-roberta-en-base} & bf16 & 0 \\
\nolinkurl{llama} & Constructor defaults & bf16 & 0.746 \\
\nolinkurl{llama4} & Constructor defaults & bf16 & 0 \\
\nolinkurl{llava} & \nolinkurl{llava-hf/llava-1.5-7b-hf} & fp16 & 0.0932 \\
\nolinkurl{llava_next} & \nolinkurl{llava-hf/llava-v1.6-mistral-7b-hf} & fp16 & 0.0961 \\
\nolinkurl{llava_next_video} & \nolinkurl{llava-hf/LLaVA-NeXT-Video-7B-hf} & bf16 & 1.34 \\
\nolinkurl{llava_onevision} & \nolinkurl{llava-hf/llava-onevision-qwen2-7b-ov-hf} & fp16 & 0.0756 \\
\nolinkurl{longcat_flash} & \nolinkurl{meituan-longcat/LongCat-Flash-Chat} & bf16 & 0.725 \\
\nolinkurl{longformer} & \nolinkurl{allenai/longformer-base-4096} & bf16 & 0 \\
\nolinkurl{longt5} & \nolinkurl{Stancld/longt5-tglobal-large-16384-pubmed-3k_steps} & bf16 & 0 \\
\nolinkurl{luke} & \nolinkurl{studio-ousia/luke-base} & bf16 & 0 \\
\nolinkurl{lw_detr} & \nolinkurl{AnnaZhang/lwdetr_small_60e_coco} & bf16 & 0 \\
\nolinkurl{lxmert} & \nolinkurl{unc-nlp/lxmert-base-uncased} & bf16 & 0 \\
\nolinkurl{m2m_100} & \nolinkurl{facebook/m2m100_418M} & bf16 & 0 \\
\nolinkurl{mamba} & \nolinkurl{state-spaces/mamba-130m-hf} & bf16 & 0 \\
\nolinkurl{mamba2} & \nolinkurl{mistralai/Mamba-Codestral-7B-v0.1} & bf16 & 0 \\
\nolinkurl{marian} & \nolinkurl{Helsinki-NLP/opus-mt-en-de} & bf16 & 0 \\
\nolinkurl{markuplm} & \nolinkurl{microsoft/markuplm-base} & fp16 & 0 \\
\nolinkurl{mask2former} & \nolinkurl{facebook/mask2former-swin-small-coco-instance} & bf16 & 0 \\
\nolinkurl{maskformer} & \nolinkurl{facebook/maskformer-swin-base-ade} & fp32 & 0 \\
\nolinkurl{maskformer_swin} & Constructor defaults & fp32 & $2.82\!\times\!10^{-5}$ \\
\nolinkurl{mbart} & \nolinkurl{facebook/mbart-large-en-ro} & bf16 & 0 \\
\nolinkurl{megatron_bert} & Constructor defaults & bf16 & 0 \\
\nolinkurl{mellum} & \nolinkurl{JetBrains/Mellum2-12B-A2.5B-Base} & bf16 & 0.817 \\
\nolinkurl{metaclip_2} & \nolinkurl{facebook/metaclip-2-worldwide-huge-quickgelu} & bf16 & 0 \\
\nolinkurl{mgp_str} & \nolinkurl{alibaba-damo/mgp-str-base} & bf16 & 0.365 \\
\nolinkurl{mimi} & \nolinkurl{kyutai/mimi} & bf16 & 0 \\
\nolinkurl{mimo_v2_flash} & \nolinkurl{XiaomiMiMo/MiMo-V2-Flash} & fp8/bf16 & 12.2$^\dagger$ \\
\nolinkurl{minicpm3} & \nolinkurl{openbmb/MiniCPM3-4B} & bf16 & 0 \\
\nolinkurl{minicpmv4_6} & \nolinkurl{openbmb/MiniCPM-V-4_6} & bf16 & 1.27 \\
\nolinkurl{minimax} & \nolinkurl{MiniMaxAI/MiniMax-Text-01-hf} & bf16 & 0.255 \\
\nolinkurl{minimax_m2} & \nolinkurl{MiniMaxAI/MiniMax-M2} & fp8/bf16 & $8.69\!\times\!10^{-5}$ \\
\nolinkurl{minimax_m3_vl} & \nolinkurl{MiniMaxAI/MiniMax-M3} & bf16 & 3.4$^\dagger$ \\
\nolinkurl{ministral} & \nolinkurl{mistralai/Ministral-8B-Instruct-2410} & bf16 & 0.768 \\
\nolinkurl{ministral3} & \nolinkurl{mistralai/Ministral-3-8B-Base-2512} & bf16 & 0.649 \\
\nolinkurl{mistral} & \nolinkurl{mistralai/Mistral-7B-v0.1} & bf16 & 0.81 \\
\nolinkurl{mistral3} & \nolinkurl{mistralai/Mistral-Small-3.1-24B-Instruct-2503} & bf16 & 0 \\
\nolinkurl{mistral4} & \nolinkurl{mistralai/Mistral-Small-4-119B-2603} & fp8/bf16/fp32 & 0 \\
\nolinkurl{mixtral} & \nolinkurl{mistralai/Mixtral-8x7B-v0.1} & bf16 & 0 \\
\nolinkurl{mlcd} & \nolinkurl{DeepGlint-AI/mlcd-vit-bigG-patch14-448} & bf16 & 1.24 \\
\nolinkurl{mllama} & Constructor defaults & bf16 & 0 \\
\nolinkurl{mm_grounding_dino} & \nolinkurl{openmmlab-community/mm_grounding_dino_tiny_o365v1_goldg_v3det} & fp32 & $4.4\!\times\!10^{-5}$ \\
\nolinkurl{mobilebert} & \nolinkurl{google/mobilebert-uncased} & bf16 & 0 \\
\nolinkurl{mobilenet_v1} & Constructor defaults & bf16 & 0 \\
\nolinkurl{mobilenet_v2} & Constructor defaults & bf16 & 0 \\
\nolinkurl{mobilevit} & Constructor defaults & bf16 & 0 \\
\nolinkurl{mobilevitv2} & Constructor defaults & bf16 & 0.701 \\
\nolinkurl{modernbert} & \nolinkurl{answerdotai/ModernBERT-base} & bf16 & 0 \\
\nolinkurl{modernbert_decoder} & \nolinkurl{blab-jhu/test-32m-dec} & bf16 & 0 \\
\nolinkurl{modernvbert} & \nolinkurl{ModernVBERT/modernvbert} & bf16 & 1.3 \\
\nolinkurl{moonshine} & \nolinkurl{UsefulSensors/moonshine-tiny} & bf16 & 0 \\
\nolinkurl{moonshine_streaming} & \nolinkurl{UsefulSensors/moonshine-streaming-tiny} & bf16 & 0 \\
\nolinkurl{moshi} & \nolinkurl{kmhf/hf-moshiko} & bf16 & 0 \\
\nolinkurl{mpnet} & \nolinkurl{microsoft/mpnet-base} & bf16 & 0 \\
\nolinkurl{mpt} & Constructor defaults & bf16 & 0 \\
\nolinkurl{mra} & \nolinkurl{uw-madison/mra-base-512-4} & fp32 & $4.1\!\times\!10^{-5}$ \\
\nolinkurl{mt5} & \nolinkurl{google/mt5-small} & bf16 & 0 \\
\nolinkurl{muse_glimmer} & \nolinkurl{meta-models/Muse-Glimmer-30B} & bf16 & 12.2$^\dagger$ \\
\nolinkurl{muse_glimmer_assistant} & \nolinkurl{meta-models/Muse-Glimmer-30B-assistant} & bf16 & 0.413 \\
\nolinkurl{musicflamingo} & \nolinkurl{nvidia/music-flamingo-2601-hf} & bf16 & 0 \\
\nolinkurl{musicgen} & \nolinkurl{facebook/musicgen-small} & bf16 & 0 \\
\nolinkurl{musicgen_melody} & \nolinkurl{facebook/musicgen-melody} & bf16 & 0 \\
\nolinkurl{mvp} & \nolinkurl{RUCAIBox/mvp} & bf16 & 0 \\
\nolinkurl{nanochat} & Constructor defaults & bf16 & 0 \\
\nolinkurl{nemotron} & \nolinkurl{nvidia/Minitron-4B-Base} & bf16 & 0.843 \\
\nolinkurl{nemotron3_5_asr} & \nolinkurl{nvidia/nemotron-3.5-asr-streaming-0.6b} & bf16 & 0.0547 \\
\nolinkurl{nemotron3_diarization} & \nolinkurl{nvidia/Nemotron-3-Diarization} & bf16 & 0 \\
\nolinkurl{nemotron_asr_streaming} & \nolinkurl{nvidia/nemotron-speech-streaming-en-0.6b} & bf16 & 0.569 \\
\nolinkurl{nemotron_h} & \nolinkurl{nvidia/Nemotron-H-8B-Reasoning-128K} & bf16 & 1.53 \\
\nolinkurl{nemotron_h_omni} & \nolinkurl{nvidia/Nemotron-3-Nano-Omni-30B-A3B-Reasoning-BF16} & bf16 & 13.2$^\dagger$ \\
\nolinkurl{neomme} & \nolinkurl{Hcompany/NeoMME-260M} & bf16 & 0.118 \\
\nolinkurl{neucodec} & Constructor defaults & bf16 & 16.1$^\dagger$ \\
\nolinkurl{nomic_bert} & Constructor defaults & bf16 & 0 \\
\nolinkurl{nystromformer} & \nolinkurl{uw-madison/nystromformer-512} & bf16 & 0 \\
\nolinkurl{olmo} & \nolinkurl{allenai/OLMo-7B-hf} & bf16 & 0.711 \\
\nolinkurl{olmo2} & \nolinkurl{allenai/OLMo-2-0425-1B} & bf16 & 0 \\
\nolinkurl{olmo3} & \nolinkurl{allenai/Olmo-3-7B-Instruct} & bf16 & 0 \\
\nolinkurl{olmo_hybrid} & \nolinkurl{allenai/Olmo-Hybrid-7B} & bf16 & 0.639 \\
\nolinkurl{olmoe} & \nolinkurl{allenai/OLMoE-1B-7B-0924} & bf16 & 0.557 \\
\nolinkurl{omdet_turbo} & \nolinkurl{omlab/omdet-turbo-swin-tiny-hf} & bf16 & 0 \\
\nolinkurl{oneformer} & \nolinkurl{shi-labs/oneformer_ade20k_swin_tiny} & bf16 & 0.302 \\
\nolinkurl{openai} & \nolinkurl{openai-community/openai-gpt} & bf16 & 0 \\
\nolinkurl{openai_privacy_filter} & \nolinkurl{openai/privacy-filter} & bf16 & 1.3 \\
\nolinkurl{opt} & \nolinkurl{facebook/opt-350m} & fp16 & 0 \\
\nolinkurl{ovis2} & \nolinkurl{thisisiron/Ovis2-2B-hf} & bf16 & 0.702 \\
\nolinkurl{owlv2} & \nolinkurl{google/owlv2-base-patch16-ensemble} & bf16 & 0 \\
\nolinkurl{owlvit} & \nolinkurl{google/owlvit-base-patch32} & bf16 & 0 \\
\nolinkurl{paddleocr_vl} & \nolinkurl{PaddlePaddle/PaddleOCR-VL} & bf16 & 0 \\
\nolinkurl{paligemma} & Constructor defaults & bf16 & 0.0153 \\
\nolinkurl{parakeet} & \nolinkurl{nvidia/parakeet-ctc-1.1b} & bf16 & 0 \\
\nolinkurl{patchtsmixer} & \nolinkurl{ibm/patchtsmixer-etth1-pretrain} & bf16 & 0 \\
\nolinkurl{patchtst} & \nolinkurl{namctin/patchtst_etth1_pretrain} & bf16 & 0.132 \\
\nolinkurl{pe_audio} & \nolinkurl{facebook/pe-av-large} & bf16 & 0 \\
\nolinkurl{pe_audio_video} & \nolinkurl{facebook/pe-av-large} & bf16 & 0 \\
\nolinkurl{pe_video} & \nolinkurl{facebook/pe-av-large} & bf16 & 0 \\
\nolinkurl{pegasus} & \nolinkurl{google/pegasus-xsum} & bf16 & 0 \\
\nolinkurl{pegasus_x} & \nolinkurl{google/pegasus-x-large} & bf16 & 0 \\
\nolinkurl{perceiver} & Constructor defaults & bf16 & 0 \\
\nolinkurl{perception_lm} & \nolinkurl{facebook/Perception-LM-1B} & bf16 & 0 \\
\nolinkurl{persimmon} & \nolinkurl{adept/persimmon-8b-base} & bf16 & 0 \\
\nolinkurl{phi} & \nolinkurl{microsoft/phi-1} & bf16 & 0 \\
\nolinkurl{phi3} & \nolinkurl{microsoft/Phi-3-mini-4k-instruct} & bf16 & 0.929 \\
\nolinkurl{phi4_multimodal (speech)} & \nolinkurl{microsoft/Phi-4-multimodal-instruct} & bf16 & 0 \\
\nolinkurl{phimoe} & \nolinkurl{microsoft/Phi-3.5-MoE-instruct} & bf16 & 0 \\
\nolinkurl{pi0} & Constructor defaults & bf16 & 0.282 \\
\nolinkurl{pix2struct} & \nolinkurl{google/pix2struct-textcaps-base} & bf16 & 0 \\
\nolinkurl{pixio} & Constructor defaults & bf16 & 0 \\
\nolinkurl{pixtral} & \nolinkurl{mistral-labs/pixtral-12b} & bf16 & 0 \\
\nolinkurl{plbart} & \nolinkurl{uclanlp/plbart-base} & bf16 & 0 \\
\nolinkurl{poolformer} & Constructor defaults & bf16 & 0 \\
\nolinkurl{pop2piano} & \nolinkurl{sweetcocoa/pop2piano} & bf16 & 0 \\
\nolinkurl{pp_doclayout_v2} & \nolinkurl{PaddlePaddle/PP-DocLayoutV2_safetensors} & bf16 & 0 \\
\nolinkurl{pp_doclayout_v3} & \nolinkurl{PaddlePaddle/PP-DocLayoutV3_safetensors} & bf16 & 0 \\
\nolinkurl{pp_formulanet} & \nolinkurl{PaddlePaddle/PP-FormulaNet_plus-L_safetensors} & bf16 & 0 \\
\nolinkurl{pp_lcnet} & Constructor defaults & bf16 & 0 \\
\nolinkurl{pp_lcnet_v3} & Constructor defaults & bf16 & 0 \\
\nolinkurl{pp_lcnet_v4} & Constructor defaults & bf16 & 0 \\
\nolinkurl{pp_ocrv5_mobile_det} & \nolinkurl{PaddlePaddle/PP-OCRv5_mobile_det_safetensors} & bf16 & 0 \\
\nolinkurl{pp_ocrv5_mobile_rec} & \nolinkurl{PaddlePaddle/PP-OCRv5_mobile_rec_safetensors} & bf16 & 0 \\
\nolinkurl{pp_ocrv5_server_det} & \nolinkurl{PaddlePaddle/PP-OCRv5_server_det_safetensors} & bf16 & 0 \\
\nolinkurl{pp_ocrv5_server_rec} & \nolinkurl{PaddlePaddle/PP-OCRv5_server_rec_safetensors} & bf16 & 0 \\
\nolinkurl{pp_ocrv6_medium_det} & \nolinkurl{PaddlePaddle/PP-OCRv6_medium_det_safetensors} & bf16 & 0 \\
\nolinkurl{pp_ocrv6_small_det} & \nolinkurl{PaddlePaddle/PP-OCRv6_small_det_safetensors} & bf16 & 0 \\
\nolinkurl{pp_ocrv6_small_rec} & \nolinkurl{PaddlePaddle/PP-OCRv6_small_rec_safetensors} & bf16 & 0 \\
\nolinkurl{pp_ocrv6_tiny_rec} & \nolinkurl{PaddlePaddle/PP-OCRv6_tiny_rec_safetensors} & bf16 & 0 \\
\nolinkurl{prompt_depth_anything} & \nolinkurl{depth-anything/prompt-depth-anything-vits-hf} & bf16 & 0.0479 \\
\nolinkurl{prophetnet} & \nolinkurl{microsoft/prophetnet-large-uncased} & bf16 & 0 \\
\nolinkurl{pvt} & Constructor defaults & bf16 & 0 \\
\nolinkurl{pvt_v2} & Constructor defaults & bf16 & 0 \\
\nolinkurl{qianfan_ocr} & \nolinkurl{baidu/Qianfan-OCR} & bf16 & 0.727 \\
\nolinkurl{qwen2} & \nolinkurl{Qwen/Qwen2-7B} & bf16 & 0 \\
\nolinkurl{qwen2_5_omni} & \nolinkurl{Qwen/Qwen2.5-Omni-7B} & bf16 & 0.005 \\
\nolinkurl{qwen2_5_vl} & \nolinkurl{Qwen/Qwen2.5-VL-7B-Instruct} & bf16 & 0.789 \\
\nolinkurl{qwen2_audio} & \nolinkurl{Qwen/Qwen2-Audio-7B} & bf16 & 0.683 \\
\nolinkurl{qwen2_moe} & \nolinkurl{Qwen/Qwen1.5-MoE-A2.7B-Chat} & bf16 & 0.56 \\
\nolinkurl{qwen2_vl} & \nolinkurl{Qwen/Qwen2-VL-7B-Instruct} & bf16 & 0.767 \\
\nolinkurl{qwen3} & \nolinkurl{Qwen/Qwen3-8B} & bf16 & 0 \\
\nolinkurl{qwen3_5} & \nolinkurl{Qwen/Qwen3.5-27B} & bf16 & 0.393 \\
\nolinkurl{qwen3_5_moe} & \nolinkurl{Qwen/Qwen3.5-35B-A3B} & bf16 & 0.353 \\
\nolinkurl{qwen3_asr} & \nolinkurl{Qwen/Qwen3-ASR-1.7B-hf} & bf16 & 0.827 \\
\nolinkurl{qwen3_moe} & \nolinkurl{Qwen/Qwen3-30B-A3B} & bf16 & 0.897 \\
\nolinkurl{qwen3_next} & \nolinkurl{Qwen/Qwen3-Next-80B-A3B-Instruct} & bf16 & 0.758 \\
\nolinkurl{qwen3_omni_moe} & \nolinkurl{Qwen/Qwen3-Omni-30B-A3B-Instruct} & bf16 & 0 \\
\nolinkurl{qwen3_vl} & \nolinkurl{Qwen/Qwen3-VL-8B-Instruct} & bf16 & 0.994 \\
\nolinkurl{qwen3_vl_moe} & \nolinkurl{Qwen/Qwen3-VL-30B-A3B-Instruct} & bf16 & 0.939 \\
\nolinkurl{qwen4_exp} & \nolinkurl{Qwen/Qwen3.8-Flash-Next} & bf16 & 12.7$^\dagger$ \\
\nolinkurl{radio} & Constructor defaults & bf16 & 0 \\
\nolinkurl{recurrent_gemma} & Constructor defaults & bf16 & 0.334 \\
\nolinkurl{reformer} & \nolinkurl{hf-internal-testing/tiny-random-reformer} & fp32 & $8\!\times\!10^{-6}$ \\
\nolinkurl{regnet} & Constructor defaults & bf16 & 0 \\
\nolinkurl{rembert} & \nolinkurl{google/rembert} & bf16 & 0 \\
\nolinkurl{resnet} & Constructor defaults & bf16 & 0 \\
\nolinkurl{rf_detr} & \nolinkurl{Roboflow/rf-detr-base} & bf16 & 113$^\dagger$ \\
\nolinkurl{roberta} & \nolinkurl{FacebookAI/roberta-base} & bf16 & 0 \\
\nolinkurl{roberta_prelayernorm} & \nolinkurl{andreasmadsen/efficient_mlm_m0.40} & bf16 & 0 \\
\nolinkurl{roc_bert} & \nolinkurl{weiweishi/roc-bert-base-zh} & bf16 & 0 \\
\nolinkurl{roformer} & \nolinkurl{junnyu/roformer_chinese_base} & bf16 & 0 \\
\nolinkurl{rt_detr} & \nolinkurl{PekingU/rtdetr_r50vd} & bf16 & 0 \\
\nolinkurl{rt_detr_resnet} & Constructor defaults & bf16 & 0 \\
\nolinkurl{rt_detr_v2} & \nolinkurl{PekingU/rtdetr_v2_r18vd} & bf16 & 0.387 \\
\nolinkurl{rwkv}$^{*}$ & --- & --- & --- \\
\nolinkurl{sam} & \nolinkurl{facebook/sam-vit-base} & bf16 & 0 \\
\nolinkurl{sam2} & \nolinkurl{danelcsb/sam2.1_hiera_tiny} & bf16 & 0 \\
\nolinkurl{sam2_video} & \nolinkurl{facebook/sam2.1-hiera-tiny} & bf16 & 0 \\
\nolinkurl{sam3} & Constructor defaults & bf16 & 0 \\
\nolinkurl{sam3_lite_text} & \nolinkurl{yonigozlan/sam3-litetext-s0} & bf16 & 0 \\
\nolinkurl{sam3_tracker} & Constructor defaults & bf16 & 0.814 \\
\nolinkurl{sam3_tracker_video} & Constructor defaults & bf16 & 0 \\
\nolinkurl{sam3_video}$^{*}$ & --- & --- & --- \\
\nolinkurl{sapiens2} & \nolinkurl{facebook/sapiens2-pretrain-0.4b} & bf16 & 0 \\
\nolinkurl{seamless_m4t} & \nolinkurl{facebook/hf-seamless-m4t-medium} & bf16 & 0 \\
\nolinkurl{seamless_m4t_v2} & \nolinkurl{facebook/seamless-m4t-v2-large} & bf16 & 0 \\
\nolinkurl{seed_oss} & \nolinkurl{ByteDance-Seed/Seed-OSS-36B-Instruct} & bf16 & 0.839 \\
\nolinkurl{segformer} & \nolinkurl{nvidia/segformer-b0-finetuned-ade-512-512} & bf16 & 0.456 \\
\nolinkurl{seggpt} & \nolinkurl{BAAI/seggpt-vit-large} & bf16 & 0 \\
\nolinkurl{sew} & Constructor defaults & bf16 & 0 \\
\nolinkurl{sew_d} & Constructor defaults & bf16 & 0 \\
\nolinkurl{shieldgemma2} & Official ShieldGemma2 converter recipe & bf16 & 0 \\
\nolinkurl{siglip} & \nolinkurl{google/siglip-base-patch16-224} & bf16 & 0.529 \\
\nolinkurl{siglip2} & Constructor defaults & bf16 & 0.755 \\
\nolinkurl{slanet} & \nolinkurl{PaddlePaddle/SLANet_plus_safetensors} & fp32 & $1.02\!\times\!10^{-4}$ \\
\nolinkurl{slanext} & \nolinkurl{PaddlePaddle/SLANeXt_wired_safetensors} & bf16 & 0 \\
\nolinkurl{smollm3} & \nolinkurl{HuggingFaceTB/SmolLM3-3B} & bf16 & 0.85 \\
\nolinkurl{smolvlm} & \nolinkurl{HuggingFaceTB/SmolVLM2-2.2B-Instruct} & bf16 & 0 \\
\nolinkurl{solar_open} & \nolinkurl{upstage/Solar-Open-100B} & bf16 & 0.61 \\
\nolinkurl{speech_to_text} & \nolinkurl{facebook/s2t-small-librispeech-asr} & bf16 & 0 \\
\nolinkurl{speecht5} & \nolinkurl{microsoft/speecht5_asr} & bf16 & 0 \\
\nolinkurl{splinter} & \nolinkurl{tau/splinter-base} & bf16 & 0 \\
\nolinkurl{squeezebert} & \nolinkurl{squeezebert/squeezebert-uncased} & bf16 & 0.798 \\
\nolinkurl{stablelm} & \nolinkurl{stabilityai/stablelm-3b-4e1t} & bf16 & 0 \\
\nolinkurl{starcoder2} & \nolinkurl{bigcode/starcoder2-7b} & bf16 & 0.815 \\
\nolinkurl{step3p7} & \nolinkurl{stepfun-ai/Step-3.7-Flash} & bf16 & 11.2$^\dagger$ \\
\nolinkurl{superglue} & \nolinkurl{magic-leap-community/superglue_outdoor} & bf16 & 0 \\
\nolinkurl{superpoint} & \nolinkurl{magic-leap-community/superpoint} & bf16 & 0 \\
\nolinkurl{swiftformer} & Constructor defaults & bf16 & 0 \\
\nolinkurl{swin} & Constructor defaults & bf16 & 0 \\
\nolinkurl{swin2sr} & Constructor defaults & bf16 & 1.16 \\
\nolinkurl{swinv2} & Constructor defaults & bf16 & 0 \\
\nolinkurl{switch_transformers} & \nolinkurl{google/switch-base-8} & bf16 & 0 \\
\nolinkurl{t5} & \nolinkurl{google-t5/t5-small} & bf16 & 0 \\
\nolinkurl{t5gemma} & Author T5GemmaPreset.GEMMA2\_2B\_2B PREFIXLM IT recipe & bf16 & 0.688 \\
\nolinkurl{t5gemma2} & \nolinkurl{jordimas/t5gemma-2-270m-270m} & bf16 & 0 \\
\nolinkurl{table_transformer} & \nolinkurl{microsoft/table-transformer-detection} & bf16 & 0 \\
\nolinkurl{tapas} & \nolinkurl{google/tapas-base} & bf16 & 0 \\
\nolinkurl{textnet} & \nolinkurl{czczup/textnet-base} & bf16 & 0 \\
\nolinkurl{timesfm} & \nolinkurl{google/timesfm-2.0-500m-pytorch} & bf16 & 0 \\
\nolinkurl{timesfm2_5} & \nolinkurl{google/timesfm-2.5-200m-transformers} & bf16 & 0 \\
\nolinkurl{timesformer} & Constructor defaults & bf16 & 1.21 \\
\nolinkurl{timm_wrapper}$^{*}$ & --- & --- & --- \\
\nolinkurl{tipsv2} & \nolinkurl{google/tipsv2-b14} & bf16 & 1.86 \\
\nolinkurl{tipsv2_dpt} & \nolinkurl{google/tipsv2-b14-dpt} & bf16 & 2.23$^\dagger$ \\
\nolinkurl{trocr} & Constructor defaults & bf16 & 0 \\
\nolinkurl{tvp} & \nolinkurl{Intel/tvp-base} & bf16 & 0 \\
\nolinkurl{udop} & \nolinkurl{microsoft/udop-large} & bf16 & 0 \\
\nolinkurl{umt5} & \nolinkurl{google/umt5-small} & bf16 & 0 \\
\nolinkurl{unispeech} & Constructor defaults & bf16 & 0 \\
\nolinkurl{unispeech_sat} & Constructor defaults & bf16 & 0 \\
\nolinkurl{univnet} & \nolinkurl{dg845/univnet-dev} & bf16 & 0 \\
\nolinkurl{upernet} & \nolinkurl{openmmlab/upernet-convnext-tiny} & bf16 & 0.323 \\
\nolinkurl{uvdoc} & \nolinkurl{PaddlePaddle/UVDoc_safetensors} & bf16 & 0 \\
\nolinkurl{vaultgemma} & Constructor defaults & bf16 & 0.621 \\
\nolinkurl{vibevoice} & \nolinkurl{vibevoice/VibeVoice-1.5B-hf} & bf16 & 1.16 \\
\nolinkurl{vibevoice_acoustic_tokenizer} & \nolinkurl{microsoft/VibeVoice-AcousticTokenizer} & bf16 & 0 \\
\nolinkurl{vibevoice_asr} & \nolinkurl{microsoft/VibeVoice-ASR-HF} & bf16 & 0 \\
\nolinkurl{video_llama_3} & \nolinkurl{lkhl/VideoLLaMA3-2B-Image-HF} & bf16 & 0.624 \\
\nolinkurl{video_llava} & \nolinkurl{LanguageBind/Video-LLaVA-7B-hf} & bf16 & 0.716 \\
\nolinkurl{videomae} & \nolinkurl{MCG-NJU/videomae-base} & bf16 & 0 \\
\nolinkurl{videomt} & \nolinkurl{tue-mps/videomt-dinov2-small-ytvis2019} & bf16 & 0 \\
\nolinkurl{videoprism} & Constructor defaults & bf16 & 0 \\
\nolinkurl{vilt} & \nolinkurl{dandelin/vilt-b32-finetuned-coco} & bf16 & 0 \\
\nolinkurl{vipllava} & \nolinkurl{llava-hf/vip-llava-7b-hf} & fp16 & 0.0971 \\
\nolinkurl{visual_bert} & \nolinkurl{uclanlp/visualbert-vqa-coco-pre} & bf16 & 0 \\
\nolinkurl{vit} & Constructor defaults & bf16 & 0 \\
\nolinkurl{vit_mae} & \nolinkurl{facebook/vit-mae-base} & bf16 & 0 \\
\nolinkurl{vit_msn} & \nolinkurl{facebook/vit-msn-small} & bf16 & 0 \\
\nolinkurl{vitdet} & Constructor defaults & bf16 & 0 \\
\nolinkurl{vitmatte} & \nolinkurl{hustvl/vitmatte-small-composition-1k} & bf16 & 0 \\
\nolinkurl{vitpose} & \nolinkurl{usyd-community/vitpose-base-simple} & bf16 & 0 \\
\nolinkurl{vitpose_backbone} & Constructor defaults & bf16 & 0 \\
\nolinkurl{vits} & \nolinkurl{facebook/mms-tts-eng} & bf16 & 0 \\
\nolinkurl{vivit} & \nolinkurl{google/vivit-b-16x2-kinetics400} & bf16 & 0 \\
\nolinkurl{vjepa2} & Constructor defaults & bf16 & 0 \\
\nolinkurl{voxtral} & \nolinkurl{mistralai/Voxtral-Mini-3B-2507} & bf16 & 0 \\
\nolinkurl{voxtral_realtime} & \nolinkurl{mistralai/Voxtral-Mini-4B-Realtime-2602} & bf16 & 0 \\
\nolinkurl{wav2vec2} & \nolinkurl{facebook/wav2vec2-large-960h-lv60-self} & bf16 & 0 \\
\nolinkurl{wav2vec2_bert} & \nolinkurl{facebook/w2v-bert-2.0} & bf16 & 0 \\
\nolinkurl{wav2vec2_conformer} & \nolinkurl{facebook/wav2vec2-conformer-rel-pos-large} & bf16 & 0 \\
\nolinkurl{wavlm} & Constructor defaults & bf16 & 0 \\
\nolinkurl{whisper} & \nolinkurl{openai/whisper-tiny.en} & bf16 & 0.795 \\
\nolinkurl{x_clip} & \nolinkurl{microsoft/xclip-base-patch32} & bf16 & 0 \\
\nolinkurl{xcodec} & \nolinkurl{hf-audio/xcodec-hubert-librispeech} & bf16 & 0 \\
\nolinkurl{xcodec2} & \nolinkurl{HKUSTAudio/xcodec2-hf} & bf16 & 10.1$^\dagger$ \\
\nolinkurl{xglm} & \nolinkurl{facebook/xglm-564M} & bf16 & 0 \\
\nolinkurl{xlm} & \nolinkurl{FacebookAI/xlm-mlm-en-2048} & bf16 & 1.11 \\
\nolinkurl{xlm_roberta} & \nolinkurl{FacebookAI/xlm-roberta-base} & bf16 & 0 \\
\nolinkurl{xlm_roberta_xl} & \nolinkurl{facebook/xlm-roberta-xl} & bf16 & 1.98 \\
\nolinkurl{xlnet} & Constructor defaults & fp32 & $2.25\!\times\!10^{-4}$ \\
\nolinkurl{xmod} & \nolinkurl{facebook/xmod-base} & bf16 & 1.48 \\
\nolinkurl{yolos} & \nolinkurl{hustvl/yolos-tiny} & bf16 & 0 \\
\nolinkurl{yoso} & Constructor defaults & bf16 & 0 \\
\nolinkurl{youtu} & \nolinkurl{tencent/Youtu-LLM-2B} & bf16 & 0.974 \\
\nolinkurl{zamba} & \nolinkurl{Zyphra/Zamba-7B-v1} & bf16 & 0.512 \\
\nolinkurl{zamba2} & Constructor defaults & bf16 & 1.22 \\
\nolinkurl{zaya} & \nolinkurl{Zyphra/ZAYA1-8B} & bf16 & 0.668 \\
\nolinkurl{zoedepth} & \nolinkurl{Intel/zoedepth-nyu-kitti} & bf16 & 0.00553 \\
\end{longtable}
\endgroup

\end{document}